\documentclass[10pt,twocolumn,letterpaper]{article}

\usepackage[]{cvpr}      

\usepackage{graphicx}
\usepackage{amsmath}
\usepackage{amssymb}
\usepackage{booktabs}
\RequirePackage[dvipsnames,table]{xcolor}
\usepackage{natbib}
\usepackage{booktabs}
\usepackage{graphicx}
\usepackage{booktabs}
\usepackage{multirow}
\usepackage{makecell}
\usepackage{booktabs}
\usepackage{colortbl}
\usepackage{rotating}
\usepackage{array}
\usepackage{enumitem}

\usepackage{pifont}
\newcommand{\xmark}{\ding{55}}

\definecolor{cvprblue}{rgb}{0.21,0.49,0.74}
\usepackage[pagebackref,breaklinks,colorlinks,citecolor=cvprblue]{hyperref}

\def\paperID{*****} 
\def\confName{CVPR}
\def\confYear{2025}

\title{SUM: Saliency Unification through Mamba for Visual Attention Modeling}

\author{
\quad Alireza Hosseini\footnotemark[1]$^{\;\, ,1}$ 
\quad Amirhossein Kazerouni\footnotemark[1]$^{\;\,, 2, 3, 4}$ 
\quad Saeed Akhavan$^{\;1}$ \\ 
\quad Michael Brudno$^{\;2, 3, 4}$ 
\quad Babak Taati$^{\;2, 3, 4}$ 
\\
${^1}$ University of Tehran
${^2}$ University of Toronto 
${^3}$ Vector Institute \\
${^4}$ University Health Network \\
{\tt\small \{arhosseini77, s.akhavan\}@ut.ac.ir, \{amirhossein, brudno\}@cs.toronto.edu} \\
{\tt\small babak.taati@uhn.ca} 
}

\begin{document}
\maketitle

\footnotetext[1]{Equal contribution}

\begin{abstract}
Visual attention modeling, important for interpreting and prioritizing visual stimuli, plays a significant role in applications such as marketing, multimedia, and robotics. Traditional saliency prediction models, especially those based on Convolutional Neural Networks (CNNs) or Transformers, achieve notable success by leveraging large-scale annotated datasets. However, the current state-of-the-art (SOTA) models that use Transformers are computationally expensive. Additionally, separate models are often required for each image type, lacking a unified approach. In this paper, we propose \textbf{S}aliency \textbf{U}nification through \textbf{M}amba \textbf{(SUM)}, a novel approach that integrates the efficient long-range dependency modeling of Mamba with U-Net to provide a unified model for diverse image types. Using a novel Conditional Visual State Space (C-VSS) block, SUM dynamically adapts to various image types, including natural scenes, web pages, and commercial imagery, ensuring universal applicability across different data types. Our comprehensive evaluations across five benchmarks demonstrate that SUM seamlessly adapts to different visual characteristics and consistently outperforms existing models. These results position SUM as a versatile and powerful tool for advancing visual attention modeling, offering a robust solution universally applicable across different types of visual content. Our code and pretrained models are available at \url{https://github.com/Arhosseini77/SUM}.

\end{abstract}    
\section{Introduction}
\label{sec:intro}
Visual attention is a critical function of the human visual system, enabling the selection of the most relevant information in a visual scene~\cite{jonides1982integrating}. Modeling of this mechanism, known as saliency prediction, plays pivotal roles in numerous applications such as marketing~\cite{hosseini2024brand, jiang2023ueyes}, multimedia~\cite{mishra2021multi}, computer vision~\cite{patel2021saliency}, and robotics~\cite{chen2021video}.

Deep learning models have succeeded in saliency prediction, by exploiting large-scale annotated datasets~\cite{borji2015cat2000, jiang2015salicon}. Typically, these models employ a pre-trained object recognition network for feature extraction~\cite{kummerer2014deep}, with the U-Net architecture as a popular choice. Most methods employ Convolutional Neural Networks (CNNs) to construct encoders and decoders for latent features, which generate visual saliency maps~\cite{hu2021fastsal, wang2017deep, kroner2020contextual, jia2020eml, droste2020unified, che2019gaze}. Recurrent architectures, such as Long-Short Term Memory (LSTM) networks, are also sometimes used to model both local and long-range visual information~\cite{cornia2018predicting, liu2018deep}, enhancing the accuracy of saliency predictions. More recently, the use of Transformer-based models has led to significant improvements, achieving SOTA performance in saliency prediction by learning spatial long-range dependencies~\cite{han2022survey, lou2022transalnet, djilali2024learning, hosseini2024brand, li2023uniar}. However, the computational demands of the standard self-attention mechanism used in these methods, which scales quadratically with image size, present a substantial challenge, especially for dense prediction tasks like saliency modeling.

Moreover, a significant limitation within current saliency prediction models lies in their design specificity for singular visual contexts. Saliency maps, and consequently the models that generate them, need to be adapted to the unique characteristics of different types of images. For example, in natural scenes, the visual attention of viewers may be driven largely by elements like color and movement, whereas in e-commerce images, textual information typically attracts more attention~\cite{jiang2022does}. Similarly, in user interface (UI) designs, the upper-left quadrant often attracts more attention due to common eye movement patterns and a left-to-right viewing bias ~\cite{jiang2023ueyes}. Although there are robust models tailored for specific datasets, such as those optimized for commercial imagery~\cite{hosseini2024brand} or UIs~\cite{jiang2023ueyes}, the research on the development of universally applicable models, that can effectively handle diverse requirements of various image types, remains limited. This gap underscores the necessity for a model, which can be universally performant across all image types and saliency datasets, thus providing a more comprehensive solution to the field of saliency prediction.

To address the challenges outlined above, we leverage the capabilities of State Space Models (SSMs)~\cite{kalman1960new} as used in Mamba~\cite{gu2023mamba, liu2024vmamba}, and introduce a novel unified Mamba-U-Net-based model for visual saliency prediction. Models like Mamba capture long-distance dependencies with linear computational complexity. Inspired by these successes, we propose the \textbf{S}aliency \textbf{U}nification through \textbf{M}amba (\textbf{SUM}), which uses Mamba to efficiently capture long-range information. To ensure universal applicability across diverse image types, we incorporate a novel Conditional Visual State Space (C-VSS) block in our design. This component effectively separates the distributions of different data types, making the model robust across various modalities. It allows SUM to dynamically adapt to distinct visual characteristics found in natural scenes, e-commerce imagery, and UIs. Validation of SUM on six large-scale datasets across different visual contexts confirms its exceptional adaptability and strong performance, positioning it as a potent tool in the advancement of visual attention modeling. These attributes make SUM a valuable tool for a range of applications in visual saliency prediction. The \textbf{main contributions} of this work are summarized as follows:
\begin{itemize}
    \item A novel efficient class-conditional unified model is proposed that employs Mamba to capture long-range visual information efficiently with linear computational complexity.
    \item A conditional component that dynamically adapts the model behavior at test time through the shift and scaling mechanisms, enhancing the adaptability of the model to various visual contexts.
    \item SUM is extensively evaluated on six diverse benchmark datasets, including natural scenes with gaze or mouse ground truth labels, web pages, and commercial images, consistently demonstrating superior or competitive performance against previous SOTA models.
\end{itemize}

\section{Related work}
\label{sec:related}

\textbf{Saliency Prediction:} Saliency prediction models are designed to identify areas within an image or video that capture human visual attention. Initially inspired by biological insights, these models historically used contrasts in color, intensity, and orientation, or low-level, hand-designed features to mimic human visual perception. This approach was based on cues from studies of how humans prioritize visual information~\cite{jiang2015image, rajashekar2008gaffe, goferman2011context}. 

With the advent of deep learning and the availability of large-scale eye-tracking datasets~\cite{jiang2021deepvs2, jiang2015salicon, borji2015cat2000}, there has been a shift towards applying deep neural networks to the problem of saliency prediction. This shift was marked by significant improvements in the accuracy and reliability of saliency models~\cite{vig2014large}. Kummerer et al.~\cite{kummerer2014deep} demonstrated that leveraging pretrained networks, originally designed for object recognition tasks, could enhance the performance of saliency prediction models. This insight paved the way for subsequent models such as EML-Net~\cite{jia2020eml}, DeepGaze II~\cite{kummerer2016deepgaze}, and SALICON~\cite{huang2015salicon}, which incorporated pretrained CNN encoders to enhance the prediction of saliency maps. Beyond the use of pretrained CNNs, researchers have explored various other network architectures for saliency prediction. These include fully convolutional networks (FCNs)~\cite{kruthiventi2017deepfix}, generative adversarial networks (GANs)~\cite{che2019gazegan, pan2017salgan}, and convolutional long short-term memory networks (ConvLSTM)~\cite{jiang2021deepvs2}. Attention mechanism and Transformer models~\cite{vaswani2017attention}, which show remarkable success in various vision tasks~\cite{han2022survey}, have also been applied to saliency prediction. Models like VGG-SSM~\cite{cornia2018predicting} and TranSalNet~\cite{lou2022transalnet}, incorporate self-attention modules and transformer-based methods, respectively. These approaches highlight the growing interest in leveraging advanced architectures that go beyond traditional CNNs to improve saliency.
 

Saliency prediction has also expanded to cover diverse types of data beyond natural scenes, including commercial advertisements~\cite{jiang2022does, leveque2019eye, liang2021fixation} and user interfaces~\cite{jiang2023ueyes, shen2014webpage}. This diversification has led to specialized models that address unique dataset challenges. For instance, Kou et al.\cite{kou2023advertising} proposed a method for integrating confidence scores in saliency predictions for advertising images, enhancing both robustness and performance. Similarly, Jiang et al.~\cite{jiang2022does} introduced salient Swin-Transformers and incorporated text detection techniques into their models, demonstrating the potential of combining various data modalities to improve prediction accuracy. Following this trend, Hosseini et al.~\cite{hosseini2024brand} proposed a model that combines pretrained CNNs and transformers with a text map detection module for advertising saliency prediction.

\begin{figure*}[t]
    \centering
    \includegraphics[width=\textwidth]{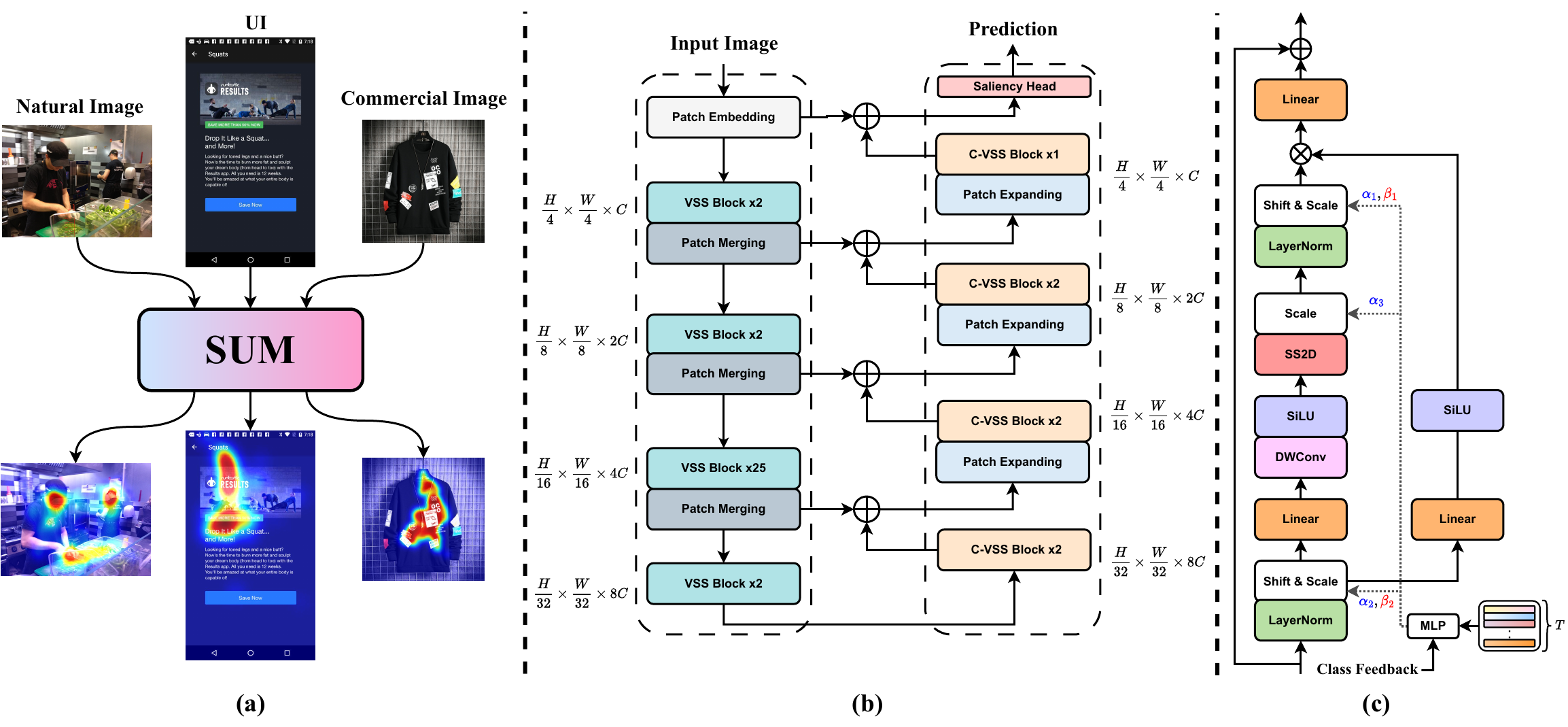}
    \caption{(a) Overview of our \textbf{SUM} model, (b) conditional-U-Net-based model for saliency prediction, and (c) C-VSS module.}
    \label{fig:SUM_Network}
    \vspace{-1em}
\end{figure*}

\noindent\textbf{Mamba:} Recent advancements in SSMs, particularly with the development of the Mamba model~\cite{gu2023mamba}, have significantly changed the landscape of computational modeling. This model offers a promising alternative to traditional attention-based models. Introduced by Gu et al.~\cite{gu2023mamba}, Mamba achieves linear computational complexity with respect to input size and is effective at capturing long-distance dependencies. This innovation has led to its broad application in fields such as language understanding and vision tasks ~\cite{pioro2024moe, liu2024vmamba, zhu2024vision, pei2024efficientvmamba, hu2024zigma, yang2024plainmamba}. The development of vision-specific SSMs such as Vision Mamba~\cite{zhu2024vision} and Vmamba~\cite{liu2024vmamba} has marked a significant step in SSM development. Notable examples include U-Mamba~\cite{ma2024u}, which combines SSMs with CNNs for medical image segmentation. SegMamba~\cite{xing2024segmamba} integrates SSMs in its encoder and uses a CNN-based decoder for 3D brain tumor segmentation. VM-UNet~\cite{ruan2024vm} explores a purely SSM-based approach in this area. Other models, like LightM-UNet~\cite{liao2024lightm}, stand out for their efficiency, outperforming previous medical segmentation models with fewer parameters. Additionally, Mamba's versatility is demonstrated in video-based applications such as video medical segmentation~\cite{yang2024vivim} and understanding~\cite{chen2024video}. However, the use of Mamba in saliency map prediction is still largely unexplored.

\noindent\textbf{Unified Models:} 
Unified models in visual saliency prediction have made significant advancements in integrating image and video saliency within a single framework. The UNISAL model~\cite{droste2020unified} is an example that addresses the integration of image and video saliency through domain adaptation. However, while UNISAL is a lightweight model, it is not a universal model for all image saliency datasets. It primarily relies on the Salicon Dataset~\cite{jiang2015salicon} for image saliency prediction, and its performance on this dataset has been outperformed by other models over time. Furthermore, UNISAL's universal model does not include diverse image types, limiting its applicability. As another notable model, UniAR~\cite{li2023uniar}, focuses on image-based saliency prediction and incorporates a multimodal transformer to capture diverse human behaviors across various visual content and tasks. While it encompasses UI and natural scene images, it overlooks incorporating e-commercial images, which have become increasingly important in recent years~\cite{jiang2022does, hosseini2024brand}. Additionally, UniAR's complexity is highlighted by its model size, with 848M parameters, making it computationally demanding and potentially limiting its practical use. Despite advancements in existing unified saliency models, there is still a significant gap in developing efficient, comprehensive models that effectively address real-world needs across diverse image types while maintaining manageable complexity. 

\section{Proposed Method}
This section provides an overview of the proposed network architecture, as shown in \autoref{fig:SUM_Network}(a). Next, we revisit the concept of VSS as introduced by Liu et al.~\cite{liu2024vmamba}. Building on this foundation, we introduce our novel C-VSS module and a conditional Mamba-U-Net-based model for visual saliency prediction.

\subsection{Model Architecture}
The architecture of \textbf{SUM}, as illustrated in \autoref{fig:SUM_Network}(b), adopts a U-Net configuration. The process initiates with an input image \(X \in \mathbb{R}^{H \times W \times 3}\), with spatial dimensions $H$ and $W$, and $3$ channels, which undergoes initial transformation via a patch embedding module, reducing its dimensions to \(\frac{H}{4} \times \frac{W}{4} \times C\). The encoder module generates four hierarchical output representations. Each stage is followed by a downsampling layer, which reduces the spatial dimensions by half while simultaneously doubling the number of channels. Transitioning to the decoder, comprises four stages of C-VSS layers, with each stage incorporating two blocks, except for the final stage, which contains a single block. Patch-expanding layers are then applied to achieve resolution upsampling while also decreasing channel dimensions by a factor of 2. Finally, a linear layer is responsible for generating the ultimate output. Our SUM architecture uses the VMamba~\cite{liu2024vmamba} weights pre-trained on ImageNet~\cite{deng2009imagenet}. This pre-training accelerates the learning process, improves the model's ability to detect salient regions more accurately, and ensures better generalization on diverse images.

\subsection{Visual State Space (VSS)}

Mamba~\cite{gu2023mamba} employs SSMs~\cite{gu2021efficiently} to shift the complexity of attention from quadratic to linear in long-sequence modeling. This has proven particularly beneficial in vision tasks due to its higher accuracy, reduced computational load, and lower memory requirements~\cite{zhu2024vision}. However, adapting Mamba's inherently 1D, causal scanning for 2D images presents challenges due to its restricted receptive field and inability to process unscanned data effectively. To address these issues, VMamba~\cite{liu2024vmamba} introduces the Cross-Scan Module, which employs bidirectional scanning along horizontal and vertical axes. This module expands the image into sequences of patches scanned in four directions, enabling each pixel to integrate information from all directions. Subsequently, these sequences are reassembled into the original 2D format to form a complete image. Termed the 2D-Selective-Scan (SS2D), this method enhances Mamba's functionality for 2D spatial processing, ensuring both local and global spatial relevance. Building upon these insights, we incorporate the VSS block as the fundamental unit in SUM. As shown in \autoref{fig:SUM_Network}(c), the VSS module can be formulated as:
\begin{equation}
    \begin{aligned}
    &\textbf{X} = \text{LN}_1(\textbf{F}), \\
    &\textbf{Attention} = \text{LN}_2(\text{SS2D}(\text{SiLU}(\text{DW-Conv}(\text{Linear}(\textbf{X}))))),\\
    &\textbf{Output} = \text{Linear}(\text{SiLU}(\text{Linear}(\textbf{X})) \otimes \textbf{Attention}) + \textbf{F},
    \end{aligned}
\end{equation}
where the input feature is denoted by $\textbf{F} \in \mathbb{R}^{H' \times W' \times C'}$. The operator $\otimes$ denotes an element-wise product operation, LN represents LayerNorm, DW-Conv stands for depth-wise convolution, and SiLU~\cite{elfwing2018sigmoid} is an activation function.

\subsection{Conditional Visual State Space (C-VSS)}
We enhance the model's adaptability to diverse visual content by conditioning the VSS block in the decoder based on input type. This is crucial for predicting saliency maps effectively, as different content types inherently attract viewer attention in distinct ways. For instance, natural scenes may focus on color and movement, e-commerce images on textual information, and UI designs on specific layout patterns such as the upper-left quadrant. To address these variations, we implement modulation of the feature map through dynamic scaling and shifting operations that adjust feature activations based on the input type. The modulated feature map can be generally defined as:
\[
\text{Modulated Feature Map} = \textcolor{blue}{\alpha_i} \odot \mathbf{F} + \textcolor{red}{\beta_i}
\]
where \( \mathbf{F} \) denotes the original feature map. Here, \textcolor{blue}{$\alpha$} is a scaling factor, \textcolor{red}{$\beta$} is a shifting factor, and $\odot$ is an element-wise multiplication.

To refine our model's ability to effectively handle different data types, we define \( T=4 \) learnable tokens, where \( D \) represents the dimensionality of each token. Each token is designated to capture distinct information about one of the following data categories: \texttt{Natural Scene-Mouse}, \texttt{Natural Scene-Eye}, \texttt{E-Commerce}, and \texttt{UI}. These tokens provide a more nuanced mechanism than a simple one-hot encoding of data types, enabling the model to adapt and learn detailed, type-specific information. We have allocated two tokens for the natural scene data because different methodologies are used in data collection for these categories, \texttt{eye} and \texttt{mouse}. Grouping them into a single token could potentially confuse the model during inference. As discussed in~\cite{tavakoli2017saliency}, mouse tracking data is less consistent and more scattered than eye tracking data, which does not fully align with eye tracking data distribution, particularly in terms of different contextual regions. Furthermore, while mouse tracking data can lead to acceptable outcomes for training existing models, it is less reliable for model selection and evaluation. Based on these insights and our experiments, we differentiate mouse and eye data of natural scenes.

Subsequently, the relevant token is fed into a Multi-Layer Perceptron (MLP) model to ensure that learning is conditioned on the specific characteristics of each data type. The MLP is composed of \( K \) hidden layers and \( p_1, p_2, \ldots, p_K \) features per layer. This MLP is designed to regress the parameters \textcolor{blue}{\( \alpha_i \)} and \textcolor{red}{\( \beta_i \)}, which modulate the model based on the diversity of inputs. The MLP, defined as \( g(\mathbf{z}; \theta): \mathbb{R}^{4 \times D} \rightarrow \mathbb{R}^{4 \times 5} \), outputs a matrix \( \mathbf{Y} \), with each row representing one of four input tokens and generating five key parameters. These parameters include pairs and individual instances of \textcolor{blue}{\( \alpha_i \)} and \textcolor{red}{\( \beta_i \)}, specifically \(\{ (\textcolor{blue}{\alpha_1}, \textcolor{red}{\beta_1}), (\textcolor{blue}{\alpha_3}), (\textcolor{blue}{\alpha_2}, \textcolor{red}{\beta_2}) \}\). An input label \( L \) determines the selection of the relevant row from \( \mathbf{Y} \), resulting in the output vector \( \mathbf{S} = \mathbf{Y}_L \). This \( 1 \times 5 \) vector contains modulation parameters finely tuned to the specifics of the designated input. These parameters are then integrated into the model to modify its behavior dynamically: \( (\textcolor{blue}{\alpha_1}, \textcolor{red}{\beta_1}) \) are used to shift and scale \( \text{LN}_1 \), \((\textcolor{blue}{\alpha_3})\) adjusts the scaling of the SS2D block to regulate feature intensity, and \((\textcolor{blue}{\alpha_2}, \textcolor{red}{\beta_2})\) shift and scale \( \text{LN}_2 \). This enables the MLP to precisely control the normalization and scaling within the model, thereby enhancing its performance and generalization across different visual content types.

\subsection{Loss Function}
Our model utilizes a composite loss function inspired by~\cite{droste2020unified,lou2022transalnet, aydemir2023tempsal} in visual saliency prediction. This function integrates five distinct components, each designed to optimize the prediction accuracy of saliency maps by targeting different aspects of the saliency prediction task. The loss function is formulated as:
\begin{equation}
\begin{aligned}
\text{Loss} = & \lambda_1 \cdot \mathcal{L_{\text{KL}}}(s^{g}, s)
             + \lambda_2 \cdot \mathcal{L_{\text{CC}}}(s^{g}, s) 
             + \lambda_3 \cdot \mathcal{L_{\text{SIM}}}(s^{g}, s) \\
             & + \lambda_4 \cdot \mathcal{L_{\text{NSS}}}(f^{g}, s)
             + \lambda_5 \cdot \mathcal{L_{\text{MSE}}}(s^{g}, s)
\end{aligned}
\label{eq:loss}
\end{equation}
where \(s^{g}\) represents the ground truth saliency map, \(f^{g}\) denotes the ground truth fixation map, and \(s\) is the network's predicted saliency map. Each component of the loss function serves a specific purpose as defined in the following.

\noindent\textbf{Kullback-Leibler Divergence (KL):} KL divergence measures the dissimilarity between the predicted and ground truth distributions, providing a method to penalize the model when its predictions deviate significantly from the actual data distribution. 
\begin{equation}
    \mathcal{L_{\text{KL}}}(s^{g}, s) = \sum_{i=1}^{n} s^{g}_i \log\left(\epsilon + \frac{s_i}{s^{g}_i + \epsilon}\right),
    \label{eq:kl}
\end{equation}
where, the regularization constant \(\epsilon\) is set to \(2.2 \times 10^{-16}\).

\noindent\textbf{Linear Correlation Coefficient (CC):} The correlation coefficient assesses the linear relationship between the predicted and ground truth saliency maps. A higher correlation indicates that the model predictions align well with the ground truth trends, improving the reliability of the saliency maps.
\begin{equation}
    \mathcal{L_{\text{CC}}}(s^{g}, s) = \frac{\text{cov}(s^{g}, s)}{\sigma(s^{g}) \cdot \sigma(s)},
    \label{eq:cc}
\end{equation}
where, \(\text{cov}(.)\) represents the covariance and \(\sigma(.)\) denotes the standard deviation.   

\noindent\textbf{Similarity (SIM):}
SIM evaluates the overlap between the predicted and actual saliency maps, emphasizing the importance of accurately predicting the salient regions.
\begin{equation}
\mathcal{L_{\text{SIM}}}(s^{g}, s) = \sum_{i=1}^{n} \min(s^{g}_i, s_i)
\label{eq:sim}
\end{equation}

\noindent\textbf{Normalized Scan-path Saliency (NSS):}
NSS measures the correlation between the normalized predicted saliency map and the actual fixation points, highlighting the model's effectiveness at capturing human attention patterns.
\begin{equation}
    \mathcal{L_{\text{NSS}}}(f^{g}, s) = \frac{1}{\sum_i (f^{g}_i)} \sum_i \left(\frac{s_i - \mu(s)}{\sigma(s)}\right)f^{g}_i
    \label{eq:nss}
\end{equation}

\noindent\textbf{Mean Squared Error (MSE):}
This component calculates the mean squared error between the predicted and actual saliency maps, directly penalizing inaccuracies in the pixel-wise saliency values.

By adjusting the weighting coefficients \(\lambda_i\) (\(i = 1, \ldots, 5\)), we aim to minimize dissimilarity metrics (KL, MSE) and maximize similarity metrics (CC, SIM, NSS). This strategy ensures that the model predicts accurate saliency maps and closely aligns with human visual attention patterns and saliency distributions.
\section{Experiments}
\noindent\textbf{Datasets:}
We leverage six benchmark large-scale datasets for training and evaluating our models, as outlined in \autoref{datasets}. This table presents a list of these datasets along with specific details about each. 

\begin{table}[t]
\caption{Comprehensive compilation of datasets used for training and testing.}
\label{tab:train_dataset}
\centering
\resizebox{\columnwidth}{!}{
\begin{tabular}{lcccccc}
\toprule
Dataset & Image domain & Acquisition Type &  \# Image & Image Resolution & \# Training Sample \\
\midrule
\textit{Salicon}~\cite{jiang2015salicon} & Natural scene & Mouse & 15,000 & 640 $\times$ 480 & 10,000 \\
\textit{MIT1003}~\cite{judd2009learning} & Natural scene & Eye & 1003 & Varied & 904 \\
\textit{CAT2000}~\cite{borji2015cat2000} & Natural scene & Eye & 2000 & 1080 $\times$ 1920 & 1600 \\
\textit{OSIE}~\cite{xu2014predicting} & Natural scene  & Eye & 700 & 800 $\times$ 600 & 500 \\
\textit{U-EYE}~\cite{jiang2023ueyes} & Web page & Eye & 1979 & Varied & 1583 \\
\textit{SalECI}~\cite{jiang2022does} & E-Commercial & Eye &  972 & 720 $\times$ 720 & 871 \\
\bottomrule
\end{tabular}
}
\label{datasets}
\end{table}

\noindent\textbf{Evaluation Metrics:} To assess the accuracy of predicted saliency maps, we use two types of metrics: \textit{location-based} and \textit{distribution-based}, followed by \cite{bylinskii2018what}. Location-based metrics, such as NSS and AUC (Area under the ROC Curve), evaluate predictions using a binary fixation map image as ground truth and focus on specific salient locations. Distribution-based metrics, including CC (Correlation Coefficient), SIM (Similarity), and KLD (Kullback-Leibler Divergence), utilize a grayscale saliency map image to measure the similarity between predicted and actual distributions. Higher values generally indicate better performance for all metrics, except for KLD, where a value closer to zero signifies a more accurate prediction.

\noindent\textbf{Implementation Details:} Our model is implemented using the PyTorch framework and is trained on A40 with 48 GB memory for 15 epochs with an early stopping after 4 epochs. We optimize the network using the Adam optimizer. The learning rate is initially set to $1 \times 10^{-4}$, and we employ a learning rate scheduler that decreases the factor by 0.1 after every four epochs. The batch size is set to 16. Additionally, we resize all data and labels to a resolution of $256 \times 256$ and combine the training data from all six datasets for model training. The optimal values for the loss function weighting coefficients \(\lambda_i\) are as follows: \(\lambda_1 = 10\), \(\lambda_2 = -2\), \(\lambda_3 = -1\), \(\lambda_4 = -1\), and \(\lambda_5 = 5\). In addition, the MLP architecture in our implementation comprises three linear layers with widths of 128, 64, and 5, respectively, interleaved with GELU \cite{hendrycks2016gaussian} activation functions. The number of tokens is set to \( T = 4 \) with each token having a dimensionality of \( D = 128 \).

\begin{table*}[t]
\centering
\caption{Saliency prediction performance across various datasets. $^\ast$ indicates that we have trained those models ourselves for fair comparison because results were not available for the corresponding dataset or the input image size was varied. $^\dagger$ signifies that the results have been taken from the paper by Hosseini et al. \cite{hosseini2024brand}, and the rest of the results are taken from their respective papers. For our model, we note the percentage (\%) change in performance relative to the second-best result, or to the best result if ours is not the top performer.}
\label{tab:saliency}
\resizebox{\textwidth}{!}{
    \begin{tabular}{c|lccccc|c}
    \toprule
    Dataset & Method & CC $\uparrow$ & KLD $\downarrow$ & AUC $\uparrow$ & SIM $\uparrow$ & NSS $\uparrow$ &  \# Parameters \\
    \midrule
    
    \textit{U-EYE}~\cite{jiang2023ueyes} & SAM$^\ast$ ~\cite{cornia2018predicting} & 0.580 & 1.490 & 0.811 & 0.520  & 1.640 & 30M \\
    (Web page) & UMSI$^\ast$~\cite{fosco2020predicting} & 0.562 & 1.580 & 0.805 & 0.510  & 1.690 & 30M \\
    & SAM++$^\ast$ ~\cite{jiang2023ueyes} & 0.580 & 1.190 & 0.800 & 0.530  & 1.660 & 42M \\
    & Transalnet$^\ast$ \cite{lou2022transalnet} & 0.696 & 0.616 & 0.839 & 0.598 & 1.601  & 72M\\
    & UMSI++$^\ast$ ~\cite{jiang2023ueyes} & 0.670 & 0.860 & 0.830 & 0.580 & 1.610 & 30M\\
    \cmidrule{2-8}
    & \cellcolor[HTML]{DADCFF} \textbf{SUM (Ours)} & \cellcolor[HTML]{DADCFF} \textbf{0.731\scriptsize{$+$5.03$\%$}} & \cellcolor[HTML]{DADCFF} \textbf{0.544 \scriptsize{$-$11.69$\%$}} & \cellcolor[HTML]{DADCFF} \textbf{0.846 \scriptsize{$+$0.83$\%$}} & \cellcolor[HTML]{DADCFF} \textbf{0.630 \scriptsize{$+$5.35$\%$}}  & \cellcolor[HTML]{DADCFF} \textbf{1.704 \scriptsize{$+$0.83$\%$}}  & \cellcolor[HTML]{DADCFF} 57.5M  \\
    \midrule
    
    \textit{SalECI}~\cite{jiang2022does} & SSM$^\dagger$ ~\cite{cornia2018predicting} &  0.720 & 0.599 & 0.830 & 0.611 & 1.396 & 42M \\
    (E-Commercial) & DeepGaze IIE$^\dagger$~\cite{linardos2021deepgaze} & 0.560 & 0.995 & 0.842 & 0.399  & 1.327  & 104M \\
    & EML-NET$^\dagger$ ~\cite{jiang2023ueyes} & 0.510 & 1.220 & 0.807 & 0.536  & 1.232 & 47M \\
    & Transalnet$^\dagger$ ~\cite{jiang2023ueyes} & 0.717 & 0.873 & 0.824 &  0.534 & 1.723 & 72M\\
    & Temp-Sal$^\dagger$ \cite{aydemir2023tempsal} & 0.719 & 0.712 &  0.813 & 0.629 & 1.768 & 242M \\
    & SSwin Transformer \cite{jiang2022does} & 0.687 & 0.652 &  0.868 & 0.606 & 1.701 & 29M   \\
    & \citet{hosseini2024brand} & 0.750 & 0.578 &  0.892 & 0.645 & 1.890 & 66M  \\
    \cmidrule{2-8}
    & \cellcolor[HTML]{DADCFF} \textbf{SUM (Ours)} & \cellcolor[HTML]{DADCFF} \textbf{0.789\scriptsize{$+$5.20$\%$}} & \cellcolor[HTML]{DADCFF} \textbf{0.473 \scriptsize{$-$18.17$\%$}} & \cellcolor[HTML]{DADCFF} \textbf{0.899 \scriptsize{$+$0.78$\%$}} & \cellcolor[HTML]{DADCFF} \textbf{0.680 \scriptsize{$+$5.43$\%$}}  & \cellcolor[HTML]{DADCFF} \textbf{2.012 \scriptsize{$+$6.46$\%$}}  & \cellcolor[HTML]{DADCFF} 57.5M  \\
    \midrule
    
    \textit{OSIE}~\cite{xu2014predicting} & UMSI~\cite{fosco2020predicting} & 0.746 & 0.513 & 0.856 & 0.631 & 1.788  & 30M \\
    (Natural scene) & EML-NET~\cite{jia2020eml} & 0.717 & 0.537 & 0.854 &  0.619 & 1.737  & 47M \\
    & SAM-ResNet~\cite{cornia2018predicting} & 0.758 & 0.480 & 0.860 & 0.648 & 1.811  & 43M \\
    & \citet{chen2023learning} & 0.761 & 0.506 & 0.860 & 0.652 & 1.840 & - \\
    & Transalnet$^\ast$ ~\cite{jiang2023ueyes} & 0.791 & 0.667 & 0.923 &  0.651 & 2.448 & 72M\\
    & UniAR \cite{li2023uniar} & 0.754 & 0.547 & 0.867 & 0.647  & 1.842 & 848M  \\
    \cmidrule{2-8}
    & \cellcolor[HTML]{DADCFF} \textbf{SUM (Ours)} & \cellcolor[HTML]{DADCFF} \textbf{0.861\scriptsize{$+$8.85$\%$}} & \cellcolor[HTML]{DADCFF} \textbf{0.340\scriptsize{$-$29.17$\%$}} & \cellcolor[HTML]{DADCFF} \textbf{0.924\scriptsize{$+$6.57$\%$}} & \cellcolor[HTML]{DADCFF} \textbf{0.727 \scriptsize{$+$11.5$\%$}}  & \cellcolor[HTML]{DADCFF} \textbf{3.416 \scriptsize{$+$39.54$\%$}}  & \cellcolor[HTML]{DADCFF} 57.5M  \\
    \midrule
    
    \textit{Salicon}~\cite{jiang2015salicon} & UniAR \cite{li2023uniar} & 0.901 & 0.215 & 0.870 & 0.792  & 1.947 & 848M  \\
    (Natural scene) & SimpleNet \cite{reddy2020tidying} & 0.907 & {0.193} & {0.871} & 0.797 & 1.926 & 116M \\
    & MDNSal \cite{reddy2020tidying} & 0.899 & 0.217 & 0.868 & 0.797 & 1.893 & -  \\
    & MSI-Net \cite{kroner2020contextual} & 0.899 & 0.307 & 0.865 & 0.784 & 1.931 & 20M   \\
    & GazeGAN\cite{che2019gazegan} & 0.879 & 0.376 & 0.864 & 0.773 & 1.899 & -  \\
    & UNISAL\cite{droste2020unified} & 0.879 & 0.354 & 0.864 & 0.775 & 1.952 & 4M  \\
    & Transalnet$^\ast$ ~\cite{lou2022transalnet} & 0.89 & 0.220 & 0.867 & 0.783 & 1.924 & 72M\\
    & DeepGaze IIE$^\ast$~\cite{linardos2021deepgaze} & 0.872 & 0.285 & 0.869 & 0.733 & \textbf{1.996} & 104M  \\
    & Temp-Sal$^\ast$ \cite{aydemir2023tempsal} & \textbf{0.911} & 0.195 & 0.869 & {0.800}  & 1.967 & 242M  \\
    \cmidrule{2-8}
    & \cellcolor[HTML]{DADCFF}  \textbf{SUM (Ours)} &  \cellcolor[HTML]{DADCFF} {0.909\scriptsize{$-$0.22$\%$}} & \cellcolor[HTML]{DADCFF} \textbf{0.192\scriptsize{$-$1.54$\%$}} & \cellcolor[HTML]{DADCFF} \textbf{0.876 \scriptsize{$+$0.64$\%$}} & \cellcolor[HTML]{DADCFF} \textbf{0.804 \scriptsize{$+$0.50$\%$}}  & \cellcolor[HTML]{DADCFF} {1.981 \scriptsize{$-$0.75$\%$}}  & \cellcolor[HTML]{DADCFF} 57.5M  \\
    \midrule
    
    \textit{CAT2000}~\cite{borji2015cat2000} & FastSal \cite{hu2021fastsal} & 0.721 & 0.552 & 0.86 & 0.603  & 1.859 & 4M \\
    (Natural scene) & SAM-Resnet \cite{cornia2018predicting} & 0.87 & 0.670 & 0.878 & 0.739  & 2.411 & 43M \\
    & MSI-Net$^\ast$ \cite{kroner2020contextual} & 0.866 & 0.428 & 0.881 & 0.730 & 2.355 & 20M  \\
    & DVA \cite{wang2017deep} & 0.861 & 0.449 & 0.878 & 0.734 & 2.345 & -  \\
    & UNISAL \cite{droste2020unified} & 0.842 & 0.530 & 0.876 & 0.721  & 2.257 & 4M \\
    & MDNSal \cite{reddy2020tidying} & \textbf{0.889} & 0.293 & 0.878 & 0.751 & 2.329 & -  \\
    & Transalnet$^\ast$ \cite{lou2022transalnet} & 0.877 & 0.287 & 0.882 & 0.744  & 2.373 & 72M \\
    \cmidrule{2-8}
    & \cellcolor[HTML]{DADCFF} \textbf{SUM (Ours)} &  \cellcolor[HTML]{DADCFF} {0.882\scriptsize{$-$0.79$\%$}} & \cellcolor[HTML]{DADCFF} \textbf{0.270 \scriptsize{$-$5.92$\%$}} & \cellcolor[HTML]{DADCFF} \textbf{0.888 \scriptsize{$+$0.68$\%$}} & \cellcolor[HTML]{DADCFF} \textbf{0.754 \scriptsize{$+$0.4$\%$}}  & \cellcolor[HTML]{DADCFF} \textbf{2.424 \scriptsize{$+$0.54$\%$}}  & \cellcolor[HTML]{DADCFF} 57.5M  \\
    \midrule
    
    \textit{MIT1003}~\cite{judd2009learning} & FastSal \cite{hu2021fastsal} & 0.590 & 1.036 & 0.875 & 0.478  & 2.008 & 4M \\
    (Natural scene) & SAM-Resnet \cite{cornia2018predicting} & {0.746} & 1.247 & 0.902 & {0.597}  & 2.752 & 43M \\
    & DVA \cite{wang2017deep} & 0.699 & 0.753 & 0.897 & 0.566 & 2.574 & -  \\
    & UNISAL \cite{droste2020unified} & 0.734 & 1.014 & 0.902 & {0.597}  & {2.759} & 4M \\
    & Transalnet$^\ast$ \cite{lou2022transalnet} & 0.722 & 0.660 & 0.903 & 0.592 & 2.631  & 72M\\
    \cmidrule{2-8}
    & \cellcolor[HTML]{DADCFF} \textbf{SUM (Ours)} &  \cellcolor[HTML]{DADCFF} \textbf{0.768\scriptsize{$+$2.95$\%$}} & \cellcolor[HTML]{DADCFF} \textbf{0.563\scriptsize{$-$14.7$\%$}} & \cellcolor[HTML]{DADCFF} \textbf{0.913 \scriptsize{$+$1.11$\%$}} & \cellcolor[HTML]{DADCFF} \textbf{0.630 \scriptsize{$+$5.53$\%$}}  & \cellcolor[HTML]{DADCFF} \textbf{2.839 \scriptsize{$+$2.9$\%$}}  & \cellcolor[HTML]{DADCFF} 57.5M  \\
    \bottomrule
    \end{tabular}
}
\end{table*}

\begin{figure*}[htb]
    \centering
    \resizebox{0.95\textwidth}{!}{
    \begin{tabular}{@{} *{7}c @{}}
    \includegraphics[width=0.18\textwidth]{sec/figures/comparison/class0_naturalscene_mouse/COCO_val2014_000000428218.jpg} &
    \includegraphics[width=0.18\textwidth]{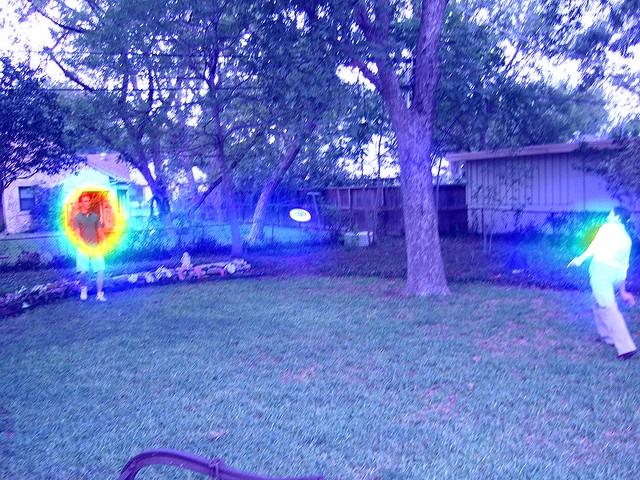} &
    \includegraphics[width=0.18\textwidth]{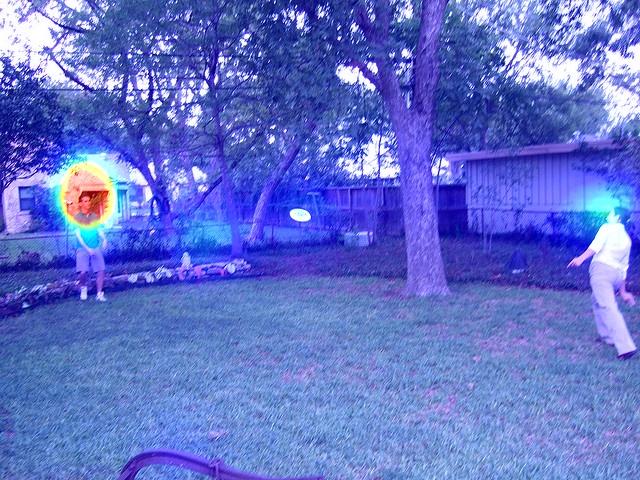} &
    \includegraphics[width=0.18\textwidth]{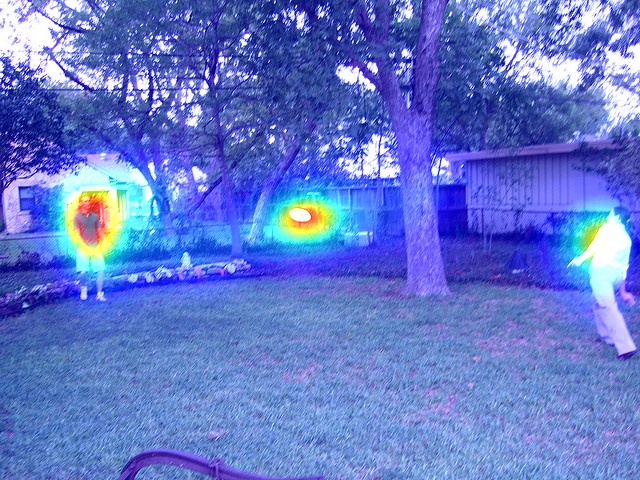} &
    \includegraphics[width=0.18\textwidth]{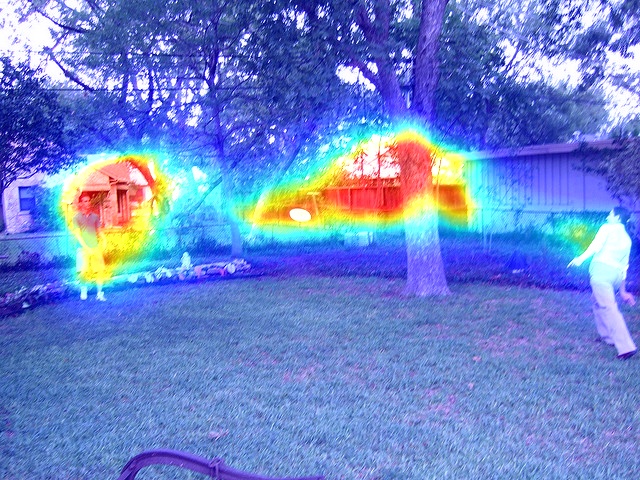} &
    \includegraphics[width=0.18\textwidth]{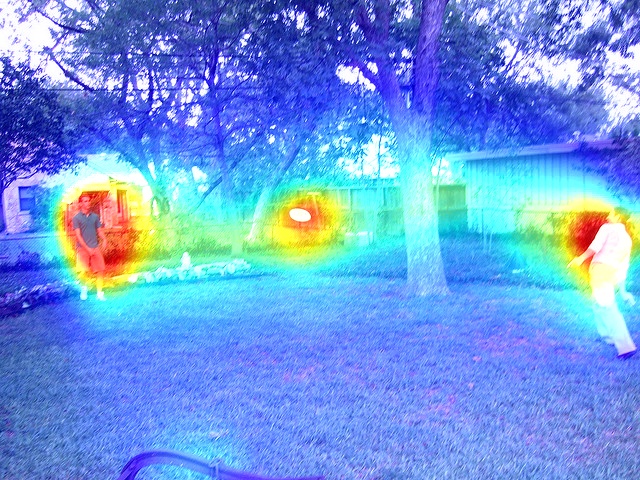} &
    \includegraphics[width=0.18\textwidth]{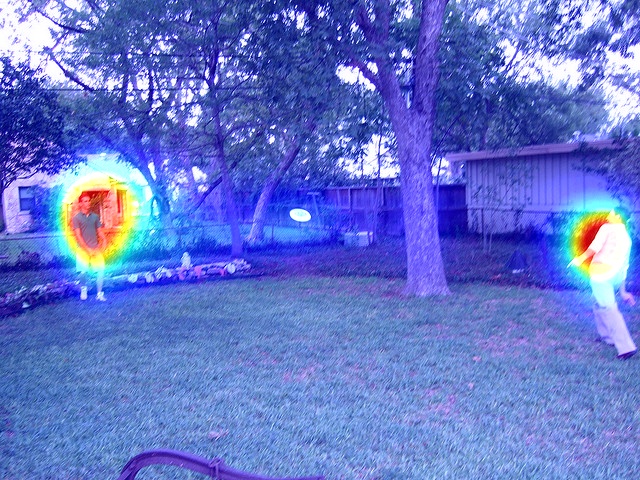} 
    \\
    { \large Input Image} & {\large Ground Truth} & {\large \textbf{SUM (Ours)}} & {\large Transalnet~\cite{lou2022transalnet}} & {\large Temp-Sal~\cite{aydemir2023tempsal}}  &  {\large DeepGaze IIE~\cite{linardos2021deepgaze}} &  {\large SimpleNet~\cite{reddy2020tidying}} \\
    \includegraphics[width=0.18\textwidth]{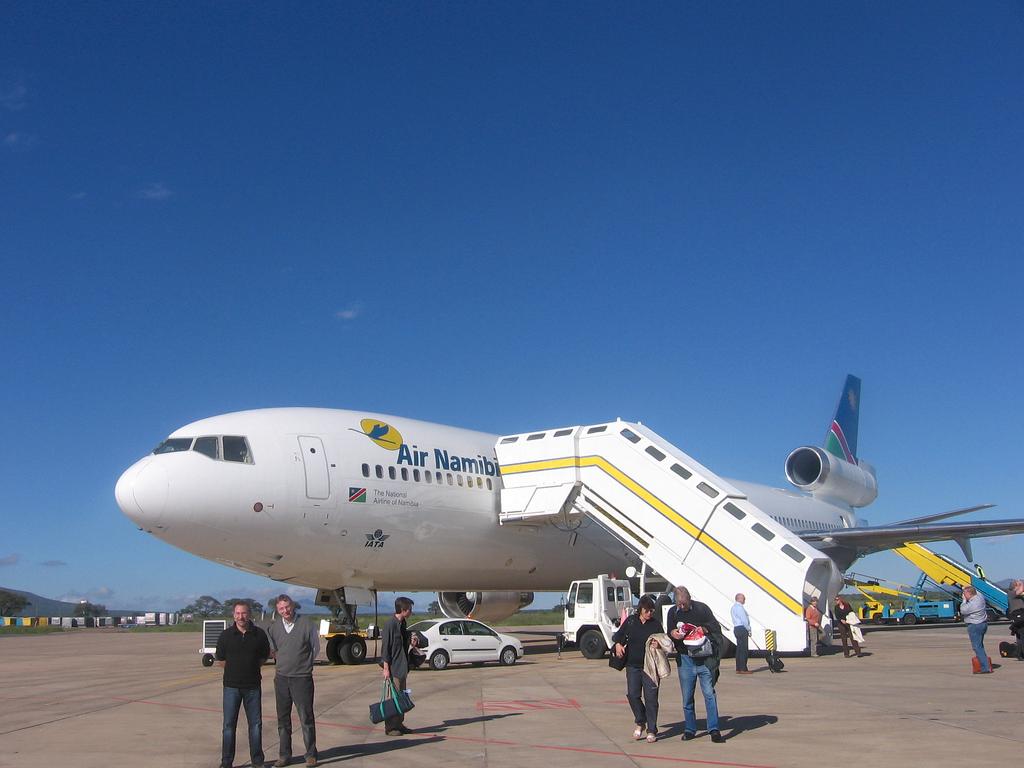} &
    \includegraphics[width=0.18\textwidth]{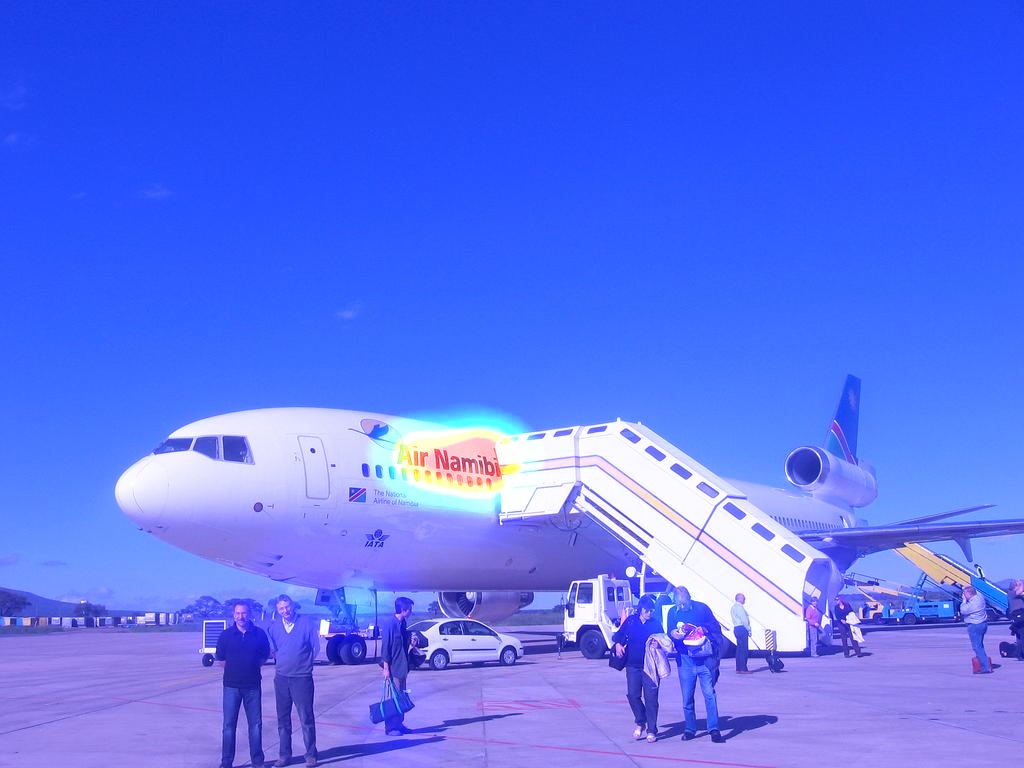} &
    \includegraphics[width=0.18\textwidth]{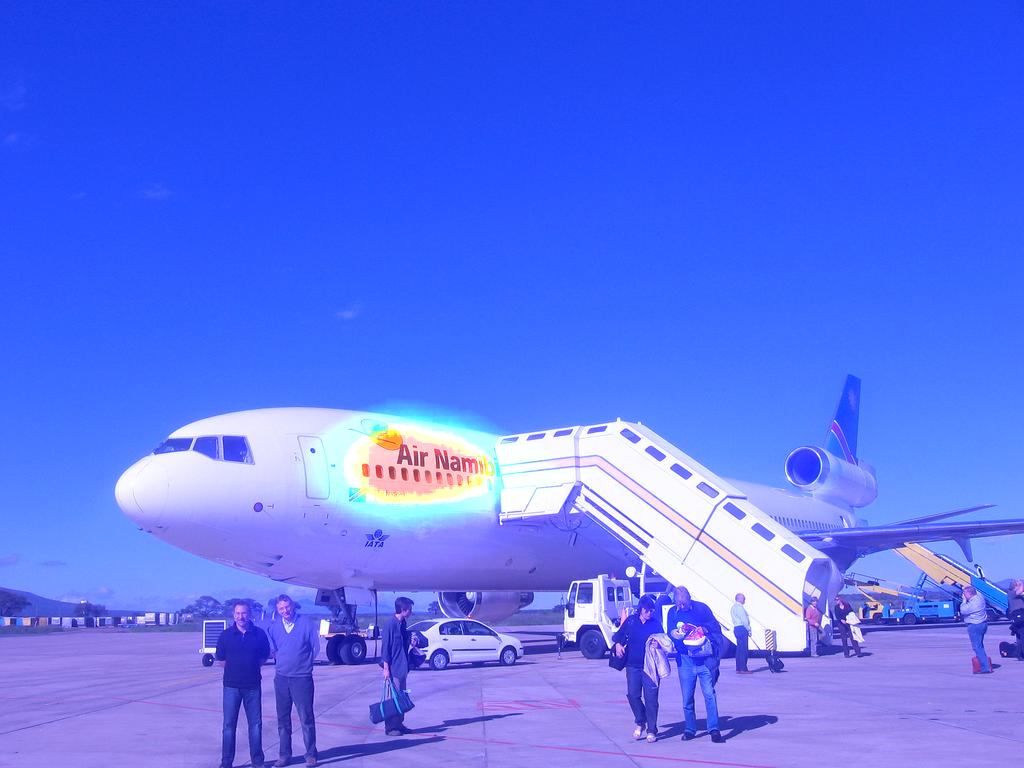} &
    \includegraphics[width=0.18\textwidth]{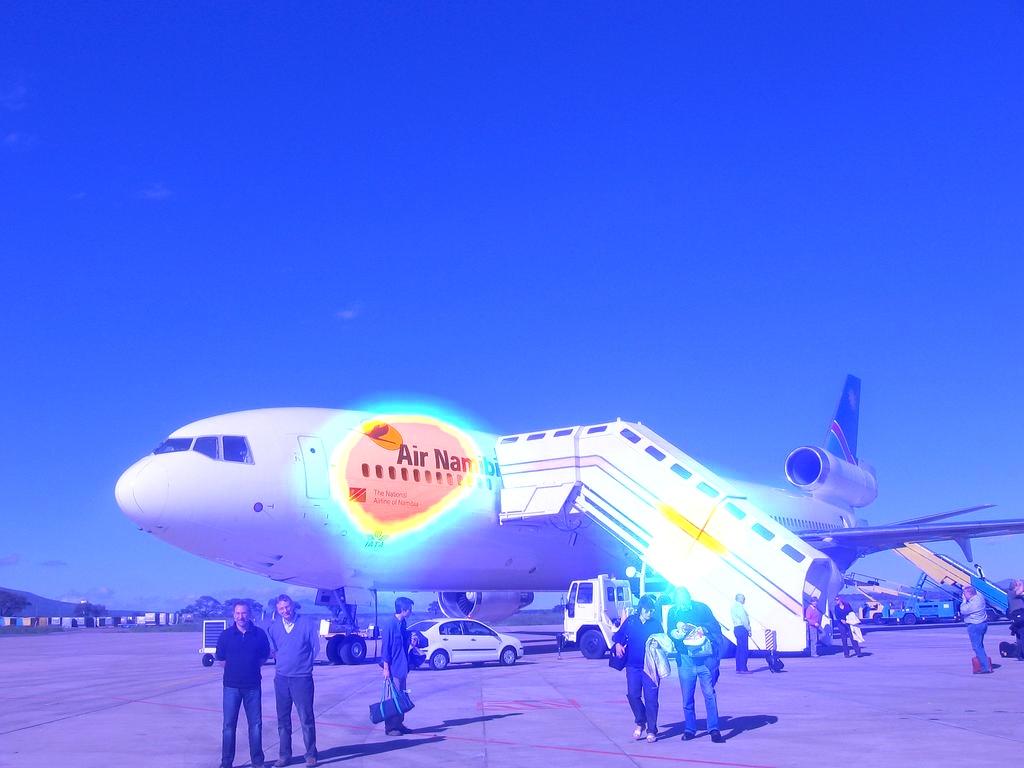} &
    \includegraphics[width=0.18\textwidth]{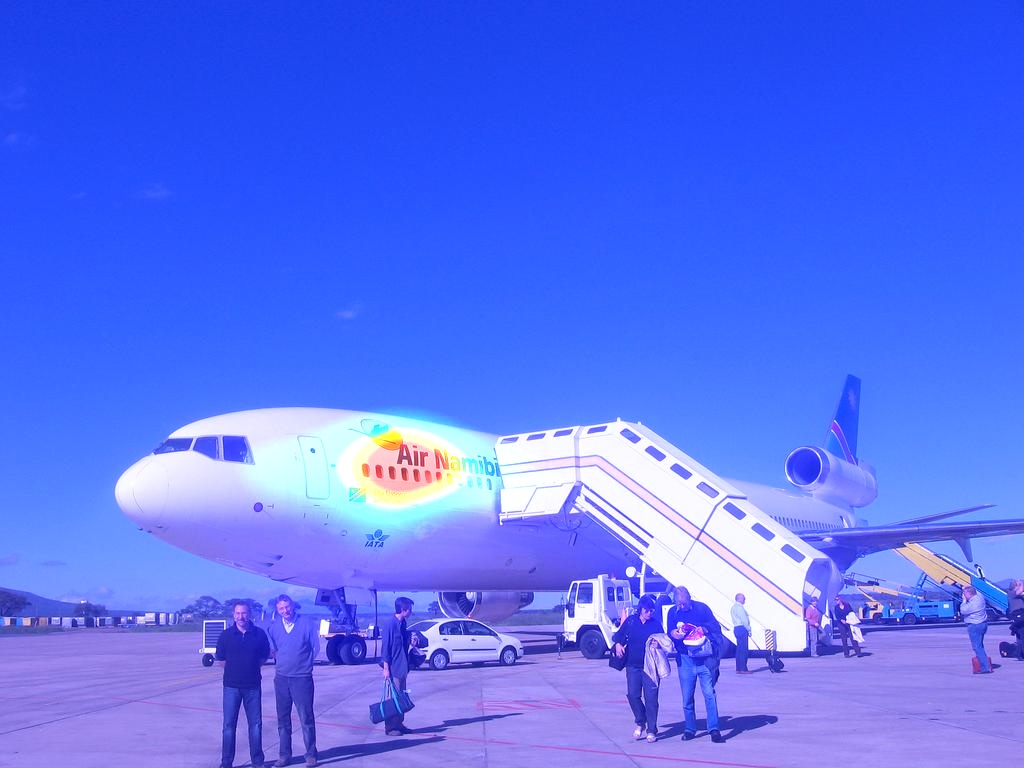} &
    \includegraphics[width=0.18\textwidth]{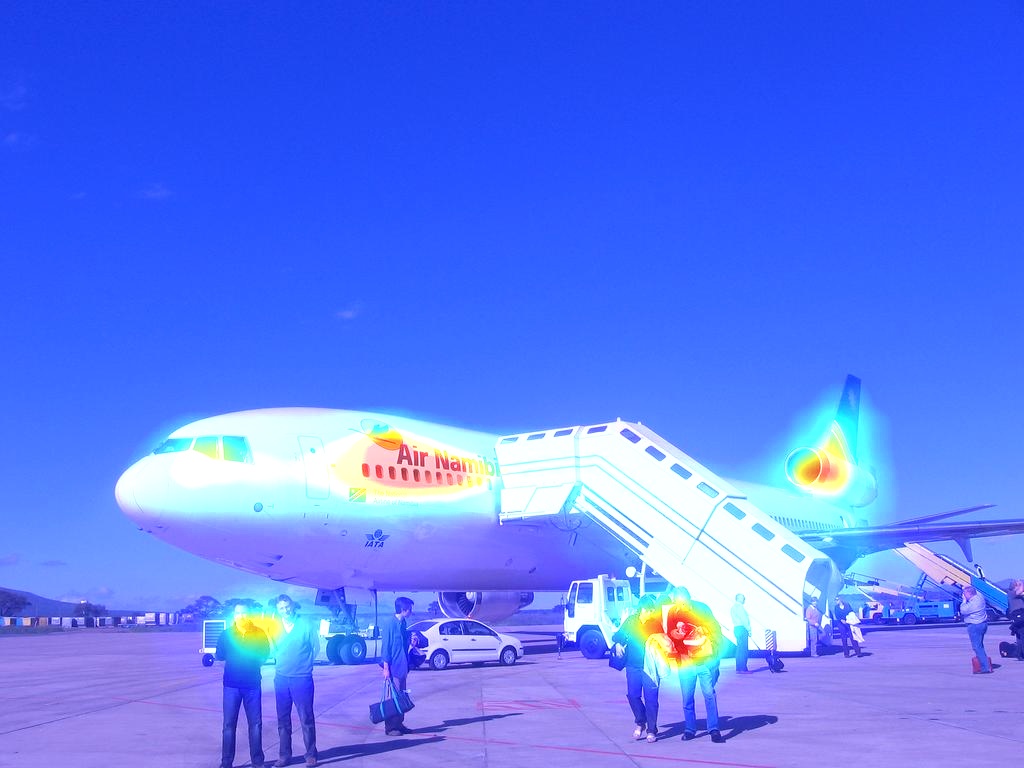} &
    \includegraphics[width=0.18\textwidth]{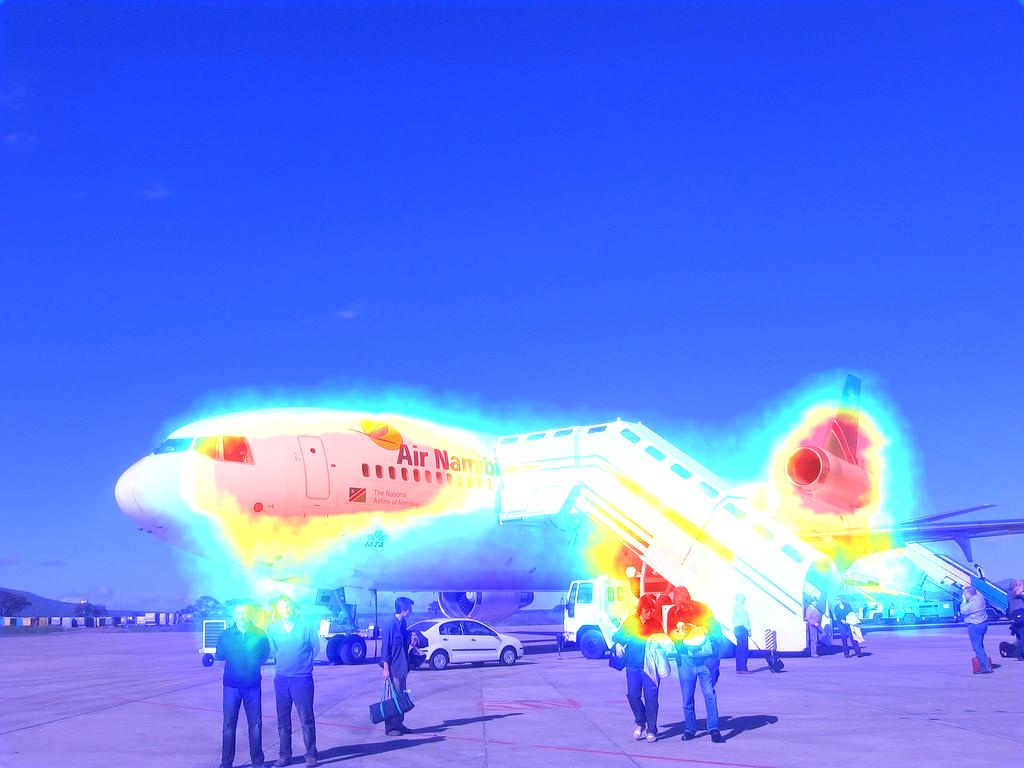} 
    \\
    { \large Input Image} & {\large Ground Truth} & {\large \textbf{SUM (Ours)}} & {\large Transalnet~\cite{lou2022transalnet}} & {\large UNISAL~\cite{droste2020unified}} & {\large EML-NET~\cite{jia2020eml}} & {\large FastSal~\cite{hu2021fastsal}} \\
    \includegraphics[width=0.18\textwidth]{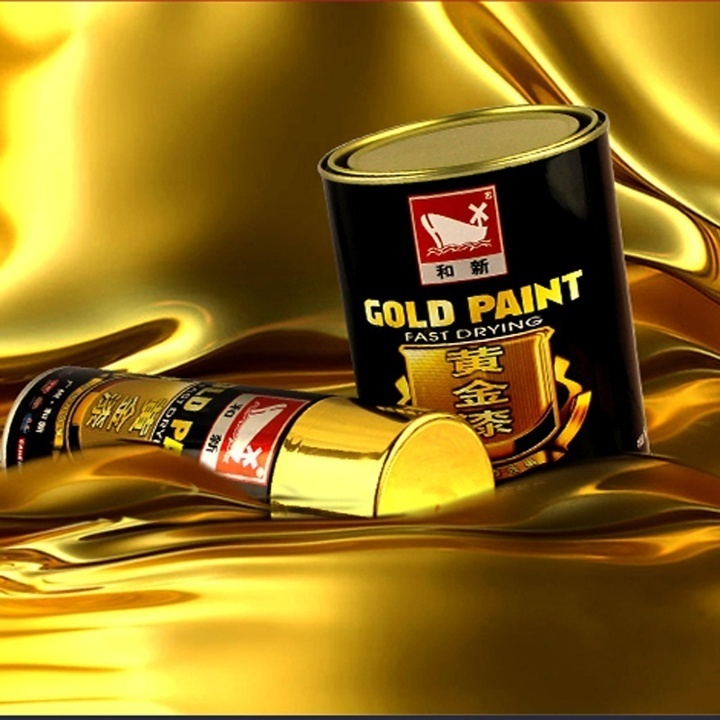} &
    \includegraphics[width=0.18\textwidth]{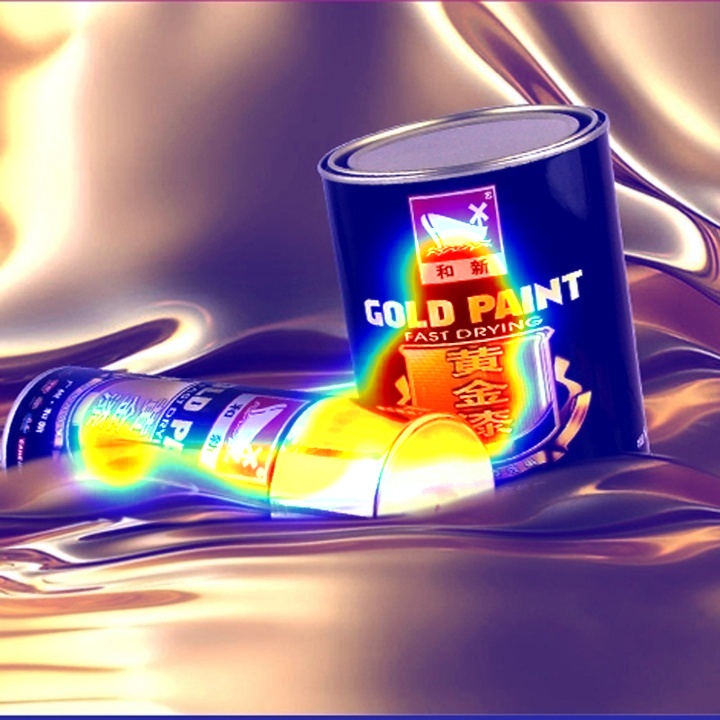} &
    \includegraphics[width=0.18\textwidth]{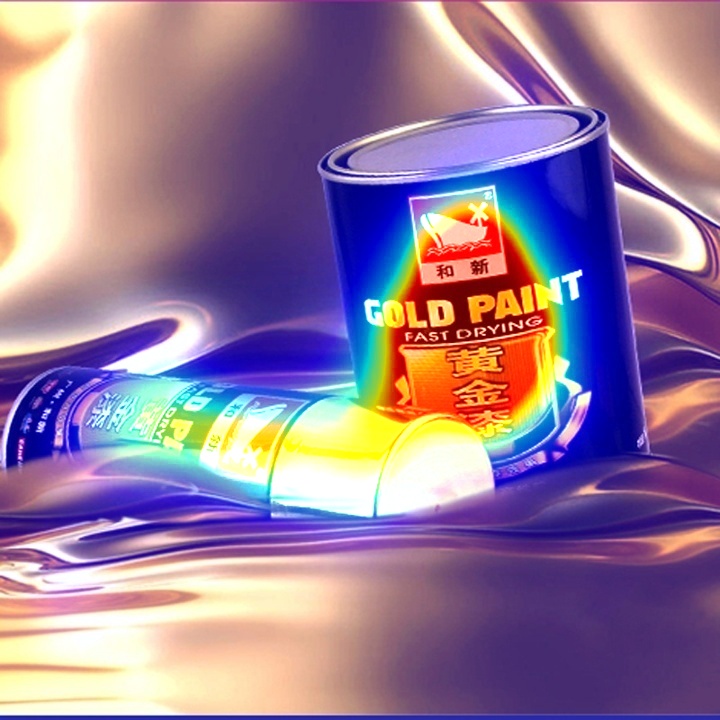} &
    \includegraphics[width=0.18\textwidth]{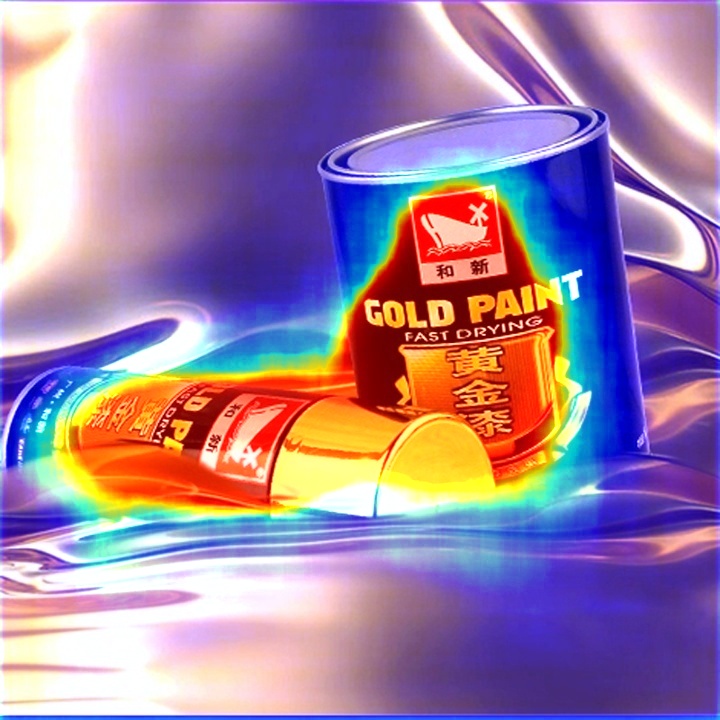} &
    \includegraphics[width=0.18\textwidth]{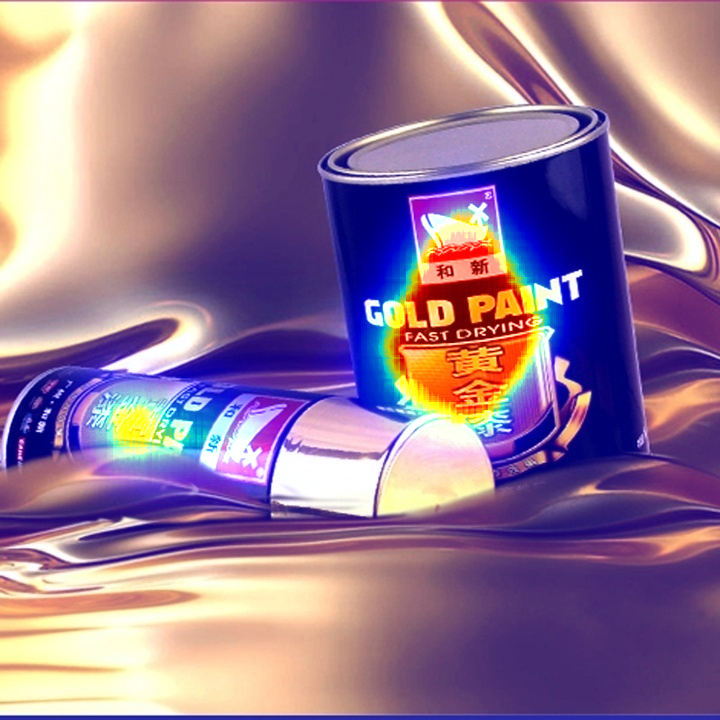} &
    \includegraphics[width=0.18\textwidth]{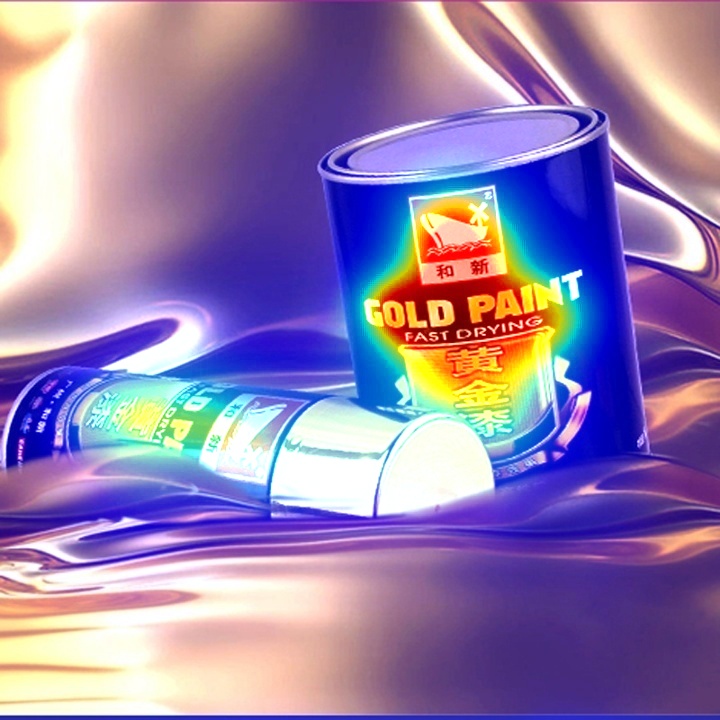} & 
    \includegraphics[width=0.18\textwidth]{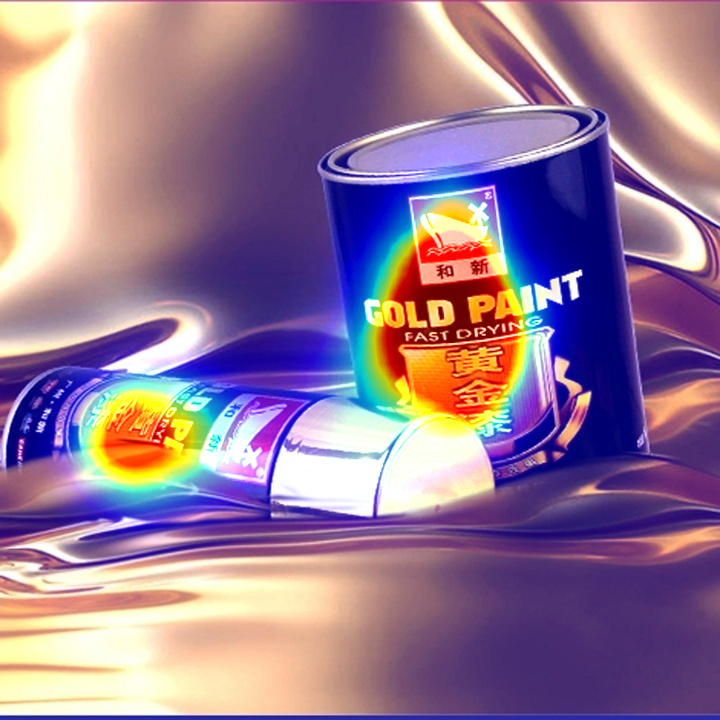} 
    \\
    { \large Input Image} & {\large Ground Truth} & {\large \textbf{SUM (Ours)}} & {\large Transalnet~\cite{lou2022transalnet}} & {\large Temp-Sal~\cite{aydemir2023tempsal}} & {\large DeepGaze IIE~\cite{linardos2021deepgaze}} & {\large \citet{hosseini2024brand}} \\
    \includegraphics[width=0.18\textwidth]{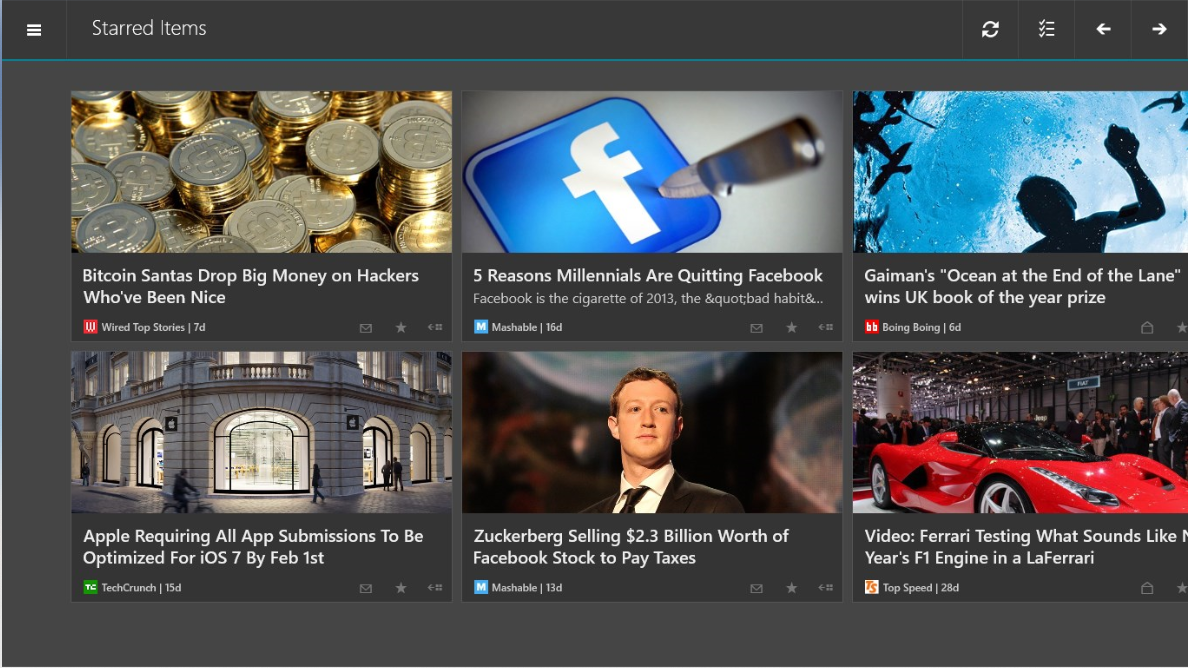} &
    \includegraphics[width=0.18\textwidth]{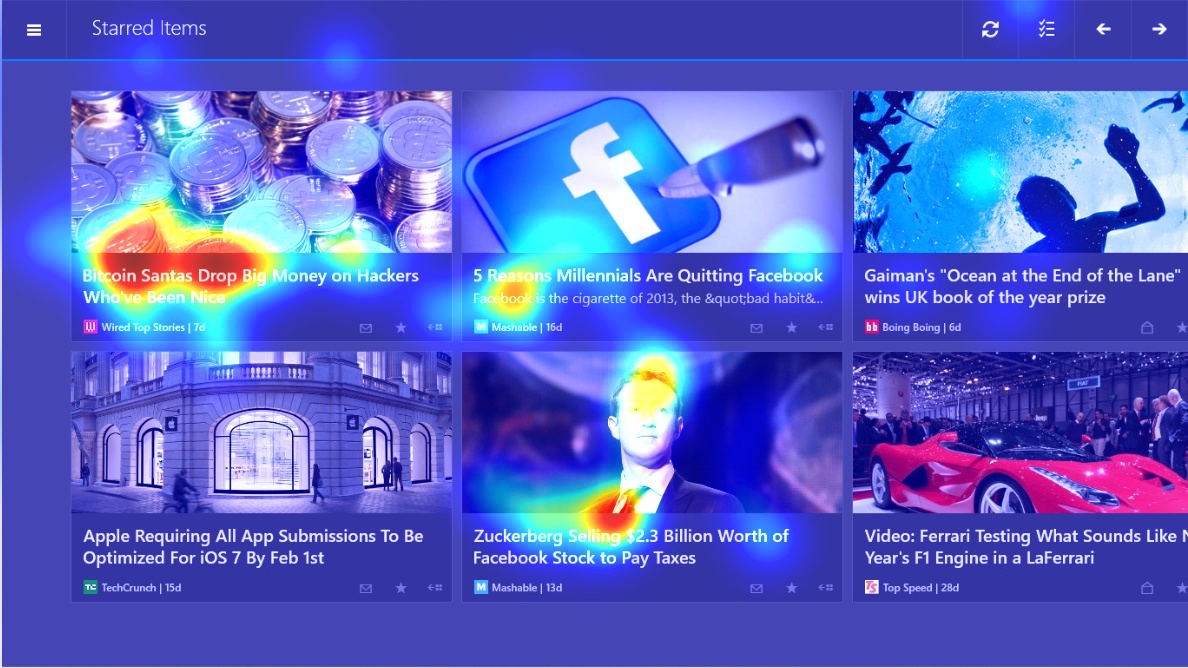} &
    \includegraphics[width=0.18\textwidth]{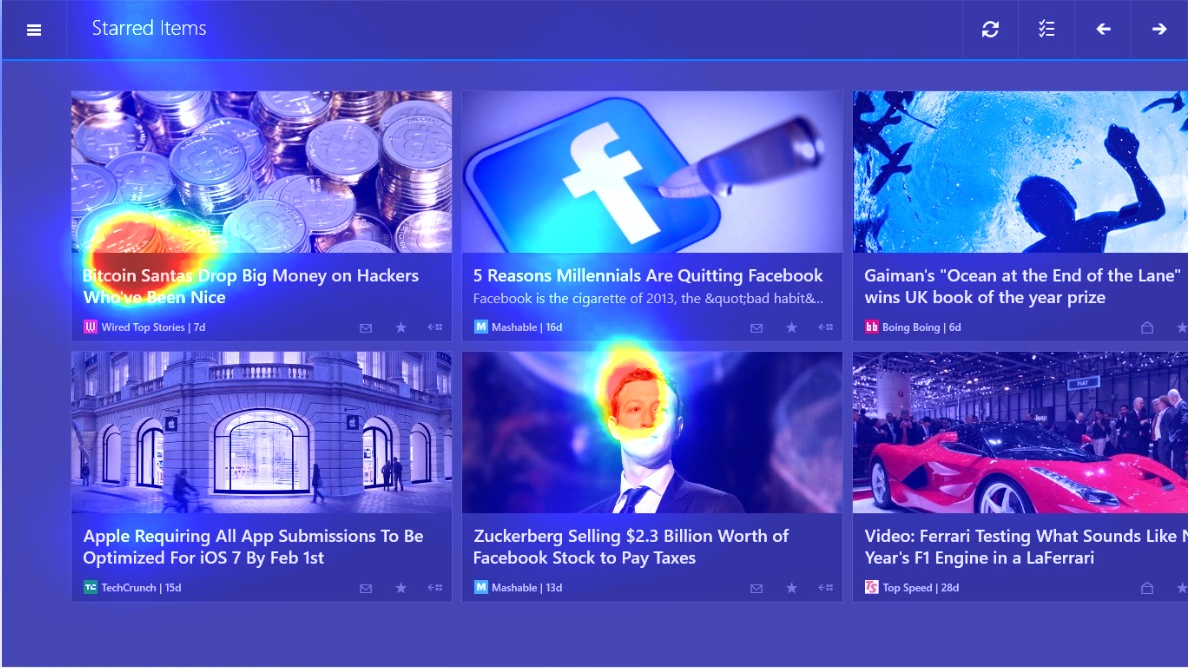} &
    \includegraphics[width=0.18\textwidth]{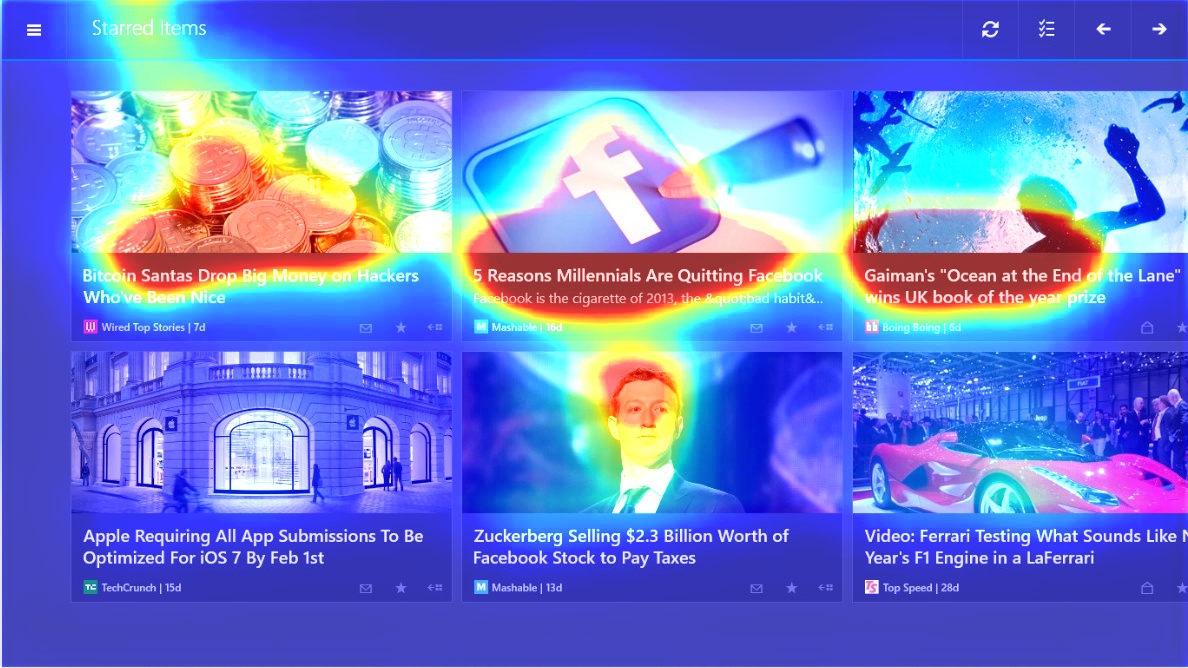} &
    \includegraphics[width=0.18\textwidth]{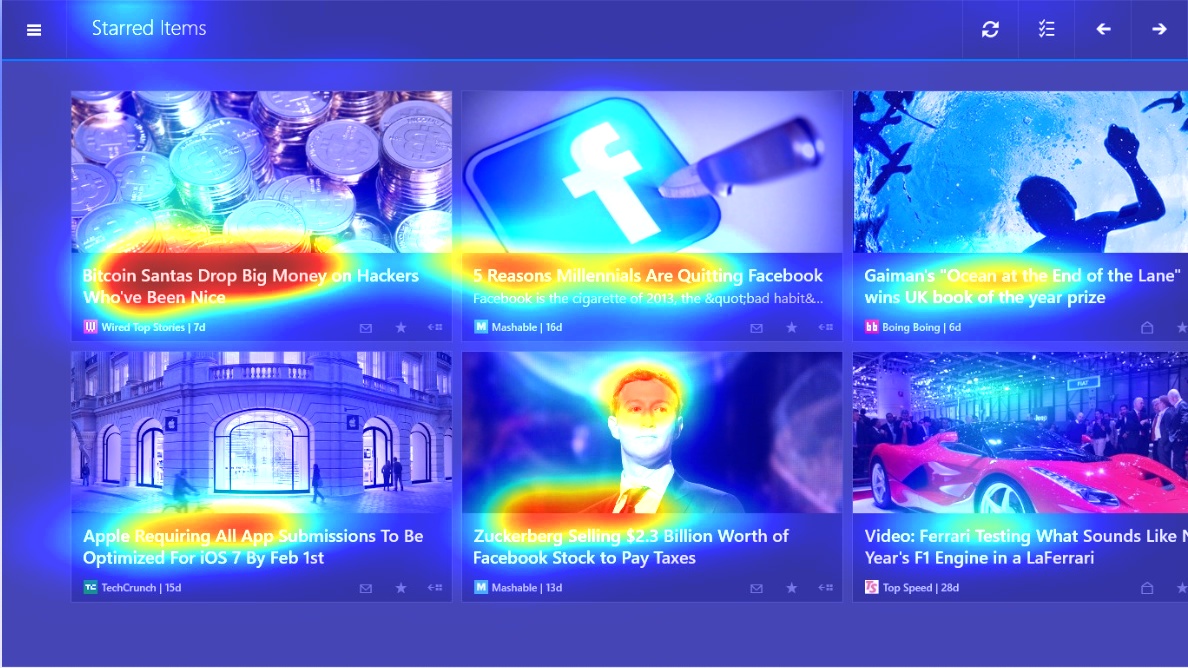} &
    \includegraphics[width=0.18\textwidth]{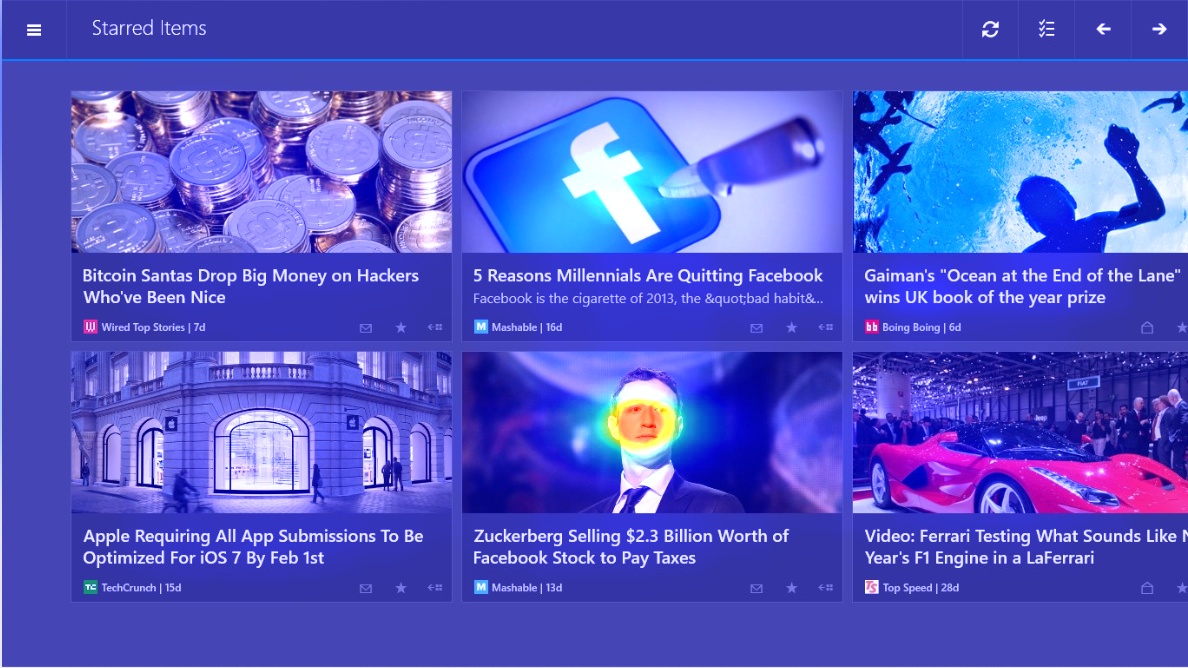} &
    \includegraphics[width=0.18\textwidth]{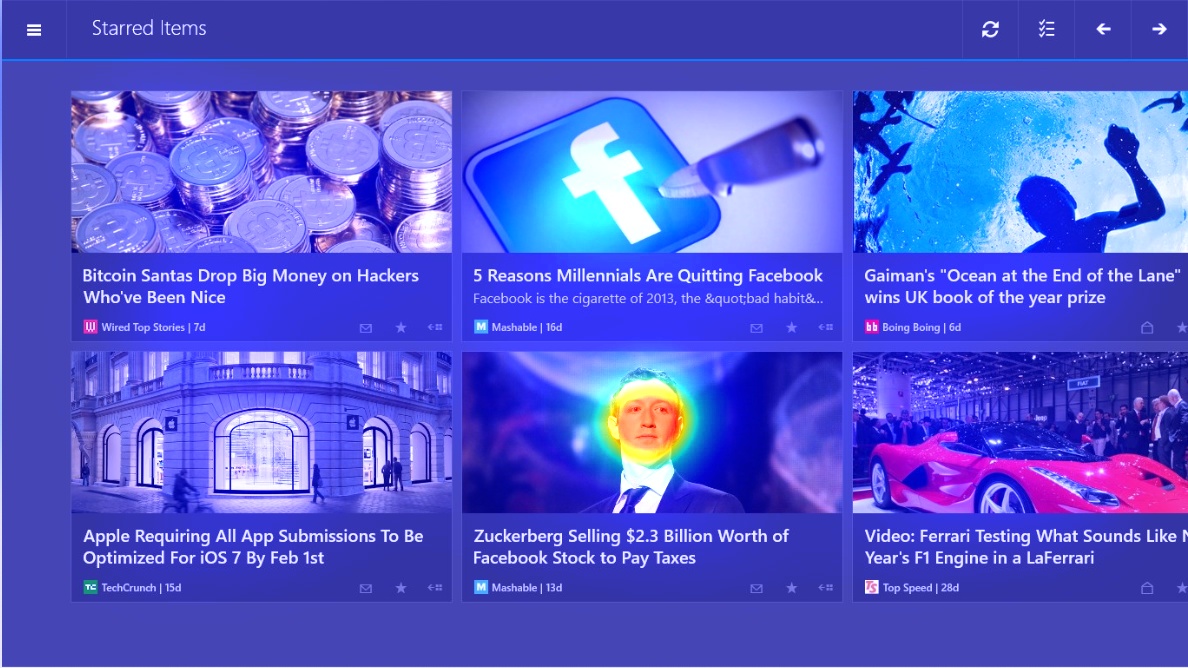} 
    \\
    { \large Input Image} & {\large Ground Truth} & {\large \textbf{SUM (Ours)}} & {\large Transalnet~\cite{lou2022transalnet}} &  {\large UMSI++~\cite{jiang2023ueyes}} & {\large SAM++~\cite{jiang2023ueyes}} & {\large UMSI~\cite{fosco2020predicting}}
    \end{tabular}
}
    \caption{Comparative visualizations of saliency predictions across different data types. The first row depicts Natural Scene-Mouse data, the second row showcases Natural Scene-Eye data, the third row features E-commerce, and the fourth row displays UI. Each row highlights the model's performance in identifying salient features within these distinct categories.}
    \label{fig:visual_sal}
\end{figure*}

\subsection{Experiment Results}
We conducted comprehensive testing of our universal model, SUM, across six different datasets, each benchmarked against state-of-the-art (SOTA) models for comparison. These datasets encompass a range of areas including natural scenes, user interfaces, and e-commerce. SUM consistently outperformed the best existing models across all datasets. In the 30 metrics presented in ~\autoref{tab:saliency}, SUM achieved SOTA results in 27 cases and secured second place in the other three. These results demonstrate that our model is highly effective and versatile across various types of data, setting a new standard for future advancements in the field. This consistency in performance underscores its robustness and capability to handle the diverse challenges presented by different datasets. Moreover, compared to counterparts like Transalnet \cite{lou2022transalnet}, Temp-Sal \cite{aydemir2023tempsal}, DeepGaze IIE \cite{linardos2021deepgaze}, and UniAR \cite{li2023uniar}, our model is relatively efficient. This efficiency underscores the advantages of our streamlined approach, which leverages Mamba's capabilities to develop a model that is efficient, robust, and universally applicable. Additionally, ~\autoref{fig:visual_sal} displays saliency prediction images selected from the validation sets, showing that our model's predictions are much closer to the ground truth than those of the SOTA models, further proving that SUM can more accurately predict human attention behavior.

\section{Ablation Study}
\textbf{Impact of different loss combinations :} We investigated the impact of different loss metric combinations on the model validation performance, as summarized in \autoref{tab:different_loss}. Our approach involved normalizing each metric using a min-max scaling technique to ensure a balanced evaluation across different metrics. The score function, described in~\autoref{eq:opt_func}, is specifically designed to maximize the beneficial metrics (CC, SIM, NSS) and minimize the detrimental metric (KL). The function's configuration is as follows:
\begin{equation}
\mathcal{F}_\text{score} = CC_{\text{scaled}} + SIM_{\text{scaled}} + NSS_{\text{scaled}} - KL_{\text{scaled}}
\label{eq:opt_func}
\end{equation}

From the results in \autoref{tab:different_loss}, it is evident that the inclusion of KL loss significantly impacts the model's performance, demonstrating its crucial role in defining saliency loss. When loss functions are used individually, the performance varies, with SIM typically showing higher values for both CC and $\mathcal{F}$ scores, indicating its strong standalone impact on model saliency. Excluding MSE, which is less directly related to saliency, still results in high performance, but the highest scores are consistently observed when MSE is included, suggesting its underlying contribution to model robustness and generalization. The integration of all five loss functions results in the highest $\mathcal{F}$ scores. This combination not only balances the enhancement and suppression of features but also stabilizes the training process, as indicated by the highest scores of 2.853 and 2.836 for Salicon and all datasets, respectively.

\begin{table*}[ht]
\centering
\caption{Evaluation of different combinations of loss functions on model performance.}
\label{tab:different_loss}
\resizebox{\textwidth}{!}{
    \begin{tabular}{ccccc|cccc|c|cccc|c}
    \multicolumn{5}{c|}{Loss Functions} & \multicolumn{5}{c|}{Avg. Performance on Salicon \cite{jiang2015salicon}} & \multicolumn{5}{c}{Avg. Performance Across All Datasets} \\
    \hline 
    KL & CC & SIM & NSS & MSE & CC $\uparrow$ & KLD $\downarrow$ & NSS $\uparrow$ & SIM  $\uparrow$ &$\mathcal{F}_\text{Score}$  $\uparrow$ & CC $\uparrow$ & KLD $\downarrow$ & NSS $\uparrow$ & SIM $\uparrow$ & $\mathcal{F}_\text{Score}$ $\uparrow$ \\
    \hline  
    \checkmark &  \xmark &  \xmark & \xmark & \xmark & 0.910 & 0.189 & 1.908 & 0.805 & 2.797 & 0.85 & 0.465 & 2.498 & 0.723 & 2.386 \\
     \xmark & \checkmark &  \xmark & \xmark  & \xmark & 0.907 & 0.732 & 1.926 & 0.787 & 1.634 & 0.851 & 1.08  & 2.532 & 0.7 & 1.218  \\
     \xmark &  \xmark & \checkmark & \xmark  &  \xmark & \textbf{0.911} & 0.447 & 1.91  & \textbf{0.807} & 2.391 & 0.85 & 0.747 & 2.469 & \textbf{0.728} & 1.917\\
     \xmark &  \xmark &  \xmark & \checkmark & \xmark  & 0.834 & 0.765 & \textbf{2.044} & 0.721 & 0 & 0.804 & 1.072 & 2.614 & 0.658 & -0.079 \\
    \xmark &  \xmark & \xmark  &  \xmark & \checkmark & 0.909 & 0.234 & 1.919 & 0.803 & 2.696 & 0.846 & 0.525 & 2.479 & 0.719 & 2.089 \\
    \hline   
    \checkmark & \xmark  & \checkmark &  \xmark &  \xmark & \textbf{0.911} & 0.196 & 1.928 & 0.806 & 2.833 & \textbf{0.852} & 0.465 & 2.337 & \textbf{0.728} & 1.972 \\
    \checkmark & \xmark  & \xmark  & \checkmark &  \xmark & 0.892 & 0.199 & 2.029 & 0.792 & 2.537 & 0.841 & 0.467 & 2.594 & 0.712 & 2.353\\
    \checkmark & \checkmark & \xmark  &  \xmark &  \xmark & \textbf{0.911} & \textbf{0.185} & 1.191 & 0.805 & 1.977 & \textbf{0.852} & 0.453 & 2.515 & 0.720 & 2.46 \\
    \checkmark & \xmark  & \xmark  & \xmark  & \checkmark & 0.909 & 0.192 & 1.917 & 0.802 & 2.755 & 0.851 & 0.456 & 2.504 & 0.723 & 2.441 \\
     \xmark & \checkmark & \checkmark &  \xmark &  \xmark & 0.910 & 0.531 & 1.921 & 0.802 & 2.188 & 0.85  & 0.871 & 2.503 & 0.721 & 1.733 \\
    \hline 
    \checkmark & \checkmark & \checkmark & \xmark  &  \xmark & 0.909 & 0.198 & 1.920 & 0.803 & 2.759 & \textbf{0.852} & 0.464 & 2.527 & 0.726 & 2.568 \\
    \checkmark & \xmark  & \checkmark & \xmark  & \checkmark & 0.909 & 0.192 & 1.919 & 0.799 & 2.722 & \textbf{0.852} & 0.461 & 2.514 & 0.726 & 2.53 \\
    \checkmark &  \xmark & \xmark  & \checkmark & \checkmark & 0.887 & 0.208 & 2.038 & 0.788 & 2.421 & 0.830 & 0.472 & \textbf{2.642} & 0.711 & 2.259 \\
    \checkmark & \checkmark & \xmark  & \xmark  & \checkmark & 0.910 & 0.188 & 1.914 & 0.803 & 2.783 & 0.851 & \textbf{0.447} & 2.511 & 0.722 & 2.464 \\
    \hline   
    \checkmark & \checkmark & \checkmark & \checkmark & \xmark  & 0.907 & 0.198 & 1.989 & 0.803 & 2.815 & 0.850 & 0.466 & 2.614 & 0.725 & 2.794 \\
    \checkmark & \checkmark & \checkmark & \xmark  & \checkmark & 0.905 & 0.208 & 1.920 & 0.798 & 2.632 & \textbf{0.852} & 0.457 & 2.510 & 0.720 & 2.437 \\
    \hline   
    \checkmark & \checkmark & \checkmark & \checkmark & \checkmark & 0.909 & 0.192 & 1.981 & 0.804 & \textbf{2.853} &  \textbf{0.852} & 0.450 & 2.602 & 0.726 & \textbf{2.836}
    \end{tabular}
}
\end{table*}

\noindent\textbf{Impact of the C-VSS module:} We compared the impact of using a C-VSS module conditioned on three and four classes against a standard VSS, which serves as the unconditional setup. The three classes include Natural Scene, UI, and E-Commerce, in contrast to our model's broader categorization into four classes. As shown in \autoref{tab:compare_classes}, the C-VSS module significantly enhances performance across all evaluated datasets compared to the standard VSS. Notably, conditioning the model on four classes yields better results than limiting it to three. This suggests that the finer categorization in the four-class setup better aligns with the varied data characteristics, especially when dealing with different data acquisition setups, thereby improving the model's predictive accuracy and robustness.

\begin{table*}[t]
\caption{Mean value and standard deviation of saliency prediction performance comparison of conditional VSS modules for three and four classes and standard VSS (no-condition) across all datasets.}
\centering
\resizebox{0.9\textwidth}{!}{
\begin{tabular}{c|l|ccccccc}
\toprule
Dataset & Method & CC $\uparrow$ & KLD $\downarrow$ & AUC $\uparrow$ & SIM $\uparrow$ & NSS $\uparrow$ \\
\midrule
\textit{U-EYE}~\cite{jiang2023ueyes} & No-condition & 0.725 \scriptsize{$\pm$ 0.035} & 0.562 \scriptsize{$\pm$ 0.062} & 0.845 \scriptsize{$\pm$ 0.012} & 0.626 \scriptsize{$\pm$ 0.023}  & 1.689 \scriptsize{$\pm$ 0.121}  \\
(Web page) & 3-class & 0.729 \scriptsize{$\pm$ 0.035} & 0.551 \scriptsize{$\pm$ 0.057} & 0.845 \scriptsize{$\pm$ 0.012} & 0.628 \scriptsize{$\pm$ 0.012}  & 1.699 \scriptsize{$\pm$ 0.012}   \\
& \cellcolor[HTML]{DADCFF} 4-class & \cellcolor[HTML]{DADCFF} \textbf{0.731\scriptsize{$\pm$ 0.037}} & \cellcolor[HTML]{DADCFF} \textbf{0.544 \scriptsize{$\pm$ 0.057}} & \cellcolor[HTML]{DADCFF} \textbf{0.846 \scriptsize{$\pm$ 0.012}} & \cellcolor[HTML]{DADCFF} \textbf{0.630 \scriptsize{$\pm$ 0.023}}  & \cellcolor[HTML]{DADCFF} \textbf{1.704 \scriptsize{$\pm$ 0.125}}  \\
\midrule
\textit{SalECI}~\cite{jiang2022does} & No-condition &  0.783\scriptsize{$\pm$ 0.046} & 0.502 \scriptsize{$\pm$ 0.112} & 0.898 \scriptsize{$\pm$ 0.014} & 0.677 \scriptsize{$\pm$ 0.039} & \textbf{2.017} \scriptsize{$\pm$ 0.168}  \\
(E-Commercial) & 3-class & 0.781 \scriptsize{$\pm$ 0.055} & 0.505 \scriptsize{$\pm$ 0.131} & 0.896 \scriptsize{$\pm$ 0.016} & 0.678 \scriptsize{$\pm$ 0.047}  & 1.99 \scriptsize{$\pm$ 0.181}   \\
& \cellcolor[HTML]{DADCFF} 4-class & \cellcolor[HTML]{DADCFF} \textbf{0.789\scriptsize{$\pm$ 0.0453}} & \cellcolor[HTML]{DADCFF} \textbf{0.473 \scriptsize{$\pm$ 0.088}} & \cellcolor[HTML]{DADCFF} \textbf{0.899 \scriptsize{$\pm$ 0.012}} & \cellcolor[HTML]{DADCFF} \textbf{0.680 \scriptsize{$\pm$ 0.041}}  & \cellcolor[HTML]{DADCFF} 2.012 \scriptsize{$\pm$ 0.161}  \\
\midrule
\textit{OSIE}~\cite{xu2014predicting} & No-condition &  0.842\scriptsize{$\pm$ 0.033} & 0.403 \scriptsize{$\pm$ 0.05} & 0.918 \scriptsize{$\pm$ 0.009} & 0.703 \scriptsize{$\pm$ 0.022}  & 3.18 \scriptsize{$\pm$ 0.32} \\
(Natural scene) & 3-class & 0.845 \scriptsize{$\pm$ 0.033} & 0.395 \scriptsize{$\pm$ 0.049} & 0.918 \scriptsize{$\pm$ 0.009} & 0.706 \scriptsize{$\pm$ 0.022}  & 3.213 \scriptsize{$\pm$ 0.323}   \\
& \cellcolor[HTML]{DADCFF} 4-class & \cellcolor[HTML]{DADCFF} \textbf{0.861\scriptsize{$\pm$ 0.029}} & \cellcolor[HTML]{DADCFF} \textbf{0.340\scriptsize{$\pm$ 0.050}} & \cellcolor[HTML]{DADCFF} \textbf{0.924\scriptsize{$\pm$ 0.008}} & \cellcolor[HTML]{DADCFF} \textbf{0.727 \scriptsize{$\pm$ 0.02}}  & \cellcolor[HTML]{DADCFF} \textbf{3.416 \scriptsize{$\pm$ 0.319}}  \\
\midrule
\textit{Salicon}~\cite{jiang2015salicon} & No-condition &  0.903\scriptsize{$\pm$ 0.012} & 0.206 \scriptsize{$\pm$ 0.028} & 0.875 \scriptsize{$\pm$ 0.014} & 0.798 \scriptsize{$\pm$ 0.013}  & 1.979 \scriptsize{$\pm$ 0.203}  \\
(Natural scene) & 3-class & 0.904 \scriptsize{$\pm$ 0.011} & 0.205 \scriptsize{$\pm$ 0.027} & 0.875 \scriptsize{$\pm$ 0.014} & 0.798 \scriptsize{$\pm$ 0.013}  & 1.981 \scriptsize{$\pm$ 0.204}   \\
& \cellcolor[HTML]{DADCFF} 4-class & \cellcolor[HTML]{DADCFF} \textbf{0.909\scriptsize{$\pm$ 0.011}} & \cellcolor[HTML]{DADCFF} \textbf{0.192\scriptsize{$\pm$ 0.025}} & \cellcolor[HTML]{DADCFF} \textbf{0.876 \scriptsize{$\pm$ 0.014}} & \cellcolor[HTML]{DADCFF} \textbf{0.804 \scriptsize{$\pm$ 0.012}}  & \cellcolor[HTML]{DADCFF} \textbf{1.981 \scriptsize{$\pm$ 0.201}}    \\
\midrule
\textit{CAT2000}~\cite{borji2015cat2000} & No-condition & 0.880\scriptsize{$\pm$ 0.014} & 0.272 \scriptsize{$\pm$ 0.022} & 0.887\scriptsize{$\pm$ 0.010} & 0.752 \scriptsize{$\pm$ 0.010}  & 2.42 \scriptsize{$\pm$ 0.141} \\
(Natural scene) & 3-class & 0.881 \scriptsize{$\pm$ 0.016} & 0.271 \scriptsize{$\pm$ 0.023} & \textbf{0.888 \scriptsize{$\pm$ 0.010}} & 0.753 \scriptsize{$\pm$ 0.011}  & \textbf{2.424 \scriptsize{$\pm$ 0.142}}   \\
& \cellcolor[HTML]{DADCFF} 4-class & \cellcolor[HTML]{DADCFF} \textbf{0.882\scriptsize{$\pm$ 0.0158}} & \cellcolor[HTML]{DADCFF} \textbf{0.270 \scriptsize{$\pm$ 0.026}} & \cellcolor[HTML]{DADCFF} \textbf{0.888 \scriptsize{$\pm$ 0.010}} & \cellcolor[HTML]{DADCFF} \textbf{0.754 \scriptsize{$\pm$ 0.011}}  & \cellcolor[HTML]{DADCFF} \textbf{2.424 \scriptsize{$\pm$ 0.142}}  \\
\midrule
\textit{MIT1003}~\cite{judd2009learning} & No-condition &  0.737\scriptsize{$\pm$ 0.035} & 0.641 \scriptsize{$\pm$ 0.083} & 0.908 \scriptsize{$\pm$ 0.010} & 0.596 \scriptsize{$\pm$ 0.024}  & 2.648 \scriptsize{$\pm$ 0.255}   \\
(Natural scene) & 3-class & 0.741 \scriptsize{$\pm$ 0.034} & 0.636 \scriptsize{$\pm$ 0.077} & 0.908 \scriptsize{$\pm$ 0.010} & 0.597 \scriptsize{$\pm$ 0.023}  & 2.678\scriptsize{$\pm$ 0.249}   \\
& \cellcolor[HTML]{DADCFF} 4-class & \cellcolor[HTML]{DADCFF} \textbf{0.768\scriptsize{$\pm$ 0.039}} & \cellcolor[HTML]{DADCFF} \textbf{0.563\scriptsize{$\pm$ 0.075}} & \cellcolor[HTML]{DADCFF} \textbf{0.913 \scriptsize{$\pm$ 0.009}} & \cellcolor[HTML]{DADCFF} \textbf{0.630 \scriptsize{$\pm$ 0.027}}  & \cellcolor[HTML]{DADCFF} \textbf{2.839 \scriptsize{$\pm$ 0.285}}   \\
\bottomrule
\end{tabular}
}
\label{tab:compare_classes}
\end{table*}

\noindent\textbf{Impact of different prompt lengths:} We explored the influence of prompt length on model performance. We experimented with various prompt lengths—32, 96, 128, 256—to determine how they impact model behavior during the training and validation phases. The results, detailed in \autoref{tab:prompt_length}, indicate that both shorter and longer prompt lengths contribute to fitting issues. Among the tested lengths, 128 demonstrated the most balanced and effective outcome.

\begin{table}[]
\centering
\caption{Impact of prompt length on model performance.}
\label{tab:prompt_length}
\resizebox{0.9\linewidth}{!}{
    \begin{tabular}{l|cccc|c}
    \hline
    \multirow{2}{*}{Prompt Length} & \multicolumn{4}{c|}{Performance Metrics} & \multirow{2}{*}{\# Parameters} \\ \cline{2-5}
     & CC $\uparrow$ & KL $\downarrow$ & NSS $\uparrow$ & SIM $\uparrow$ &  \\ \hline
    \multicolumn{6}{c}{Salicon \cite{jiang2015salicon}} \\
    \hline
    64 & \textbf{0.909} & 0.196 & 1.98 & \textbf{0.804} & 57.4M \\
    96 & \textbf{0.909} & \textbf{0.188} & 1.958 & 0.802 & 57.4M \\
    \rowcolor[HTML]{DADCFF} \textbf{128} & \textbf{0.909} & 0.192 & \textbf{1.981} & \textbf{0.804} & 57.5M \\
    256 & 0.906 & 0.195 & 1.953 & 0.801 & 58M \\
    \hline
    \multicolumn{6}{c}{Average Performance Across Datasets} \\
    \hline
    64 & 0.849 & 0.463 & 2.601 & 0.725 & 57.4M \\
    96 & 0.847 & 0.455 & 2.567 & 0.722 & 57.4M \\
    \rowcolor[HTML]{DADCFF} \textbf{128} & \textbf{0.852} & \textbf{0.450} & \textbf{2.602} & \textbf{0.726} & 57.5M \\
    256 & 0.850 & 0.456 & 2.558 & 0.723 & 58M \\
    \hline
    \end{tabular}
}
\end{table}

\noindent\textbf{Comparison of Prompt vs. one-hot encoding:} In our experiments, we compared two approaches: one using generated prompts tailored to specific conditions, and another using a one-hot vector to represent class conditions. Our goal was to see how these methods influence the model’s ability to handle different types of data. \autoref{tab:compare_prompt_onehot} illustrates the results of this comparison. Using the prompt-based approach, the model demonstrates higher performance across all metrics. This method helps the model better distinguish between the diverse data distributions in each domain, as opposed to the more straightforward one-hot vector method.

\begin{table}[!t]
    \centering
    \caption{Prompt vs. One-Hot Encoding.}
    \label{prompt_onehot}
    \resizebox{\linewidth}{!}{
    \begin{tabular}{c|cccc|cccc|c}
    \multirow{2}{*}{\begin{tabular}[c]{@{}l@{}}Method\end{tabular}}  & \multicolumn{4}{c|}{Performance Metrics} & \multirow{2}{*}{\begin{tabular}[c]{@{}l@{}}\# Parameters\end{tabular}}  \\
    \cline{2-5} 
    & CC $\uparrow$ & KL $\downarrow$ & NSS $\uparrow$ & SIM $\uparrow$ &  \\
    \hline  
    \multicolumn{6}{c}{Avg. Performance on Salicon \cite{jiang2015salicon}} \\
    \hline
    SUM - One-hot & 0.902\scriptsize{$\pm$ 0.012} & 0.201\scriptsize{$\pm$ 0.024} & 1.97\scriptsize{$\pm$ 0.972} & 0.795\scriptsize{$\pm$ 0.012} & 57.3M \\
    \rowcolor[HTML]{DADCFF} \textbf{SUM - Prompt}  & \textbf{0.909\scriptsize{$\pm$ 0.011}} & \textbf{0.192\scriptsize{$\pm$ 0.025}} & \textbf{1.981\scriptsize{$\pm$ 0.201}} & \textbf{0.804\scriptsize{$\pm$ 0.012}} & 57.5M \\
    \hline
    \multicolumn{6}{c}{Avg. Performance Across Datasets} \\
    \hline
    SUM - One-hot & 0.843\scriptsize{$\pm$ 0.034} & 0.485\scriptsize{$\pm$ 0.046} & 2.583\scriptsize{$\pm$ 0.222} & 0.716\scriptsize{$\pm$ 0.023} & 57.3M \\
    \rowcolor[HTML]{DADCFF} \textbf{SUM - Prompt}  & \textbf{0.852\scriptsize{$\pm$ 0.029}} & \textbf{0.45\scriptsize{$\pm$ 0.053}} & \textbf{2.602\scriptsize{$\pm$ 0.206}} & \textbf{0.726\scriptsize{$\pm$ 0.022}} & 57.5M \\
    \end{tabular}
}
\label{tab:compare_prompt_onehot}
\end{table}

\noindent\textbf{Optimal C-VSS Placement in U-Net:}
We evaluated the impact of deploying the C-VSS in different sections of our U-Net structure: solely in the bottleneck, across all blocks of the decoder, and in every block of both the encoder and decoder. Our objective was to ascertain the optimal placement of the C-VSS for enhancing model performance. As summarized in \autoref{ablation_dit}, incorporating C-VSS in the encoder, in addition to the decoder, tends to undermine the features in the encoder, leading to suboptimal performance. This observation suggests that integrating C-VSS throughout the entire U-Net may disrupt the model's ability to leverage its foundational pre-trained features effectively. Conversely, limiting the use of C-VSS to the bottleneck provides some benefits but does not fully capitalize on the potential enhancements the module offers. The most effective strategy, as indicated by our results, is employing C-VSS across all decoder blocks. This approach allows the model to better adapt to the unique characteristics of each input domain, resulting in superior performance metrics compared to the other configurations tested.

\begin{table}[!t]
\centering
\caption{Comparison of C-VSS placement in the proposed U-Net structure.}
\label{ablation_dit}
\resizebox{0.9\linewidth}{!}{
    \begin{tabular}{l|cccc|c}
    \hline
    Configuration & CC $\uparrow$ & KL $\downarrow$ & NSS $\uparrow$ & SIM $\uparrow$ & \# Parameters \\
    \hline
    \multicolumn{6}{c}{Avg. Performance on Salicon \cite{jiang2015salicon}} \\
    \hline
    Bottleneck & \textbf{0.909} & 0.195 & 1.97 & \textbf{0.804} & 57.37M \\
    \rowcolor[HTML]{DADCFF} Decoder & \textbf{0.909} & \textbf{0.192} & \textbf{1.981} & \textbf{0.804} & 57.5M \\
    All-Blocks & 0.907 & 0.198 & 1.975 & 0.801 & 58.5M \\
    \hline
    \multicolumn{6}{c}{Avg. Performance Across Datasets} \\
    \hline
    Bottleneck & 0.847 & 0.466 & 2.581 & 0.724 & 57.37M \\
    \rowcolor[HTML]{DADCFF} Decoder & 0.852 & \textbf{0.450} & \textbf{2.602} & \textbf{0.726} & 57.5M \\
    All-Blocks & \textbf{0.854} & 0.458 & 2.601 & 0.724 & 58.5M \\
    \hline
    \end{tabular}
}
\end{table}

\section{Conclusion}
In this paper, we have presented \textbf{SUM}, a novel approach designed to address the limitations of traditional saliency prediction models. By integrating the Mamba architecture with U-Net and enhancing it with a Conditional Visual State Space (C-VSS) block, SUM adapts dynamically to various image types, making it universally applicable across diverse visual contexts. Our extensive evaluations across six benchmark datasets demonstrated SUM's superior performance, consistently outperforming existing models. The model excelled in both location-based and distribution-based metrics, proving its robustness and adaptability to use in real-world problems.

{\small
\bibliographystyle{ieeenat_fullname}
\bibliography{main}
}

\clearpage
\setcounter{page}{1}
\maketitlesupplementary

\appendix

\section{Experimental Results}

\subsection{Impact of different loss combinations}
In addition, to provide additional details about the coefficients used for each loss combination, we conducted several experiments to determine the optimal coefficients for each combination. The best coefficients for each combination are depicted in \autoref{tab:loss_coef}.

\subsection{More visualization results}
\begin{figure*}[htb]
    \centering
    \resizebox{0.9\textwidth}{!}{
    \begin{tabular}{@{} *{7}c @{}}
    \includegraphics[width=0.25\textwidth]{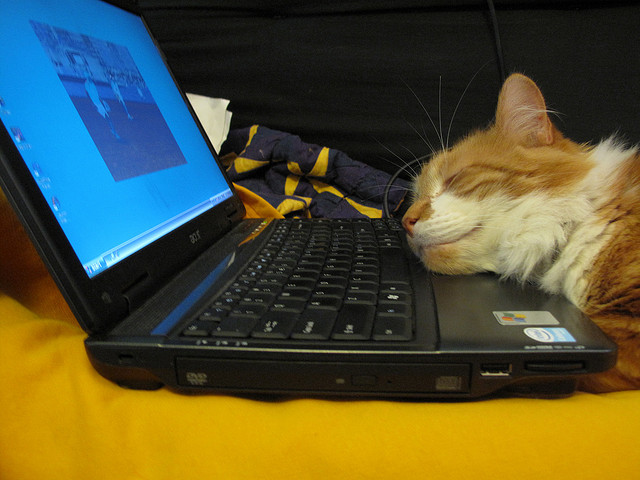} &
    \includegraphics[width=0.25\textwidth]{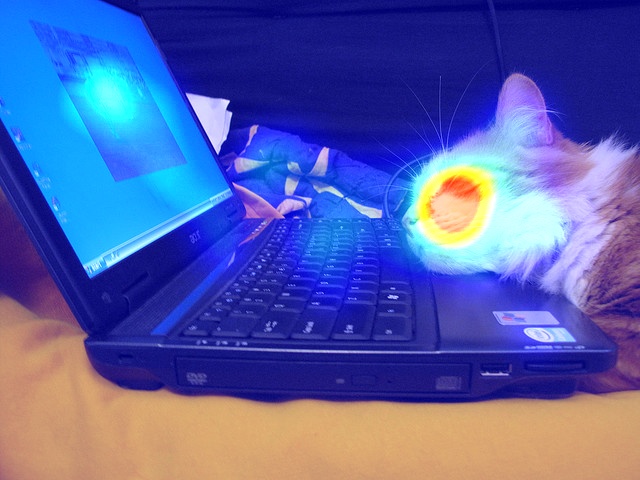} &
    \includegraphics[width=0.25\textwidth]{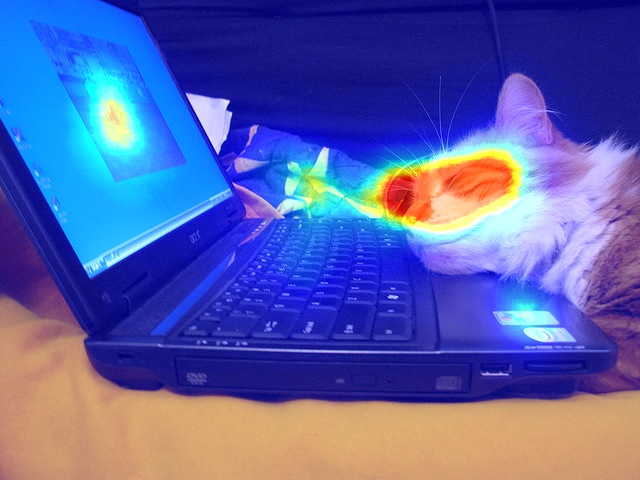} &
    \includegraphics[width=0.25\textwidth]{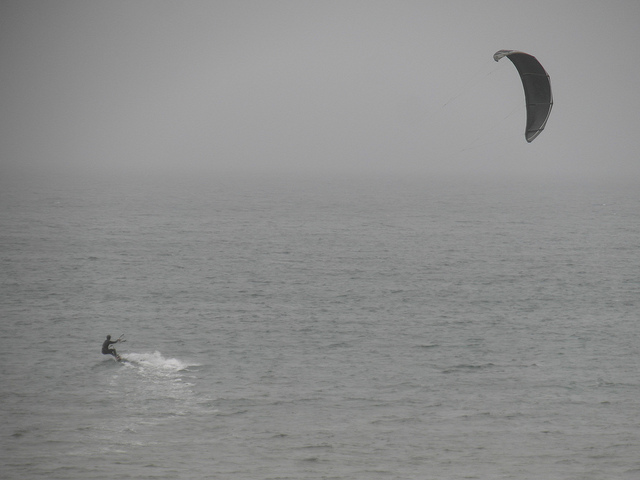} &
    \includegraphics[width=0.25\textwidth]{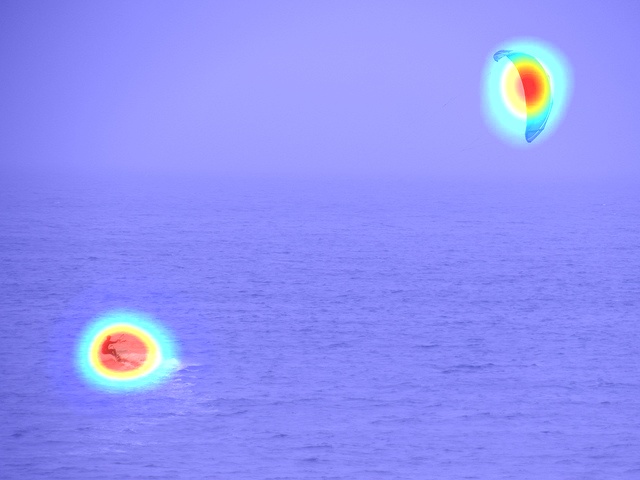} &
    \includegraphics[width=0.25\textwidth]{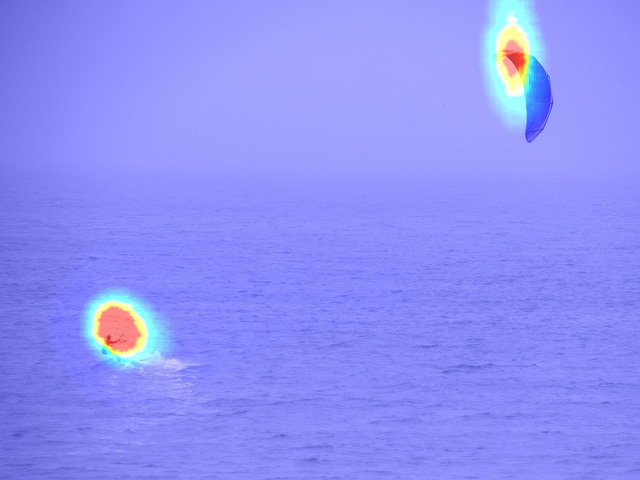} 
    \\
    \\
     \includegraphics[width=0.25\textwidth]{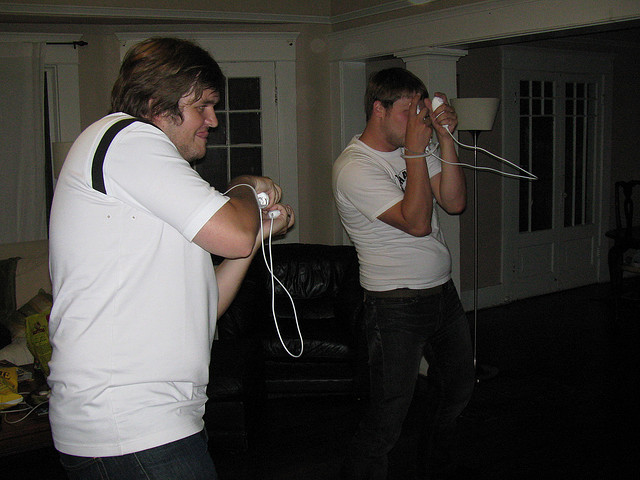} &
    \includegraphics[width=0.25\textwidth]{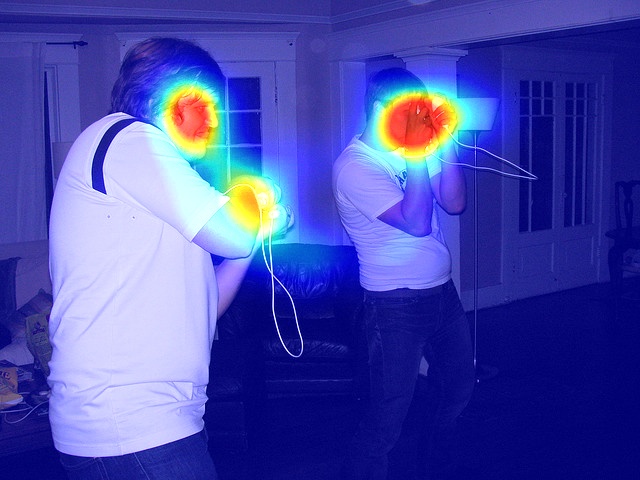} &
    \includegraphics[width=0.25\textwidth]{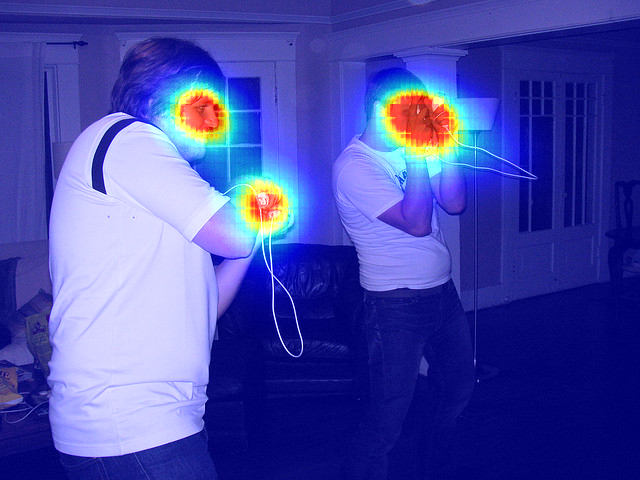} &
    \includegraphics[width=0.25\textwidth]{sec/figures/subb_visualization/natural_scene_mouse/1_4_org.jpg} &
    \includegraphics[width=0.25\textwidth]{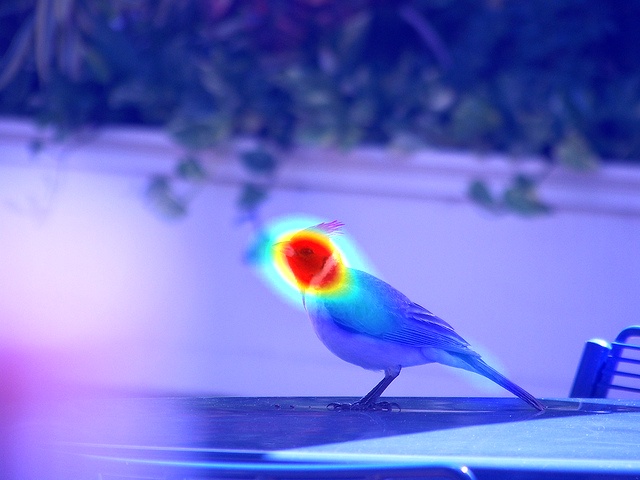} &
    \includegraphics[width=0.25\textwidth]{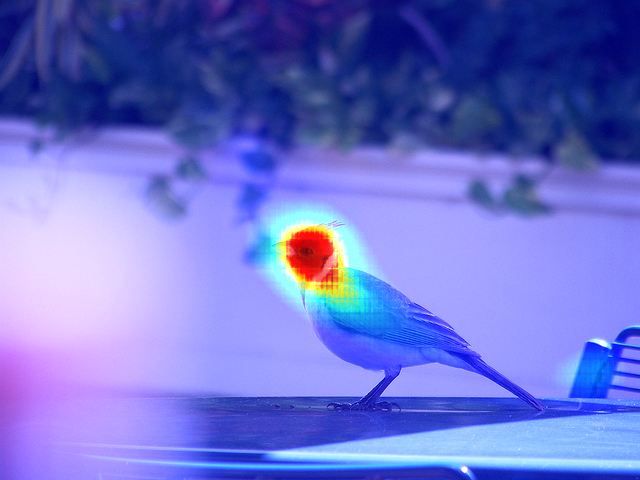} 
    \\
    \\
    \includegraphics[width=0.25\textwidth]{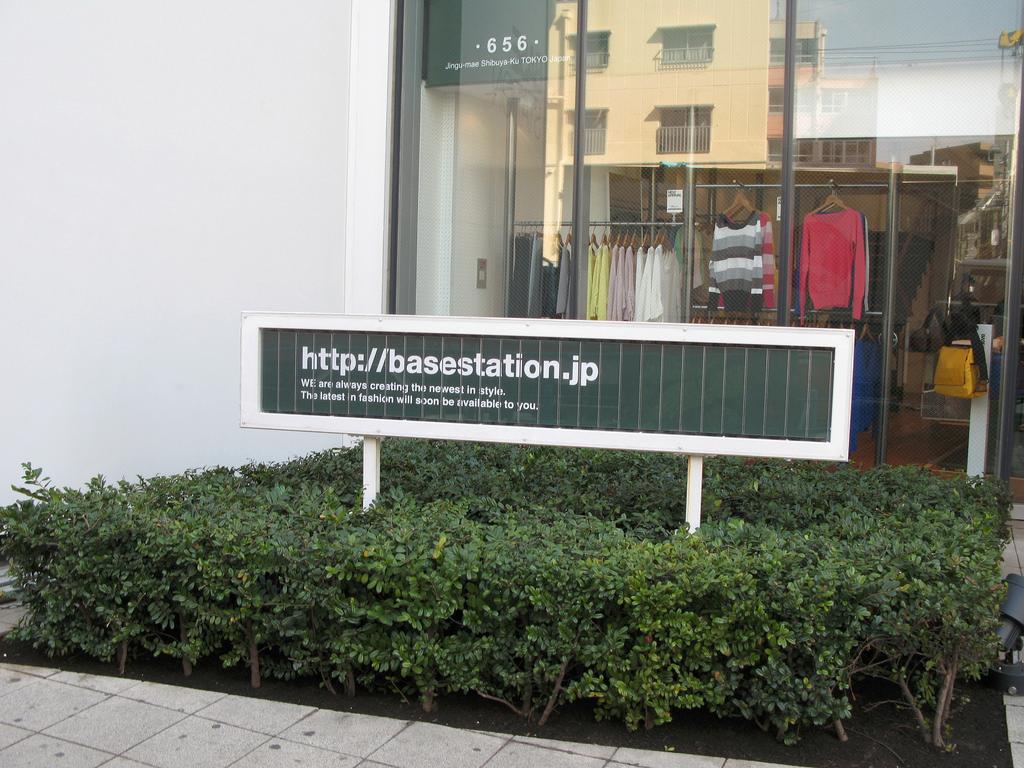} &
    \includegraphics[width=0.25\textwidth]{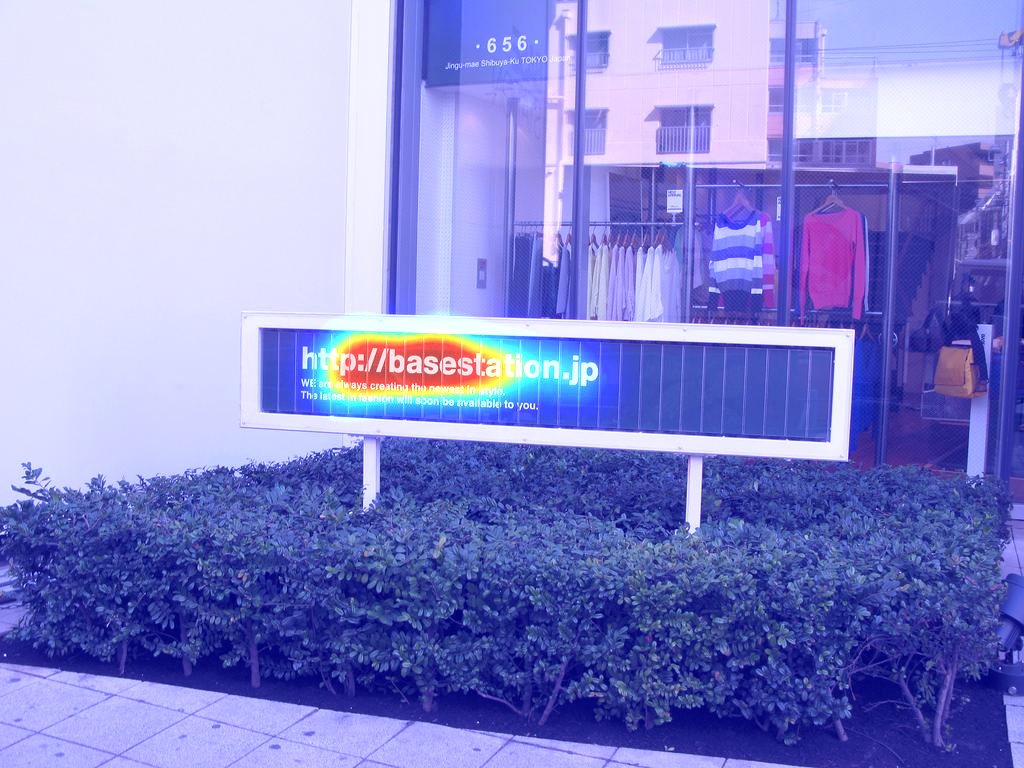} &
    \includegraphics[width=0.25\textwidth]{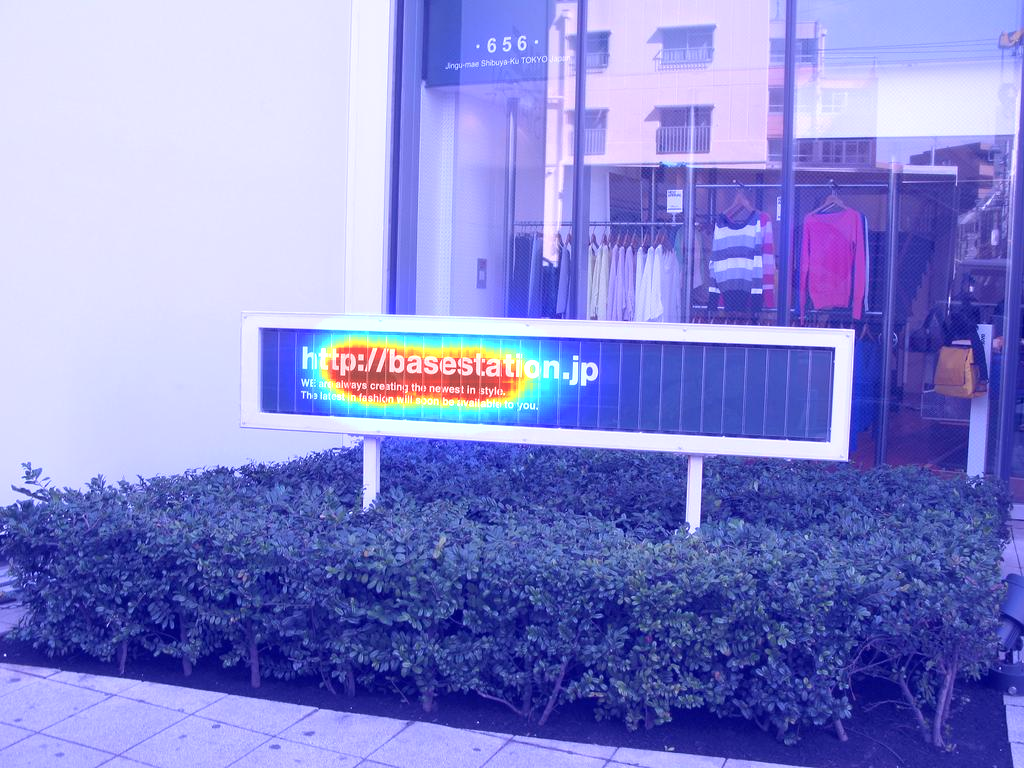} &
    \includegraphics[width=0.25\textwidth]{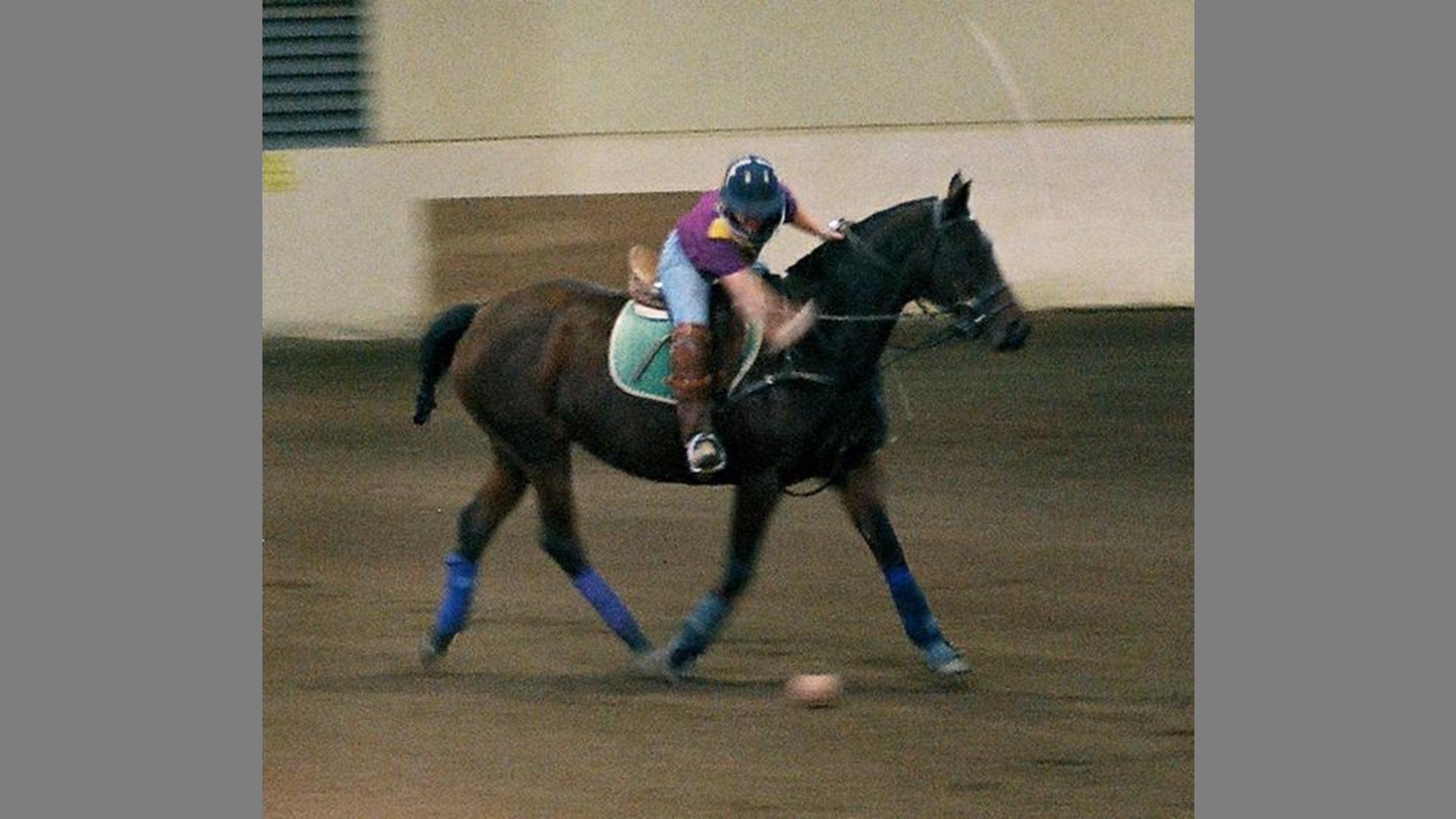} &
    \includegraphics[width=0.25\textwidth]{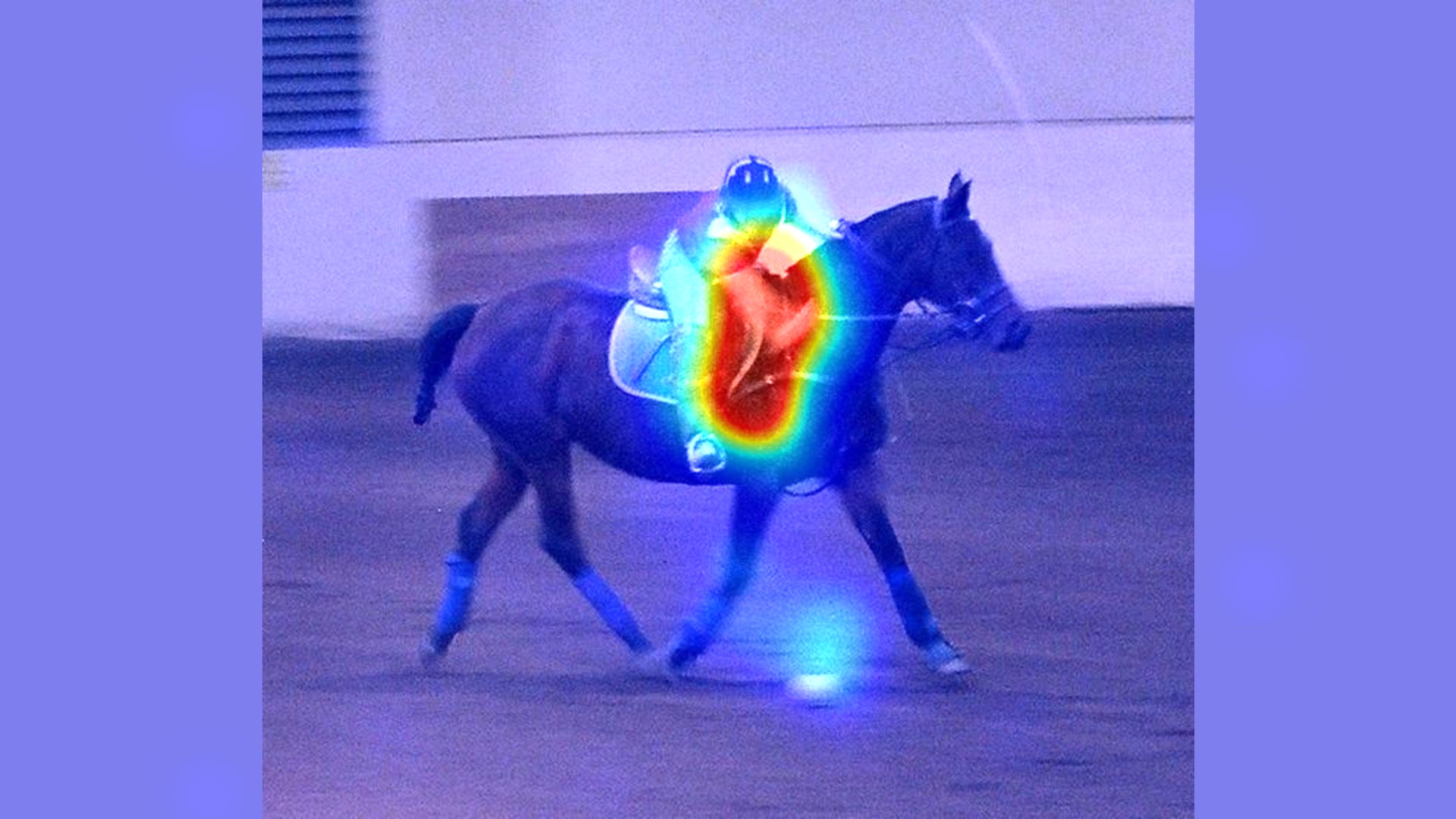} &
    \includegraphics[width=0.25\textwidth]{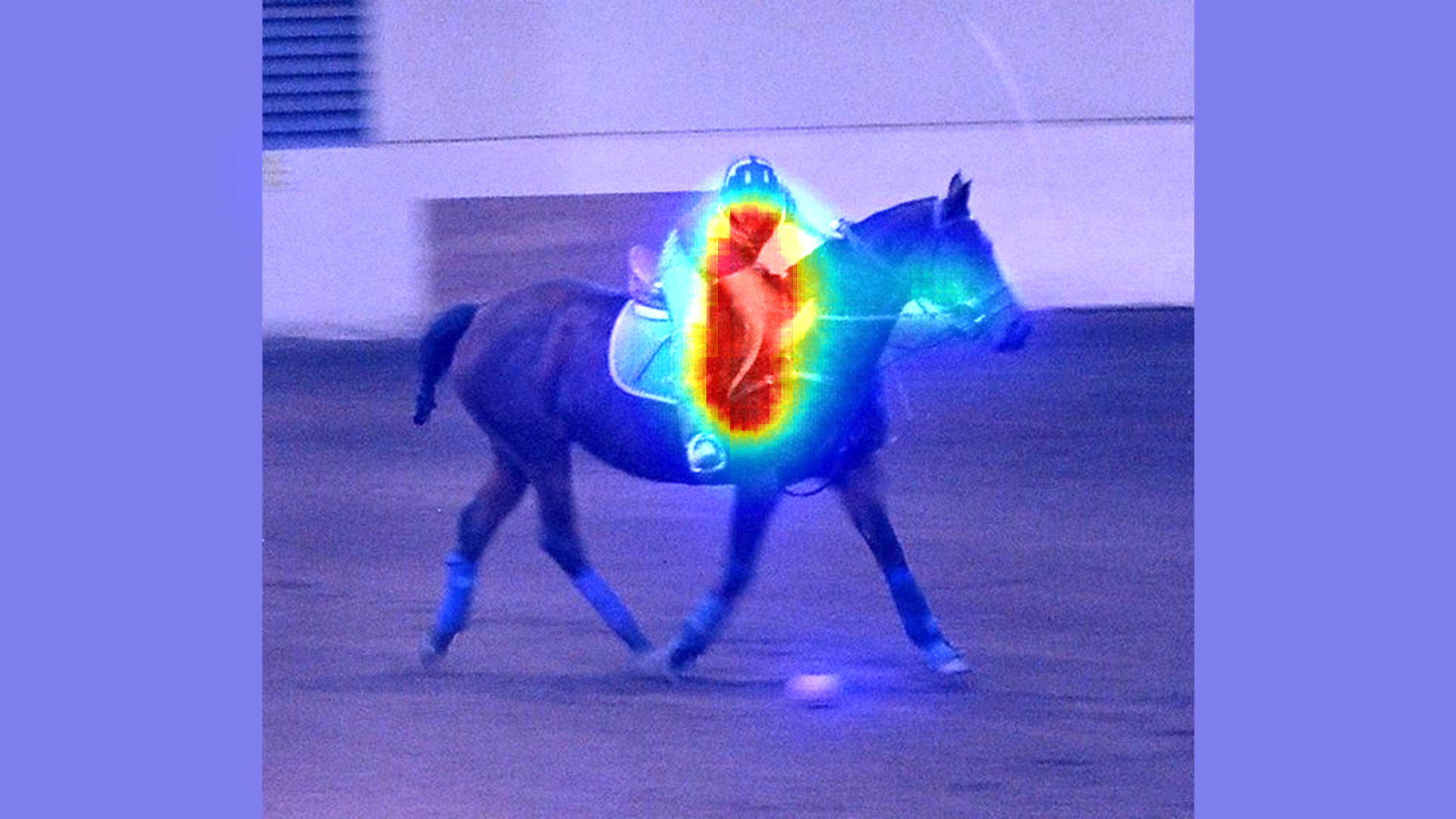} 
    \\
    \\
    \includegraphics[width=0.25\textwidth]{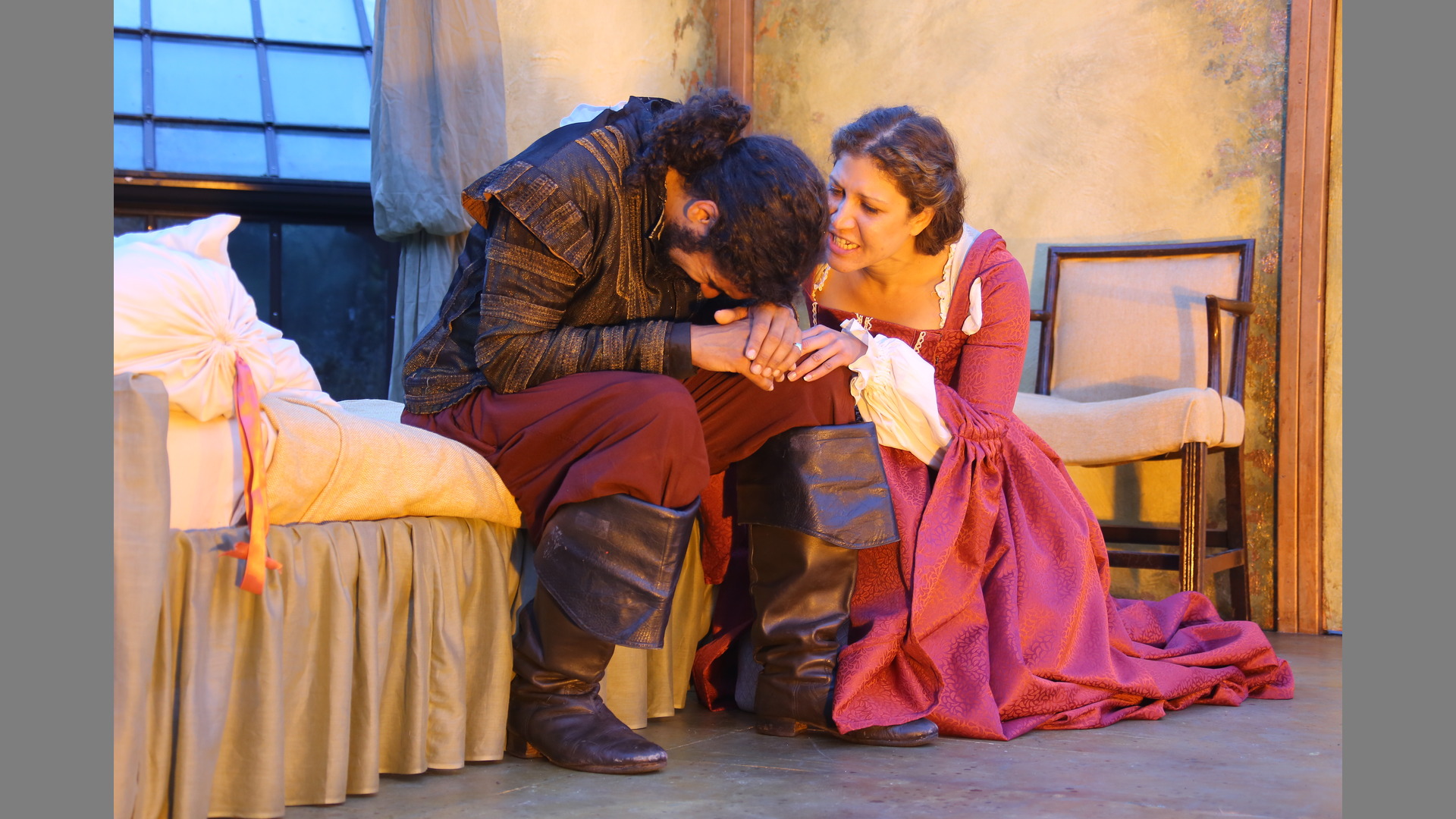} &
    \includegraphics[width=0.25\textwidth]{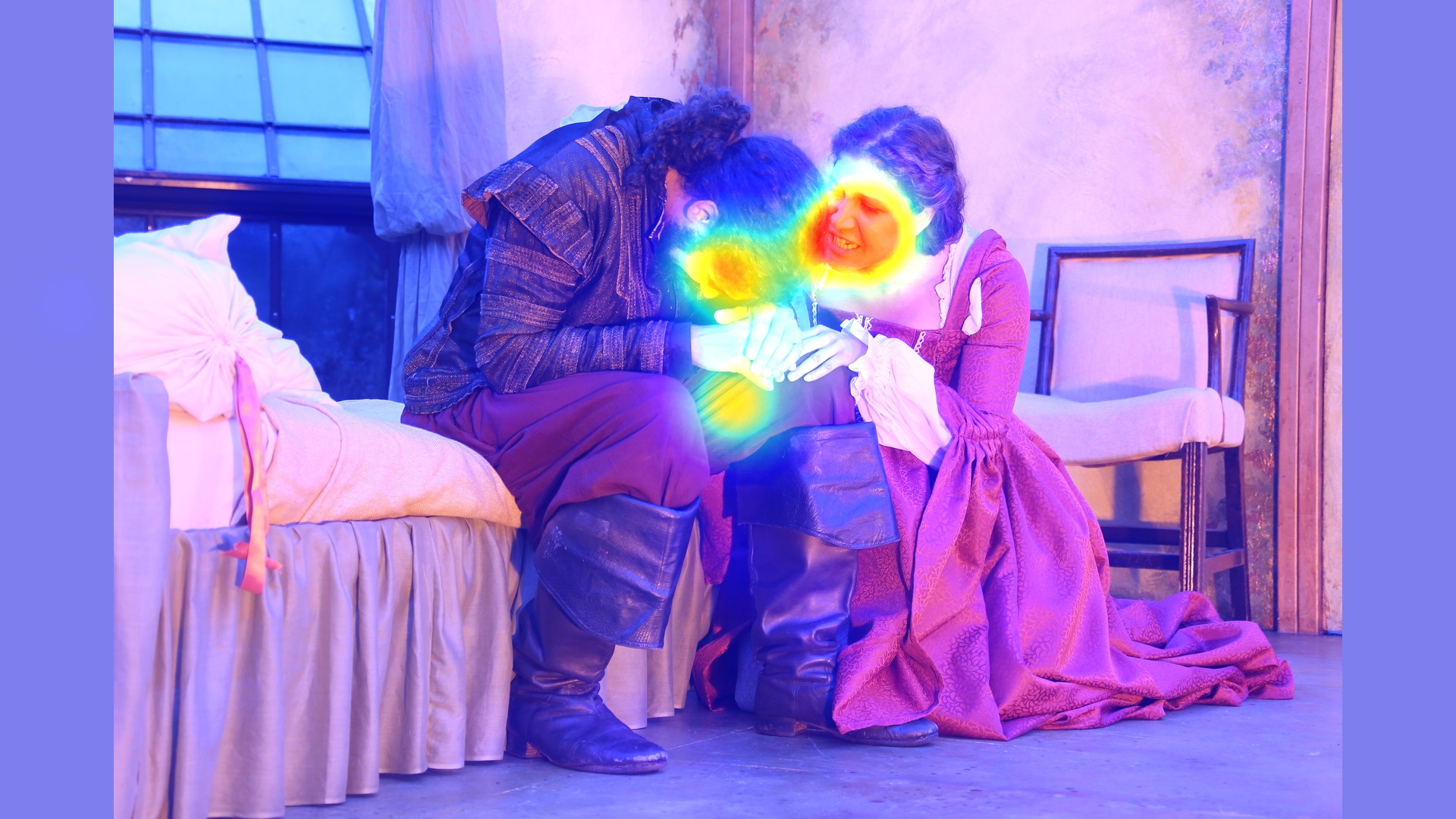} &
    \includegraphics[width=0.25\textwidth]{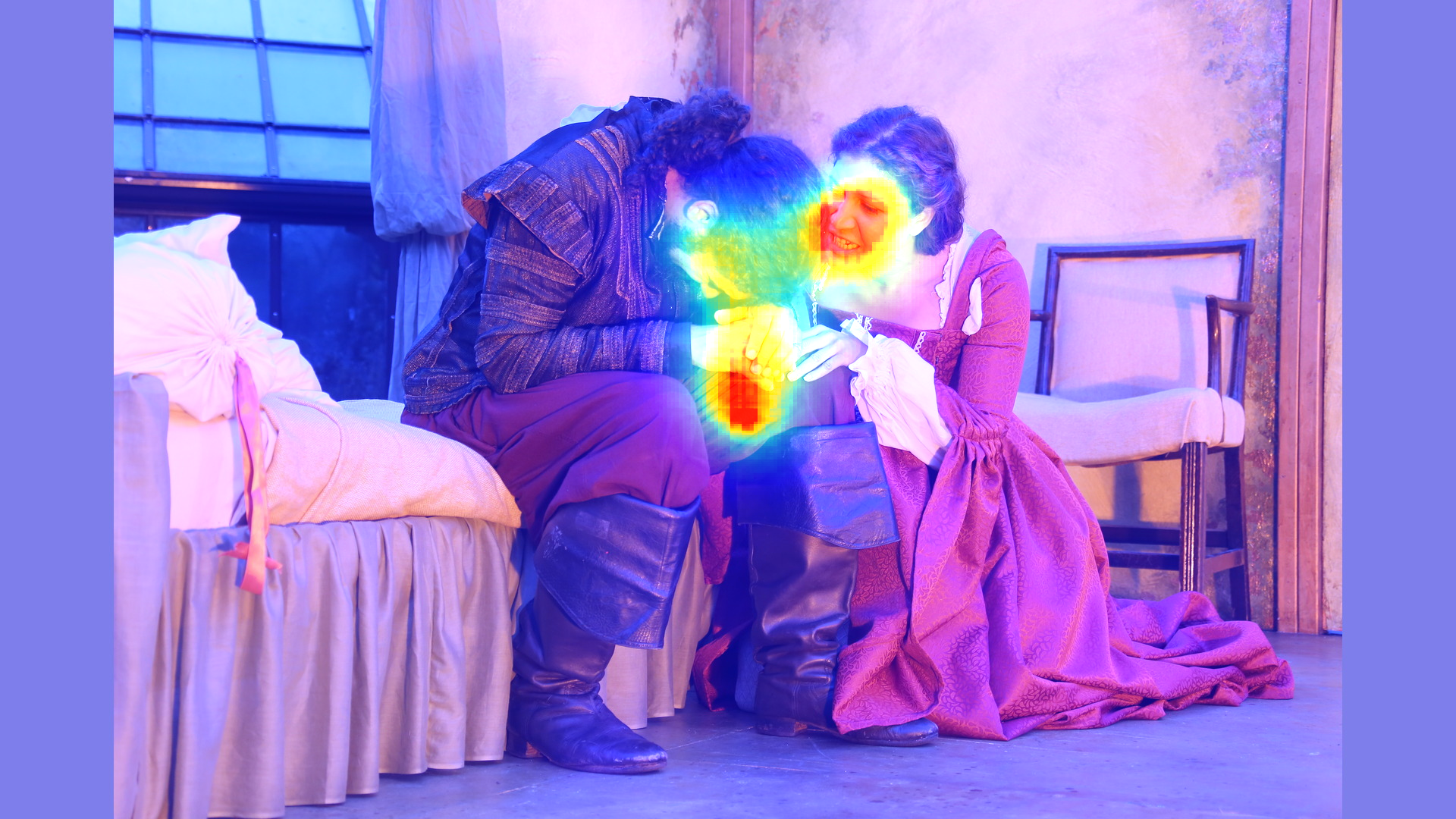} &
    \includegraphics[width=0.25\textwidth]{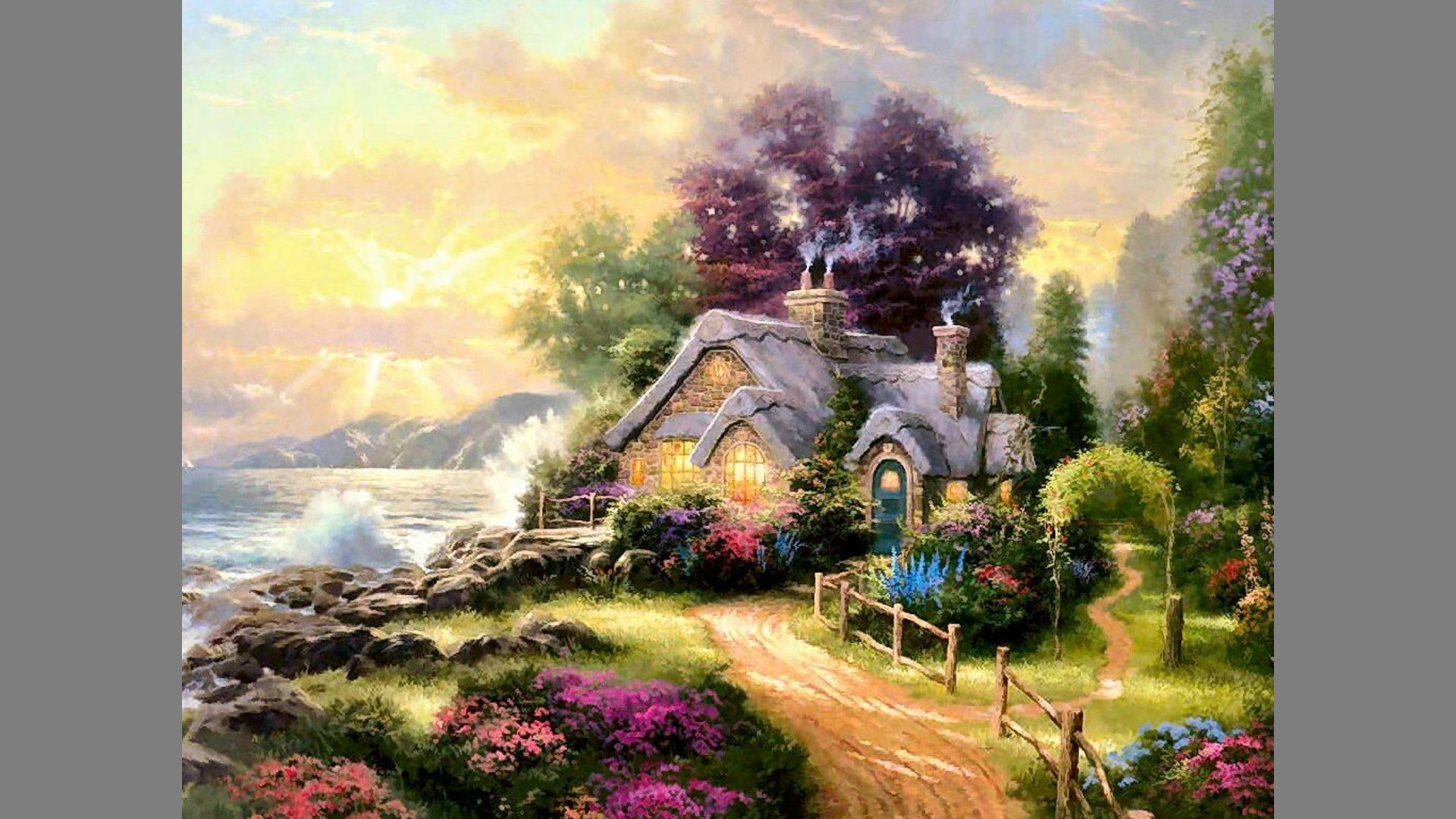} &
    \includegraphics[width=0.25\textwidth]{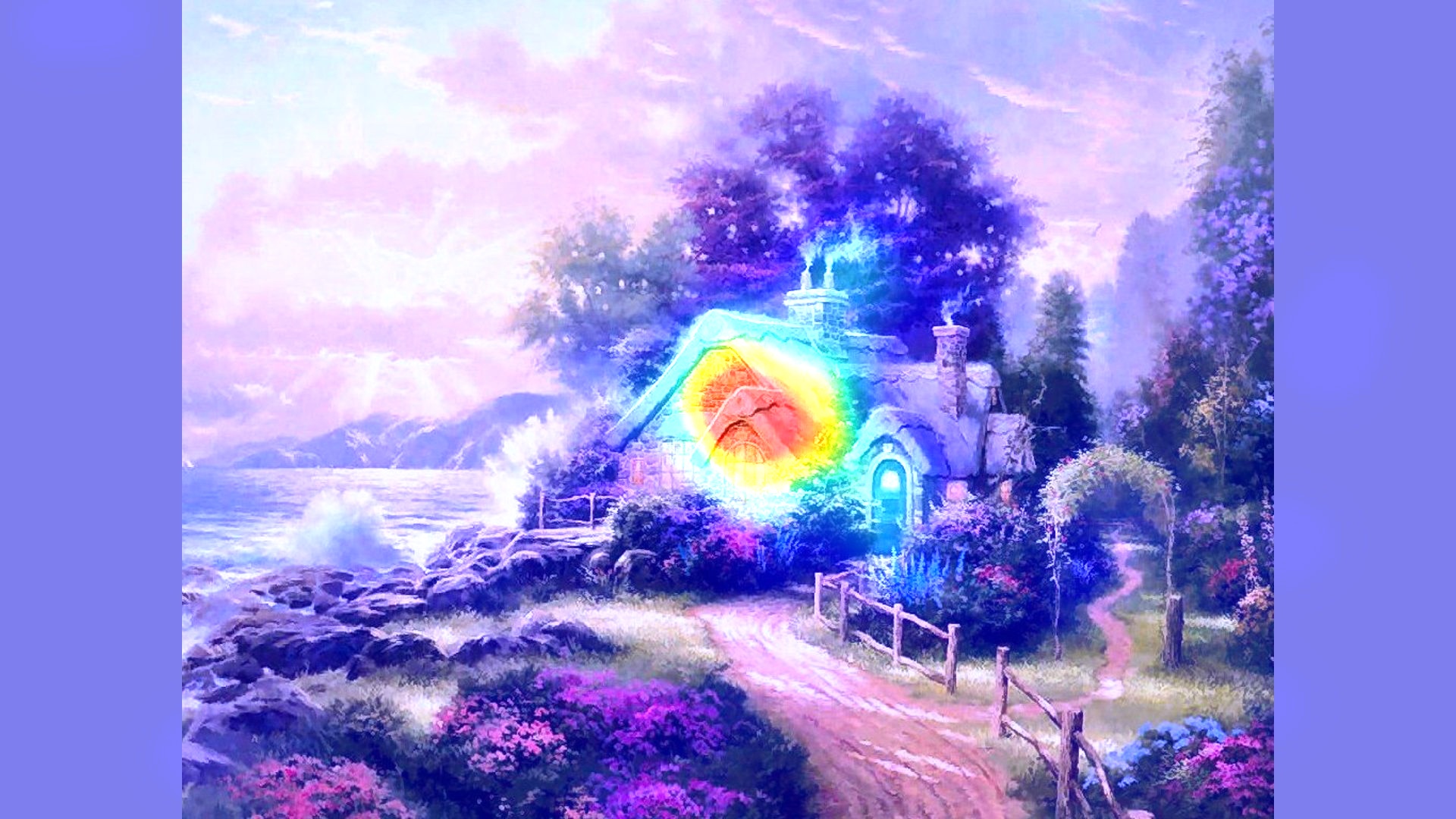} &
    \includegraphics[width=0.25\textwidth]{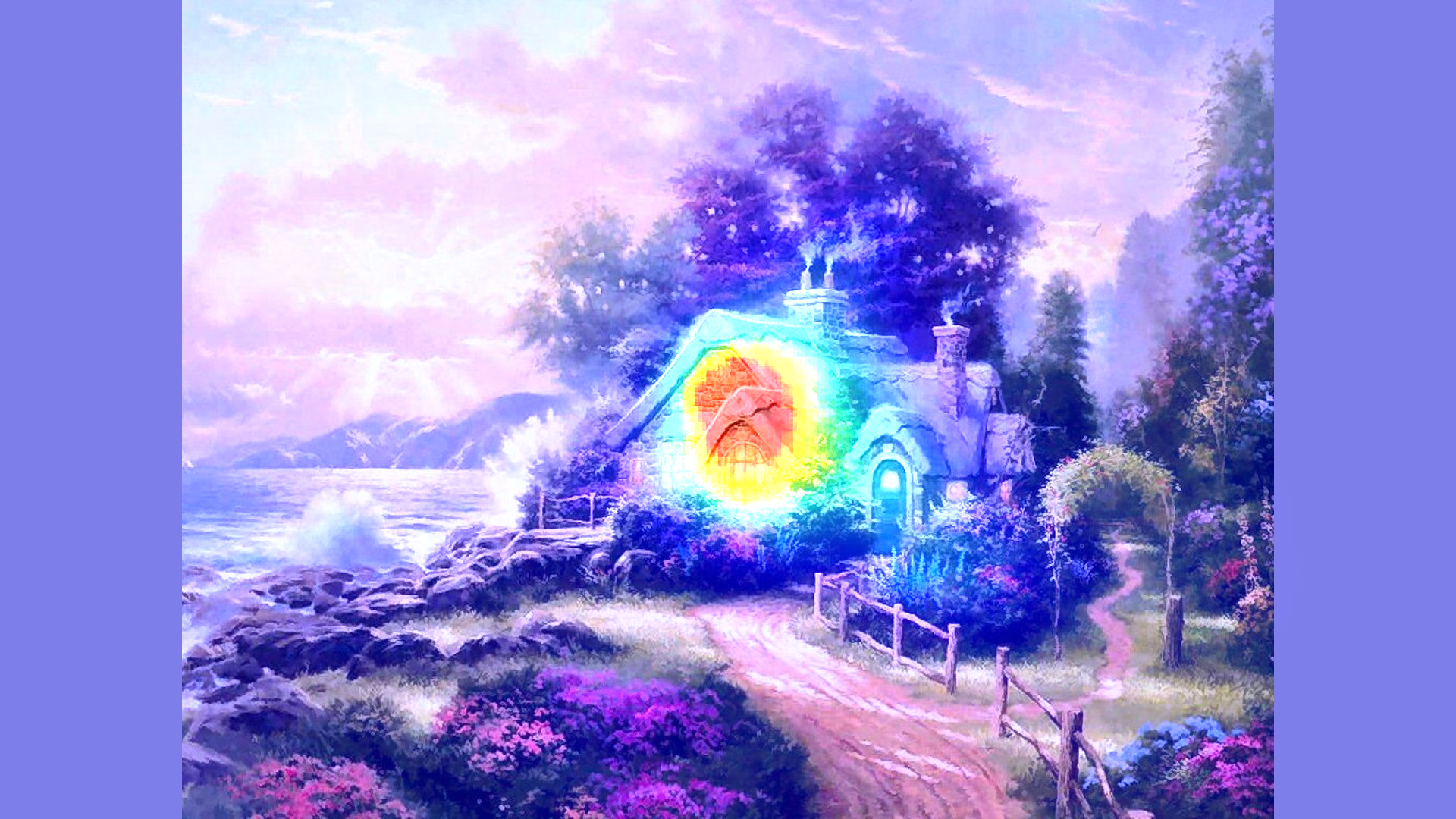}
    \\
    \\
    \includegraphics[width=0.25\textwidth]{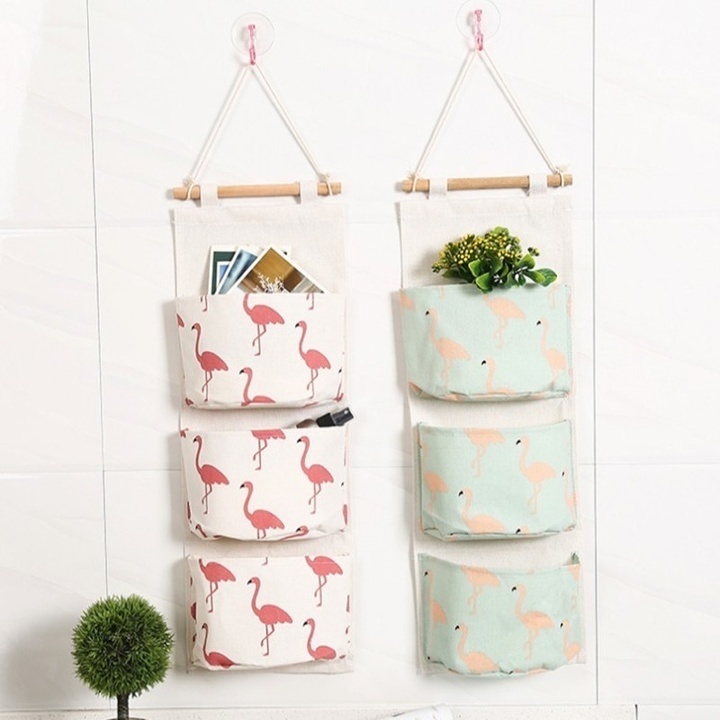} &
    \includegraphics[width=0.25\textwidth]{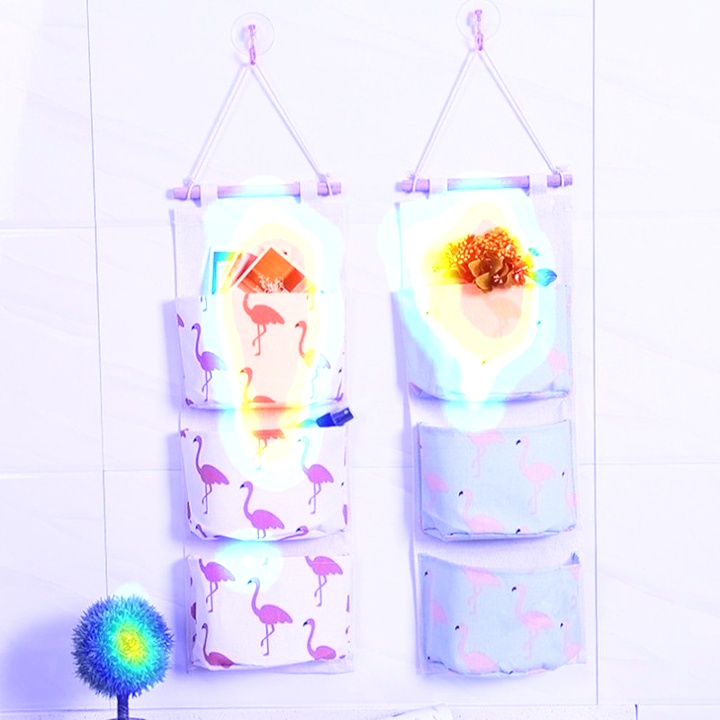} &
    \includegraphics[width=0.25\textwidth]{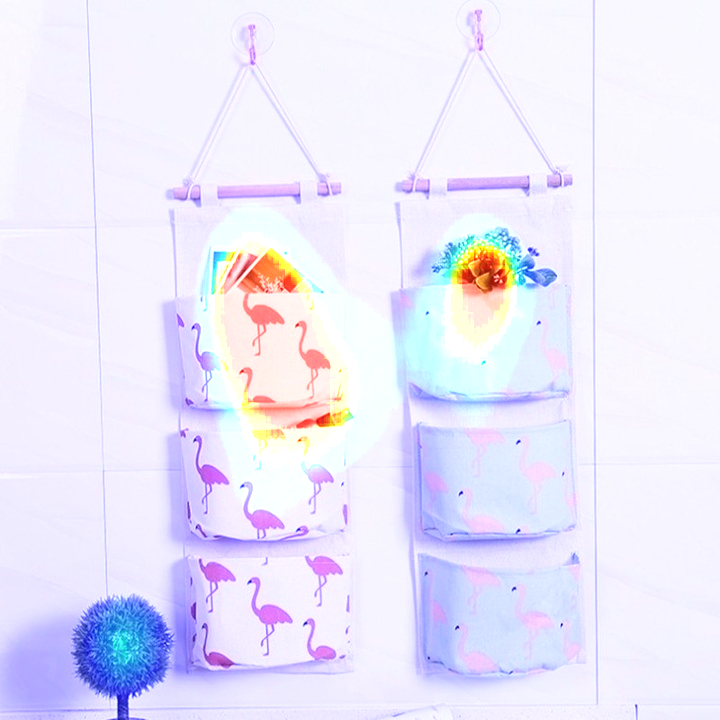} &
    \includegraphics[width=0.25\textwidth]{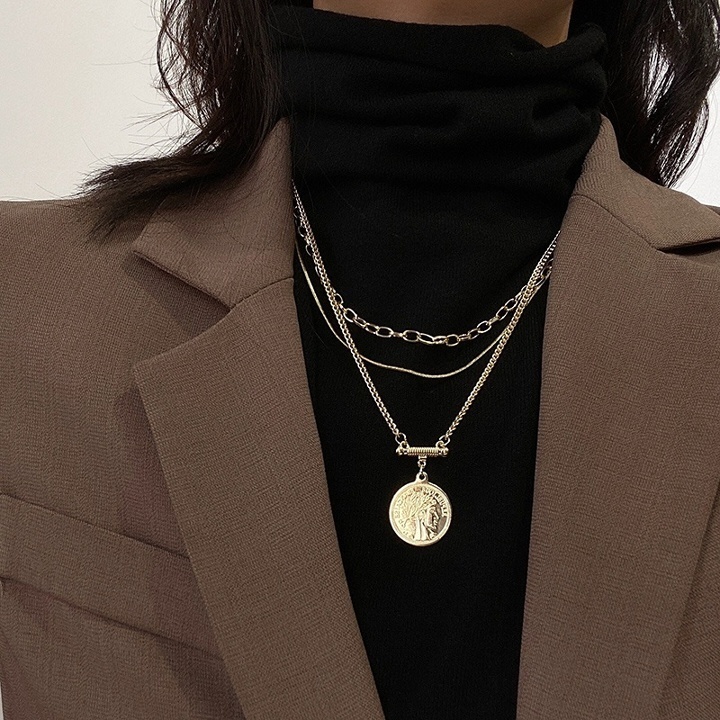} &
    \includegraphics[width=0.25\textwidth]{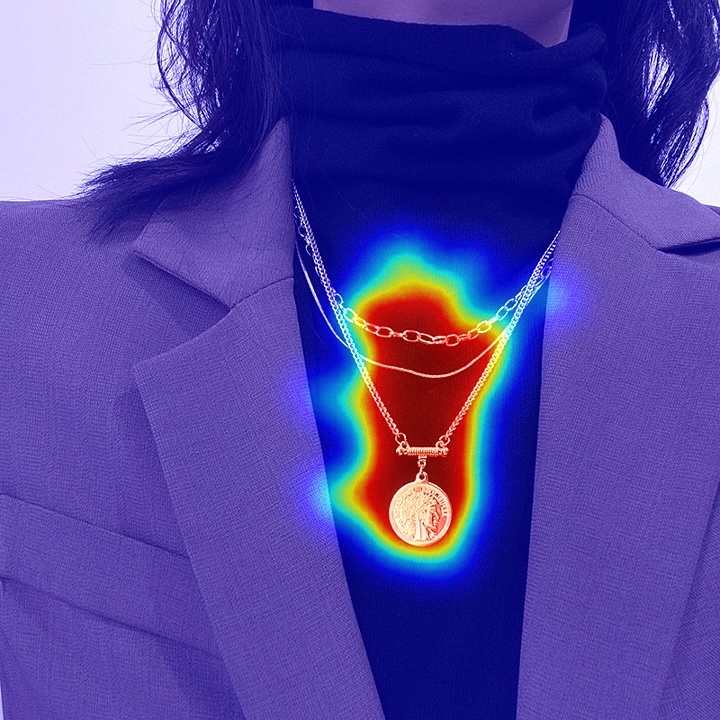} &
    \includegraphics[width=0.25\textwidth]{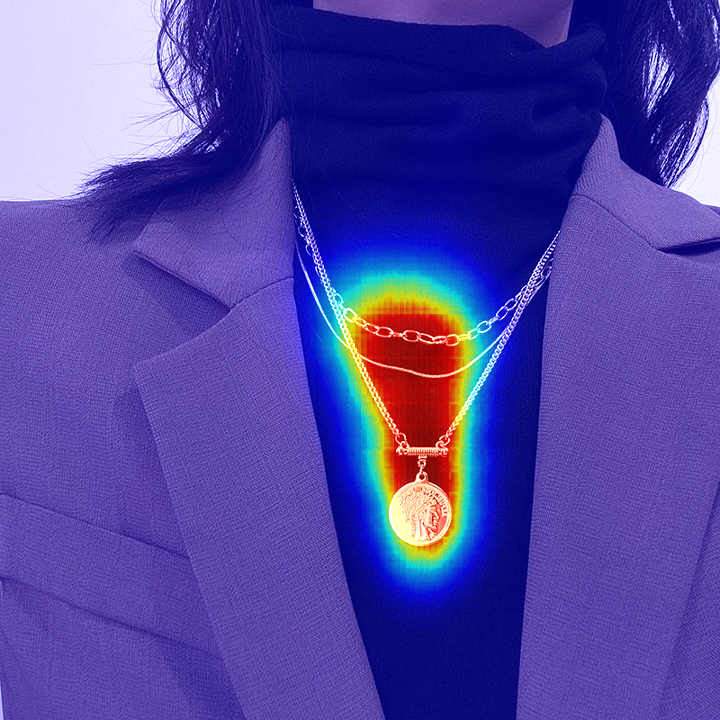}
    \\
    \\
    \includegraphics[width=0.25\textwidth]{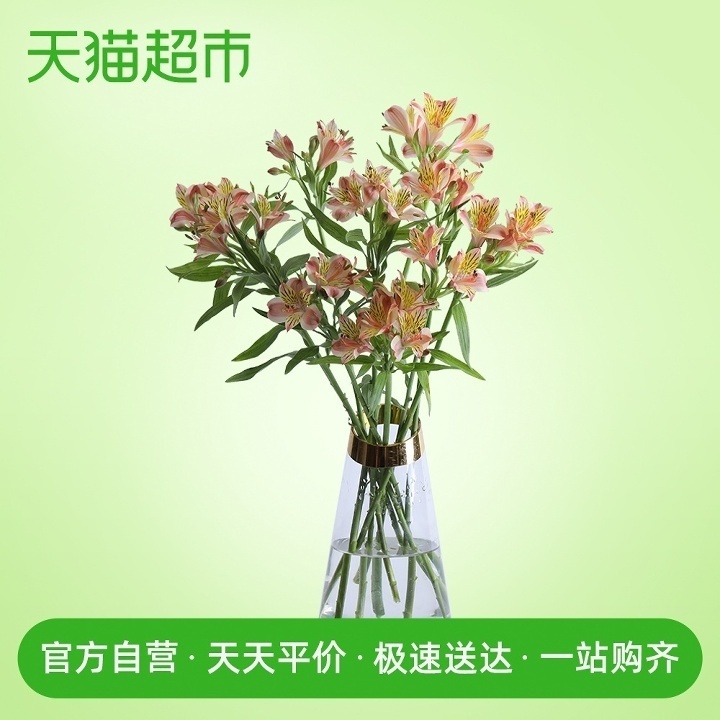} &
    \includegraphics[width=0.25\textwidth]{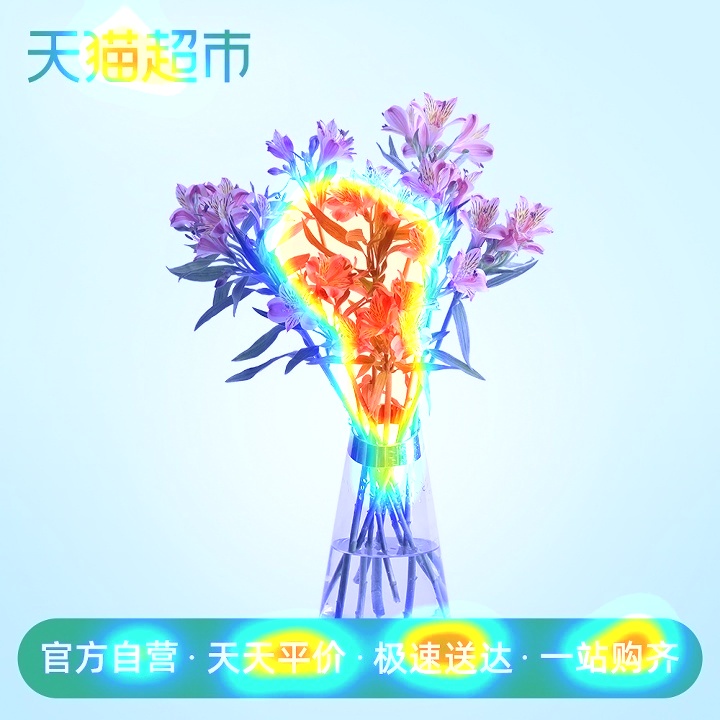} &
    \includegraphics[width=0.25\textwidth]{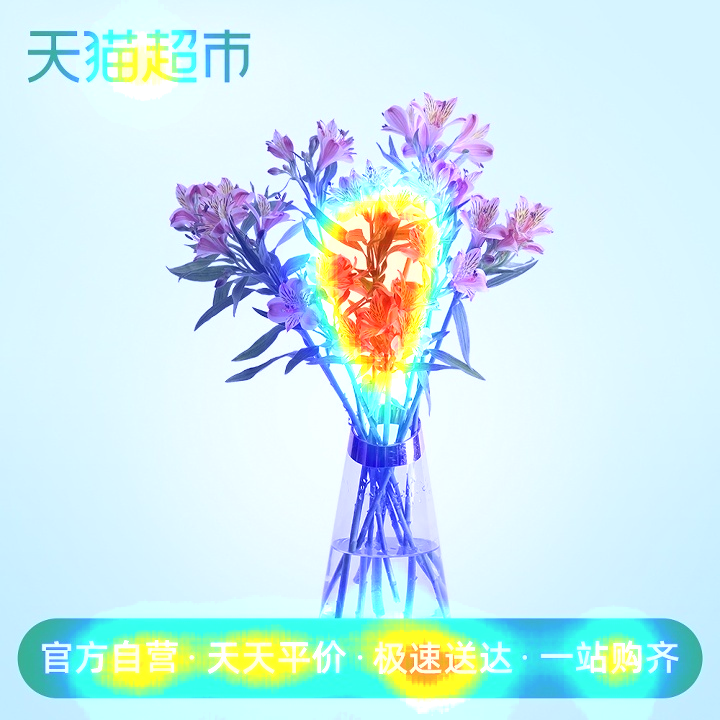} &
    \includegraphics[width=0.25\textwidth]{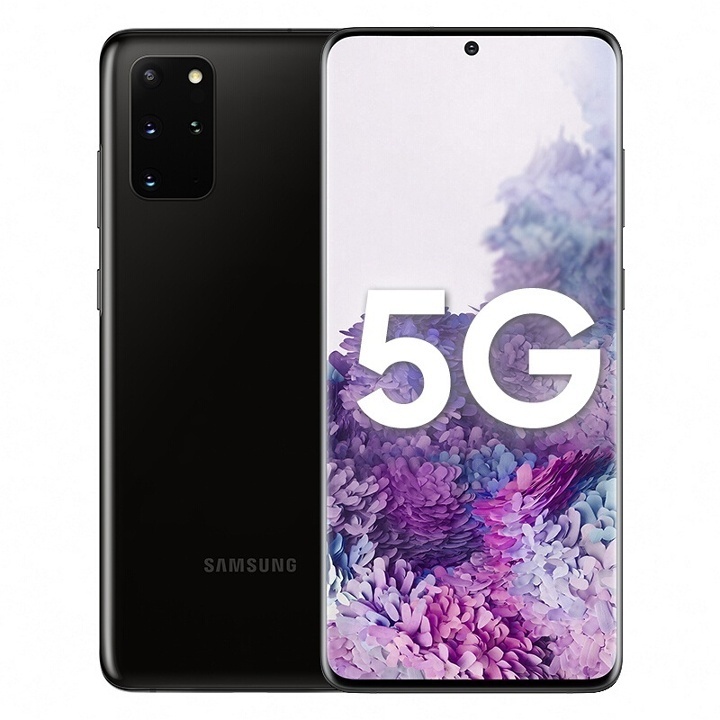} &
    \includegraphics[width=0.25\textwidth]{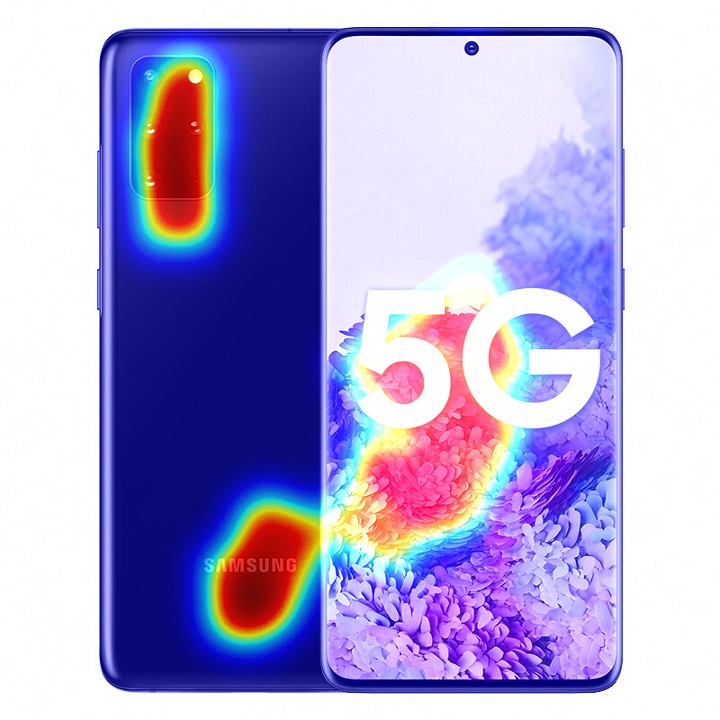} &
    \includegraphics[width=0.25\textwidth]{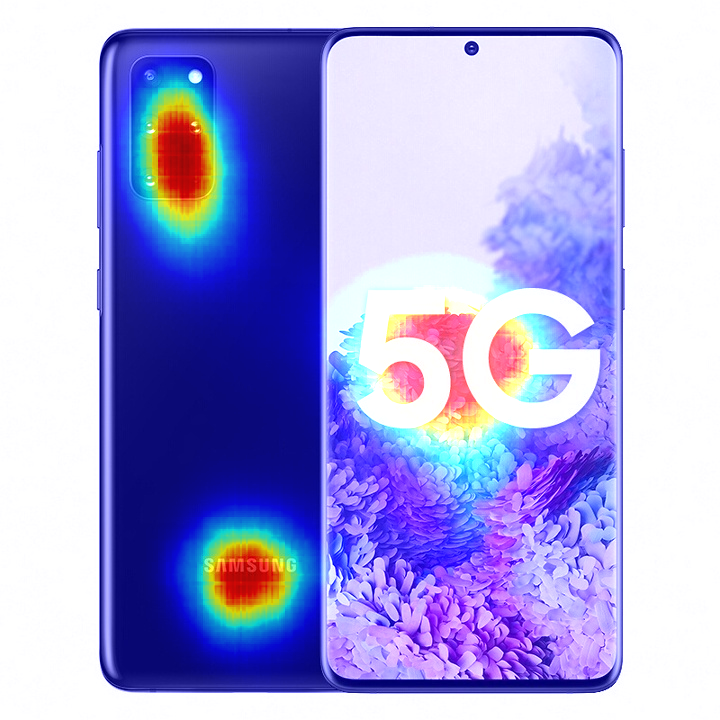}
    \\
    \\
    \includegraphics[width=0.25\textwidth]{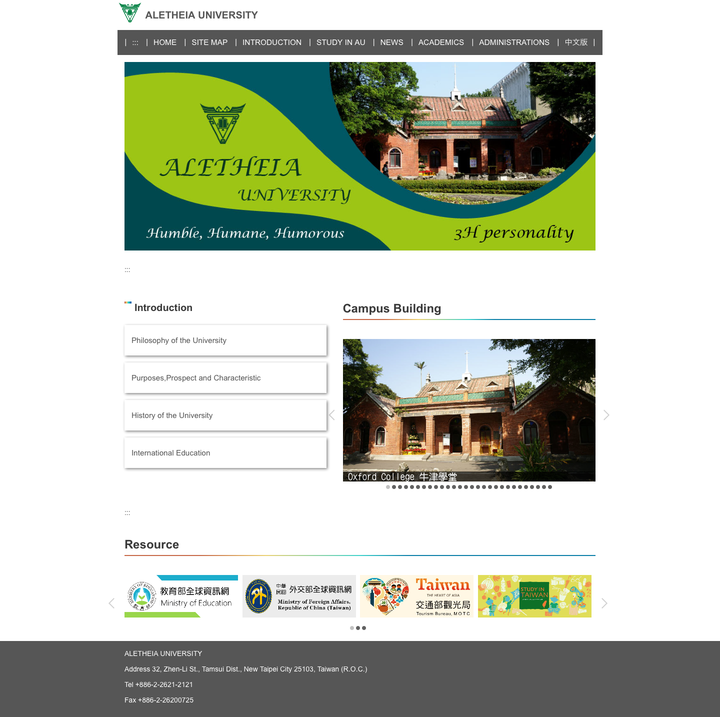} &
    \includegraphics[width=0.25\textwidth]{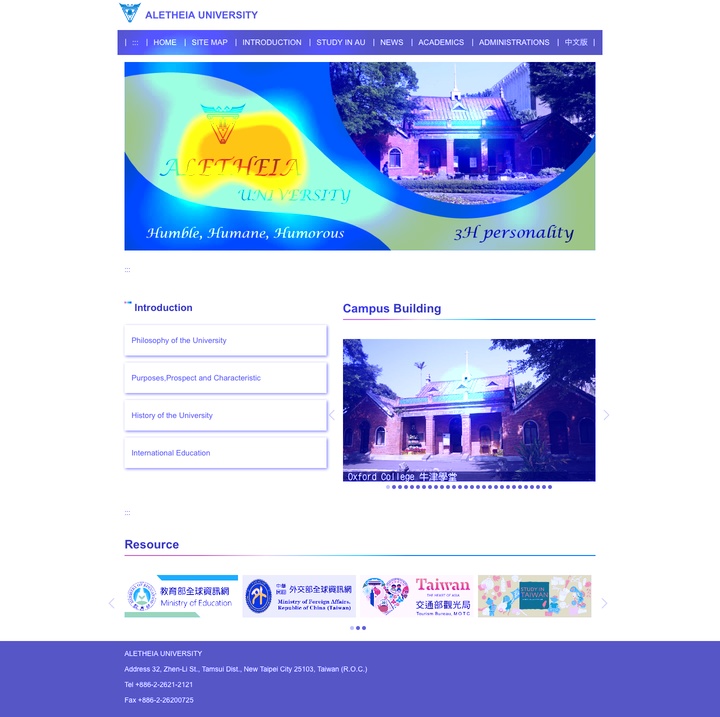} &
    \includegraphics[width=0.25\textwidth]{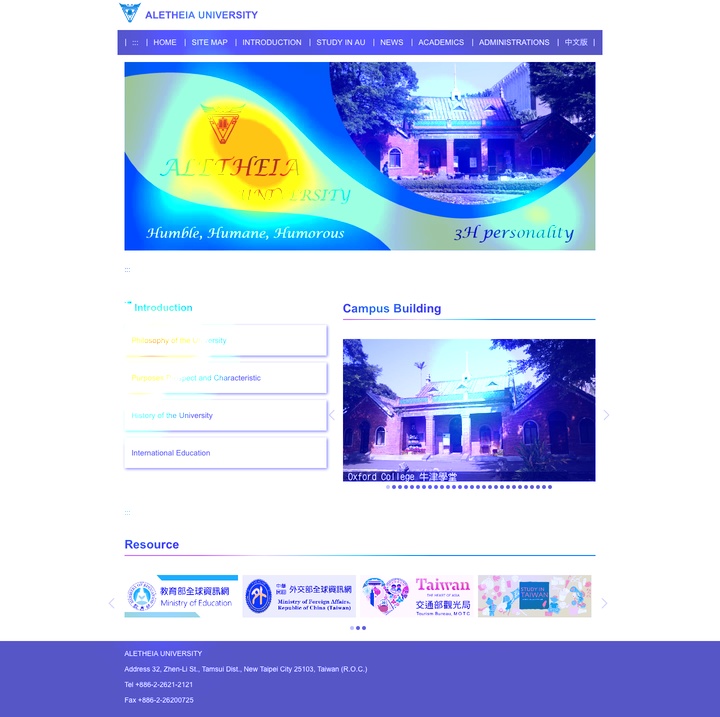} &
    \includegraphics[width=0.1\textwidth]{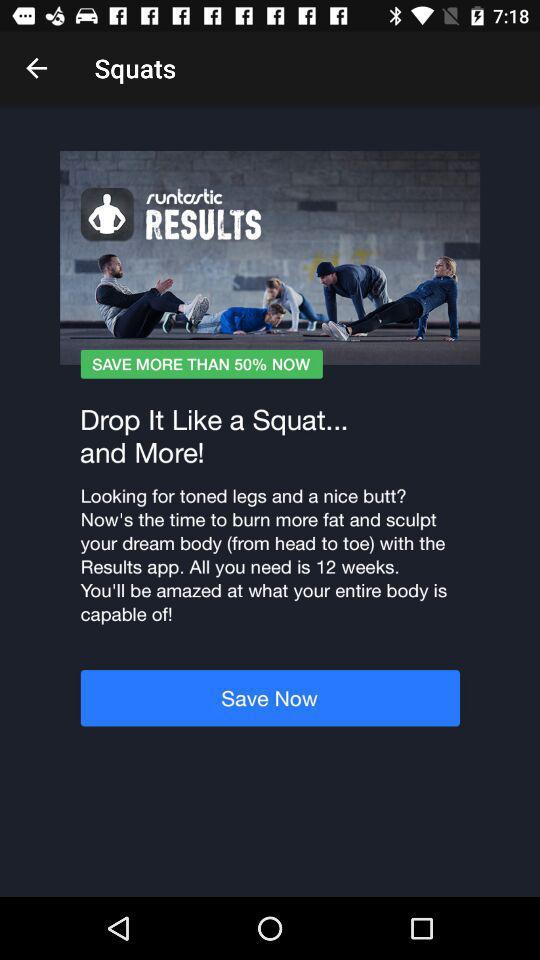} &
    \includegraphics[width=0.1\textwidth]{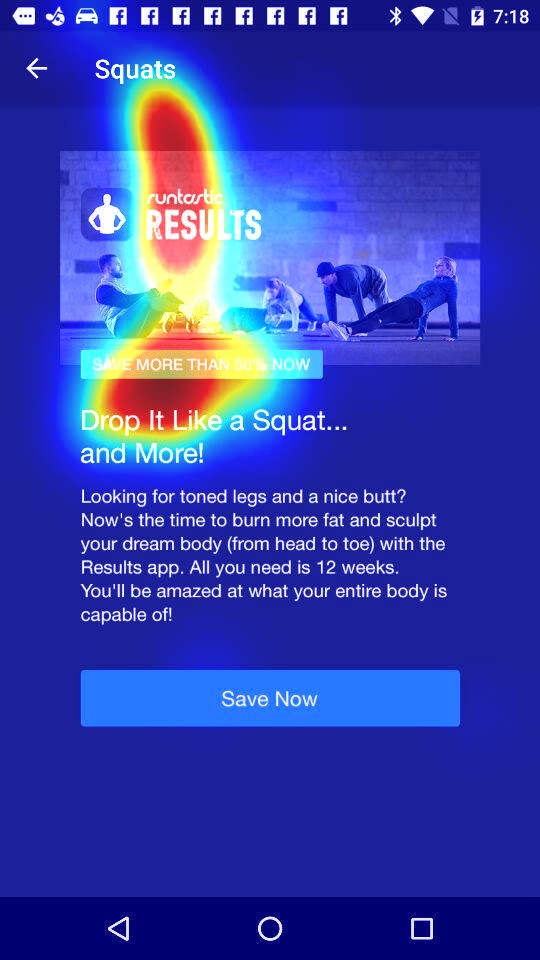} &
    \includegraphics[width=0.1\textwidth]{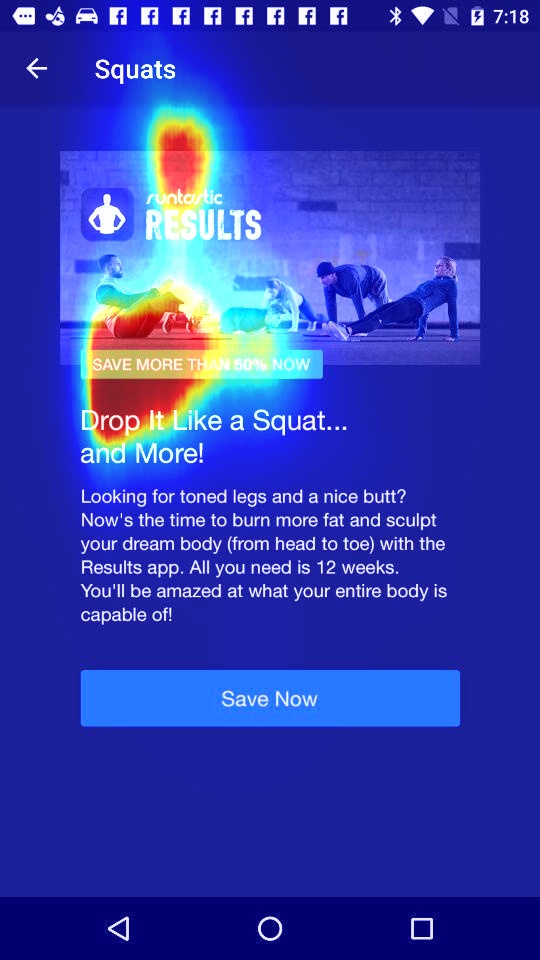}
    \\
    \\
    \includegraphics[width=0.25\textwidth]{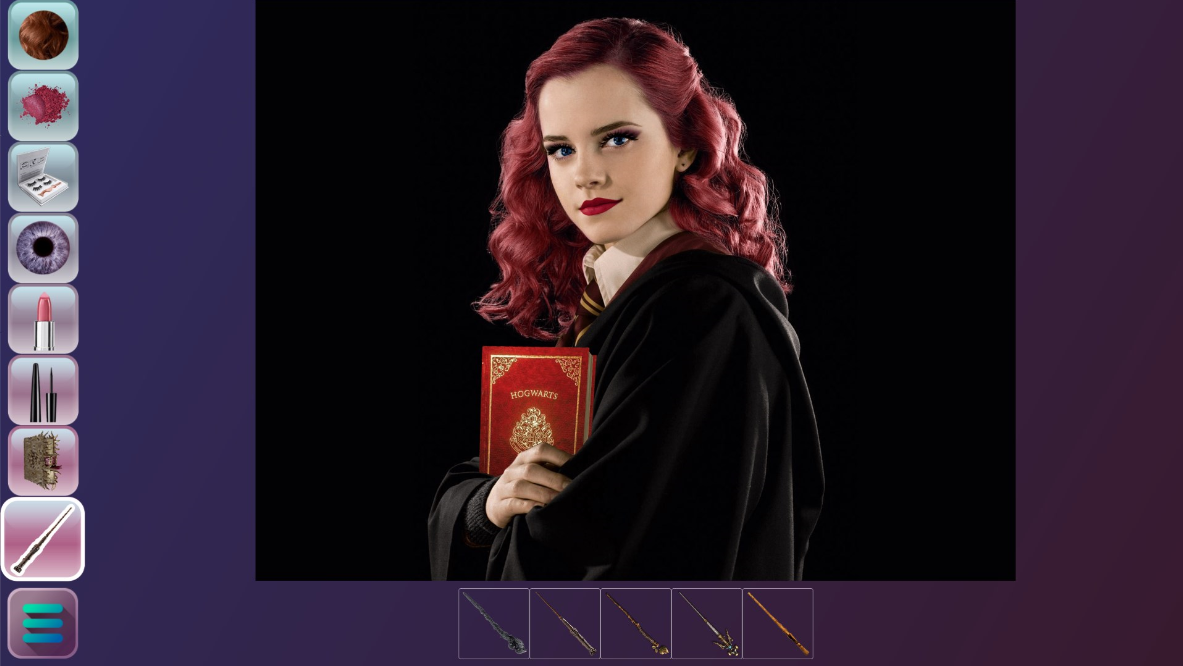} &
    \includegraphics[width=0.25\textwidth]{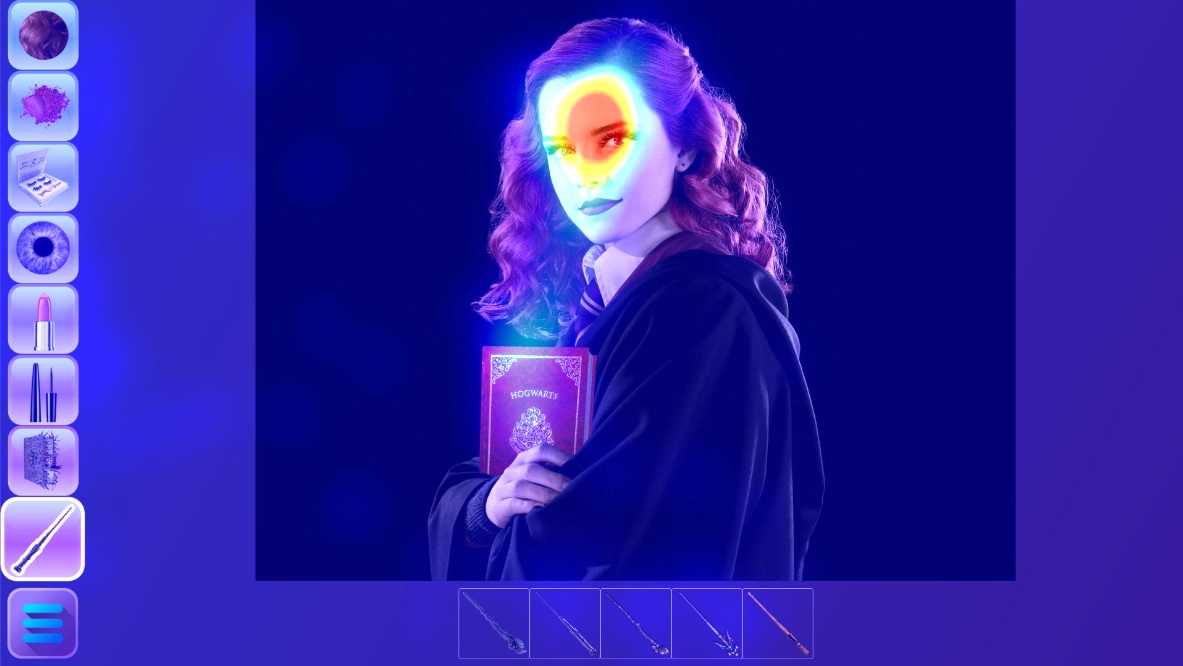} &
    \includegraphics[width=0.25\textwidth]{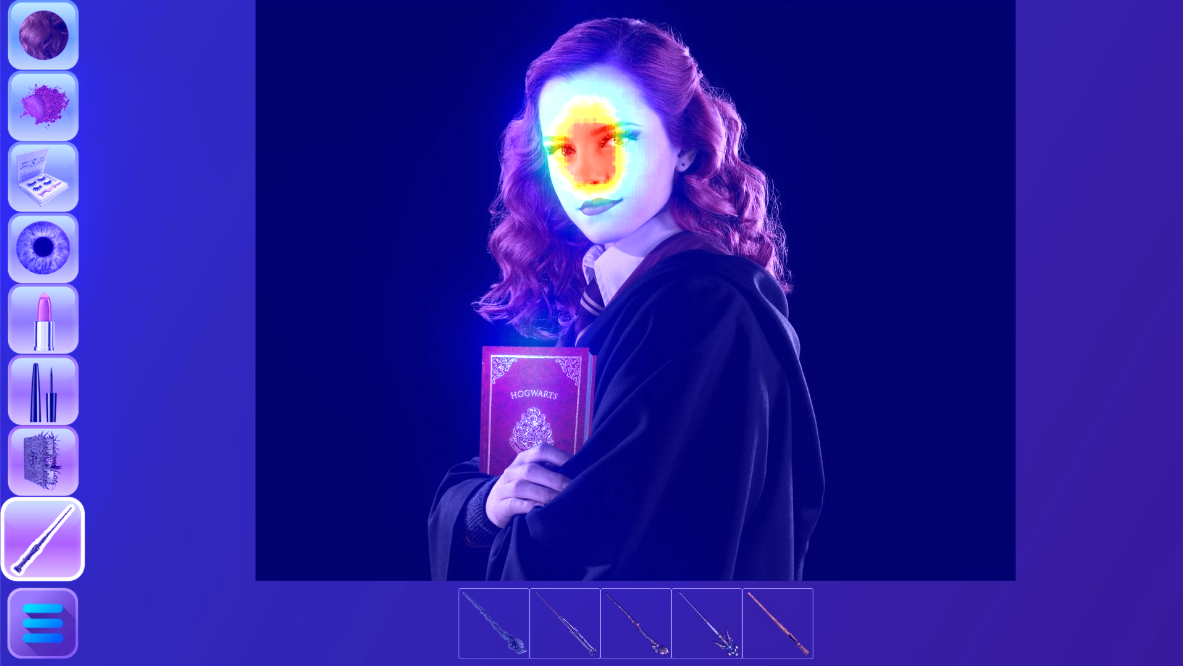} &
    \includegraphics[width=0.1\textwidth]{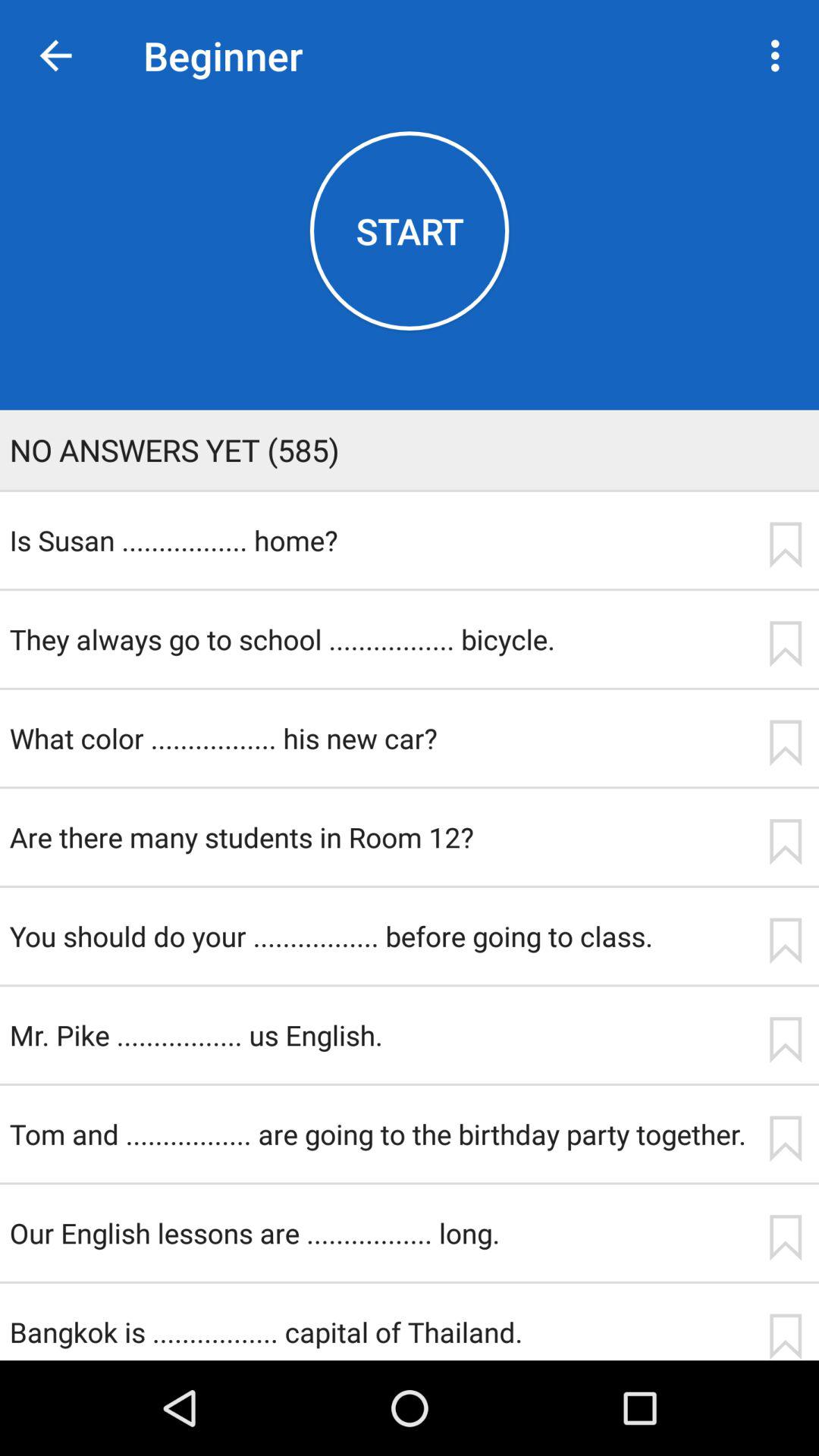} &
    \includegraphics[width=0.1\textwidth]{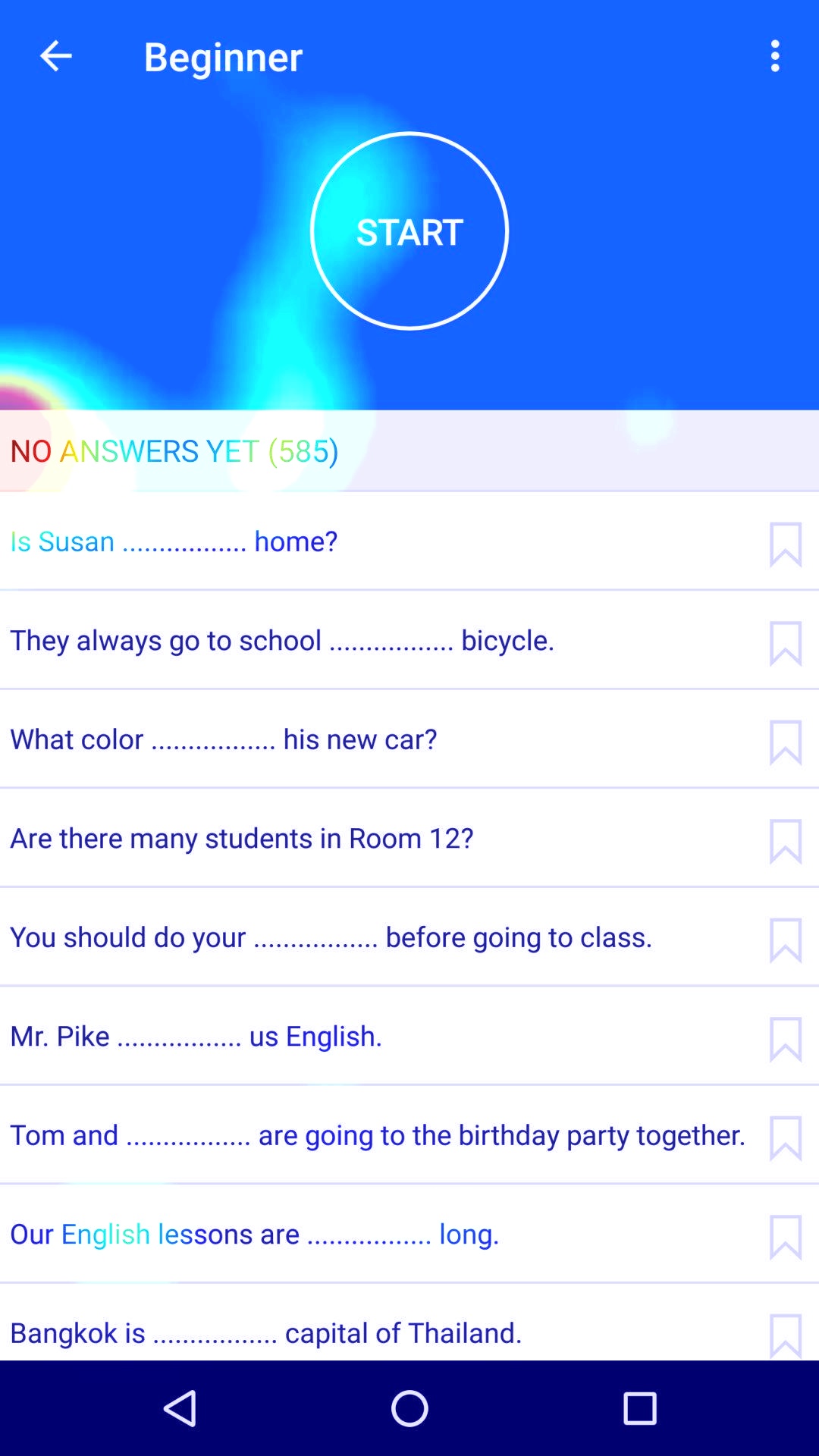} &
    \includegraphics[width=0.1\textwidth]{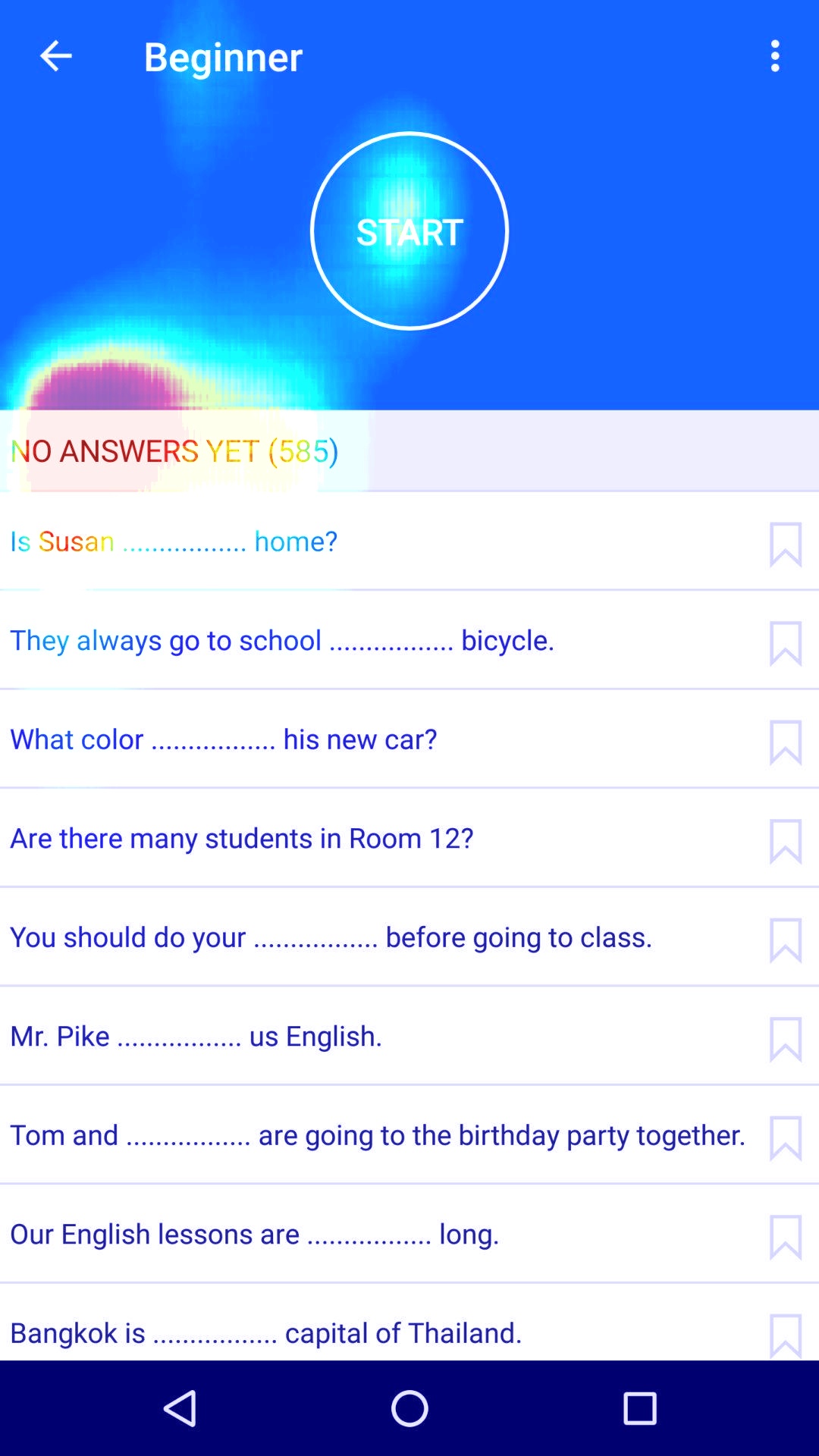}
    \\
    \\
    { \Large Input Image} & {\Large Ground Truth} & {\Large \textbf{SUM}} & 
    {\Large Input Image} & {\Large Ground Truth} & {\Large \textbf{SUM}}\\
    \end{tabular}
}
    \caption{Visualizations of SUM’s predictions across different datasets. The first and second rows depict Natural Scene-Mouse data, while the third and fourth rows showcase Natural Scene-Eye data. The fifth and sixth rows present E-commerce data, and the seventh and eighth rows display UI data.}
    \label{fig:visual_sal_subb}
\end{figure*}

\begin{figure*}[htb]
    \centering
    \resizebox{0.9\textwidth}{!}{
    \begin{tabular}{@{} *{7}c @{}}
    \includegraphics[width=0.25\textwidth]{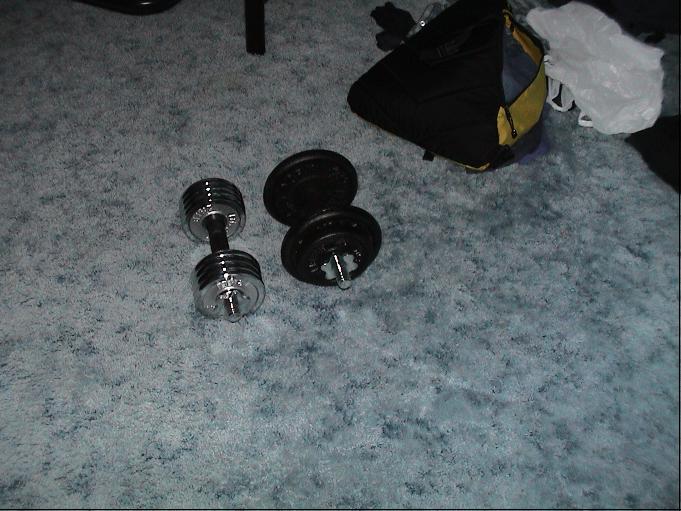} &
    \includegraphics[width=0.25\textwidth]{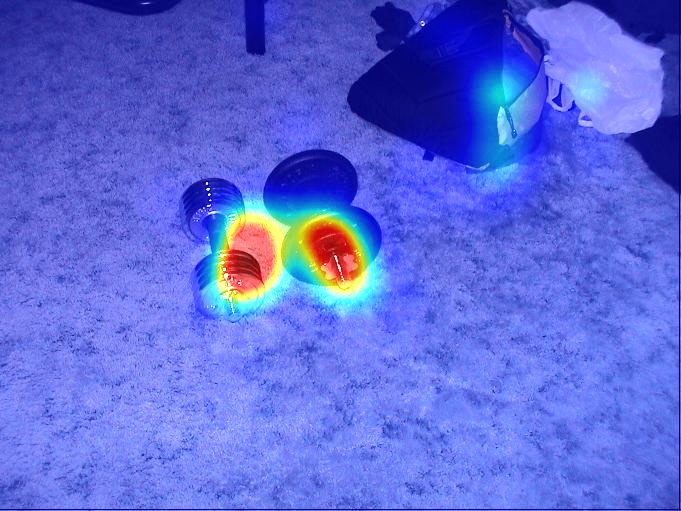} &
    \includegraphics[width=0.25\textwidth]{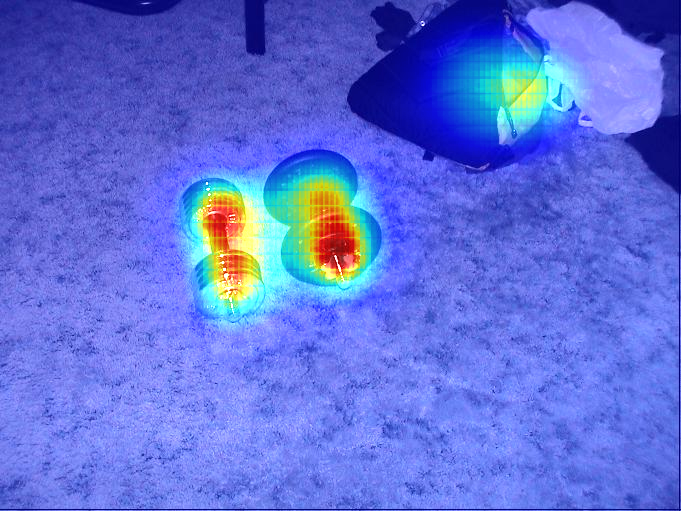} &
    \includegraphics[width=0.25\textwidth]{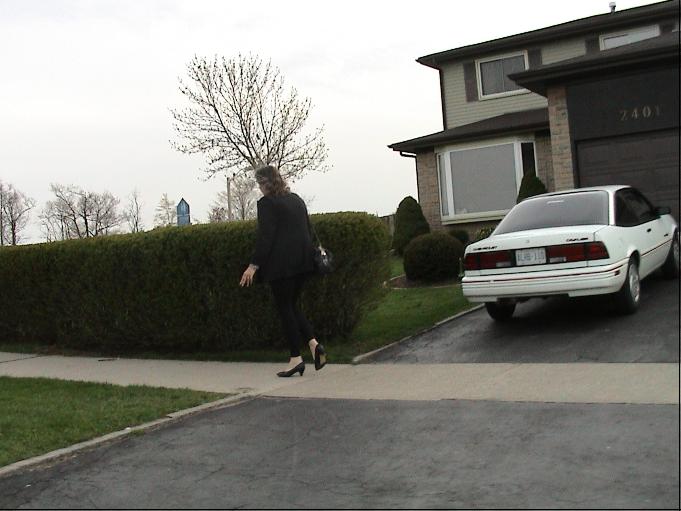} &
    \includegraphics[width=0.25\textwidth]{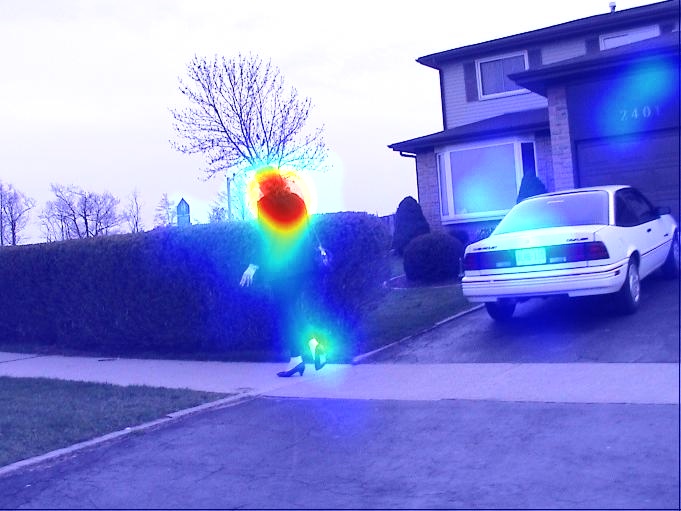} &
    \includegraphics[width=0.25\textwidth]{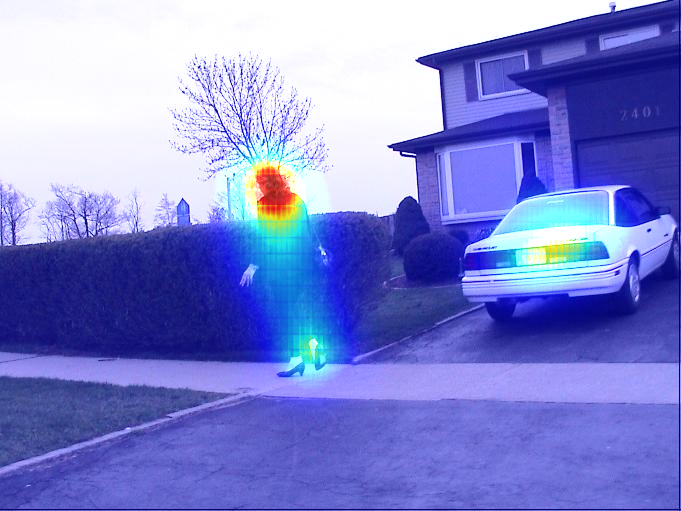} 
    \\
    \\
    \includegraphics[width=0.25\textwidth]{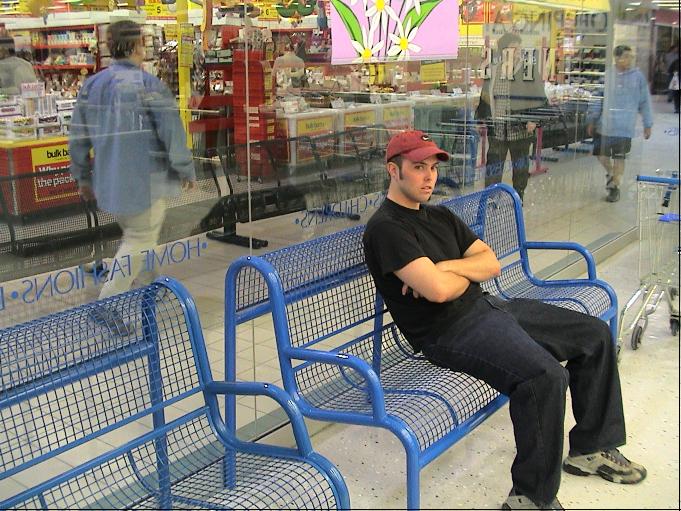} &
    \includegraphics[width=0.25\textwidth]{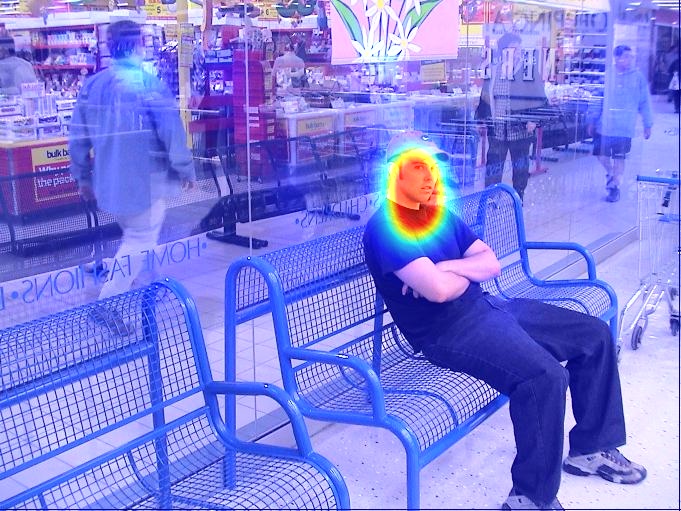} &
    \includegraphics[width=0.25\textwidth]{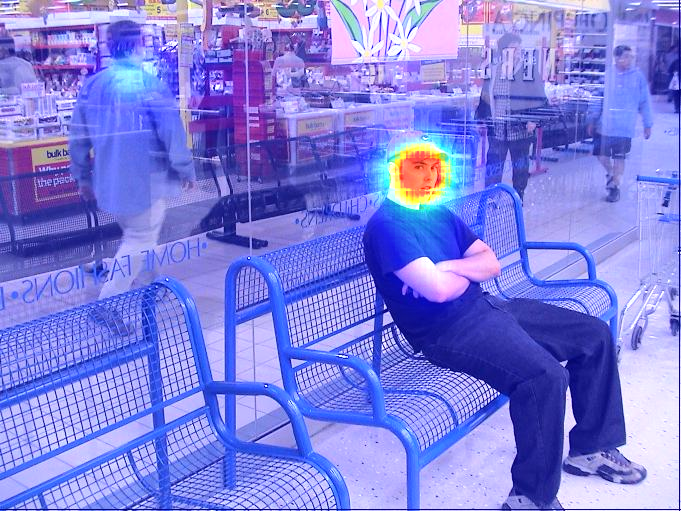} &
    \includegraphics[width=0.25\textwidth]{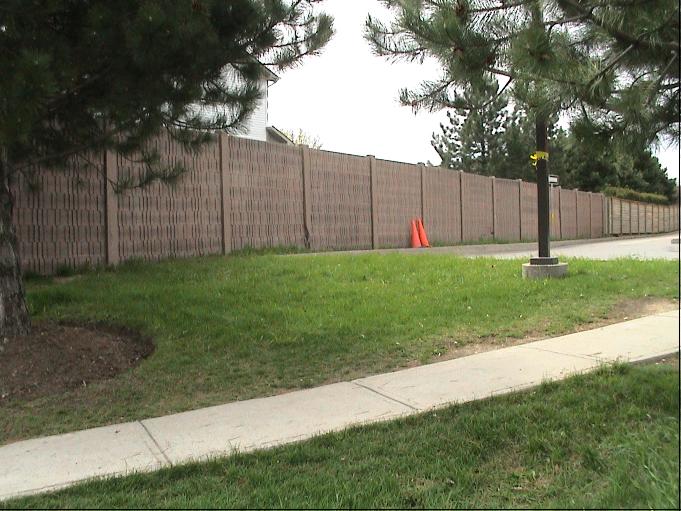} &
    \includegraphics[width=0.25\textwidth]{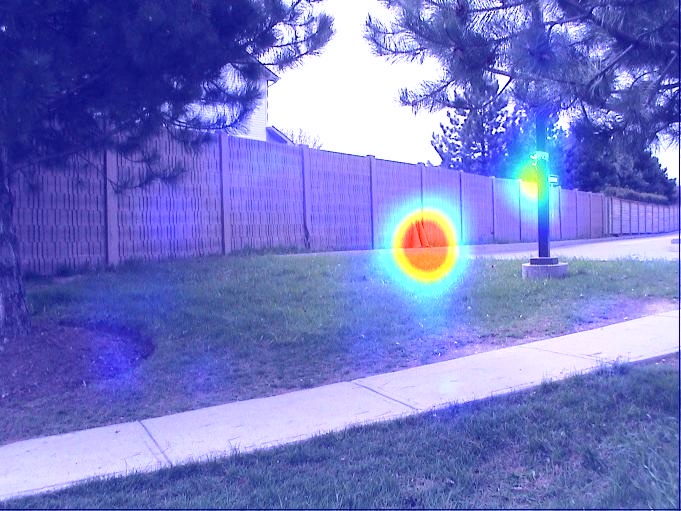} &
    \includegraphics[width=0.25\textwidth]{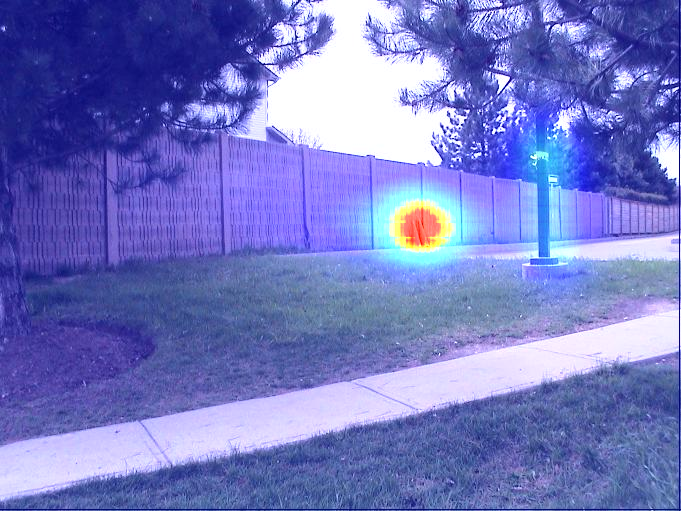} 
    \\
    \\
    \includegraphics[width=0.25\textwidth]{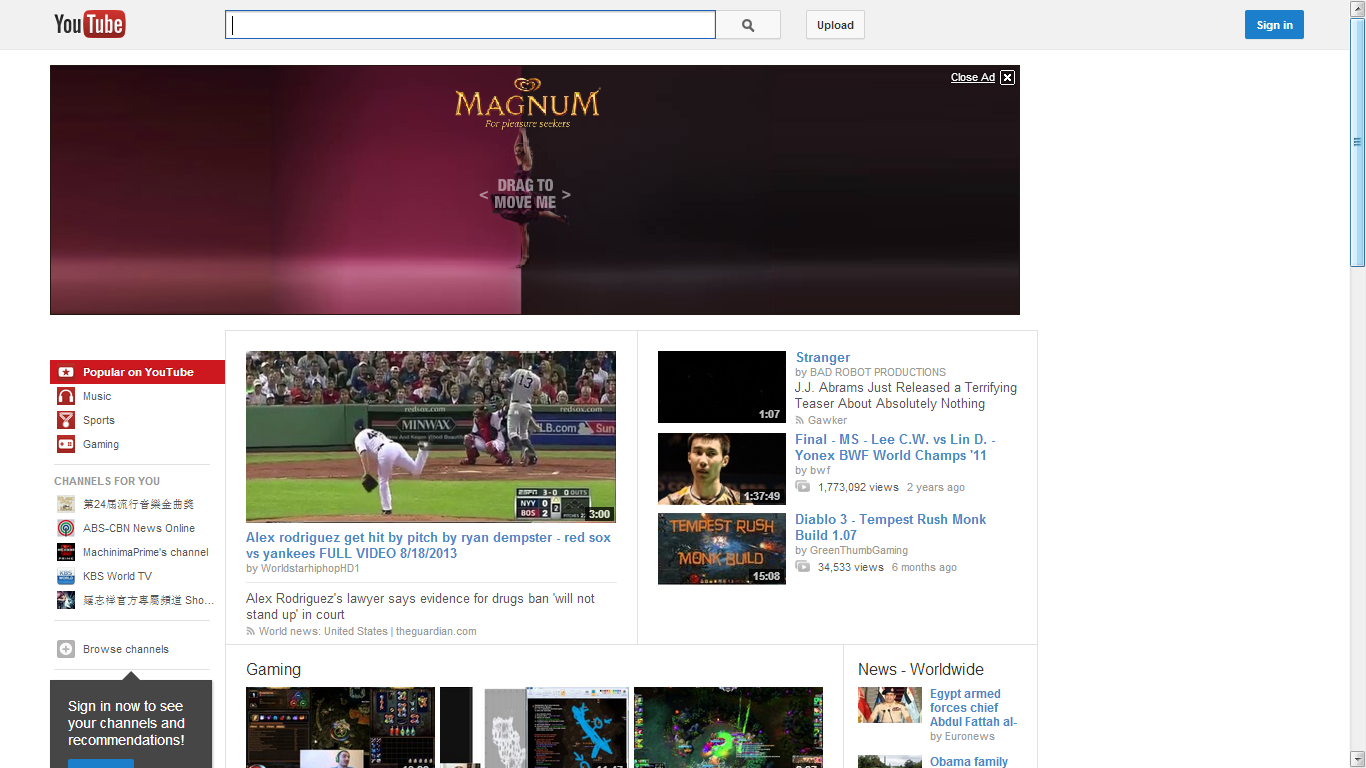} &
    \includegraphics[width=0.25\textwidth]{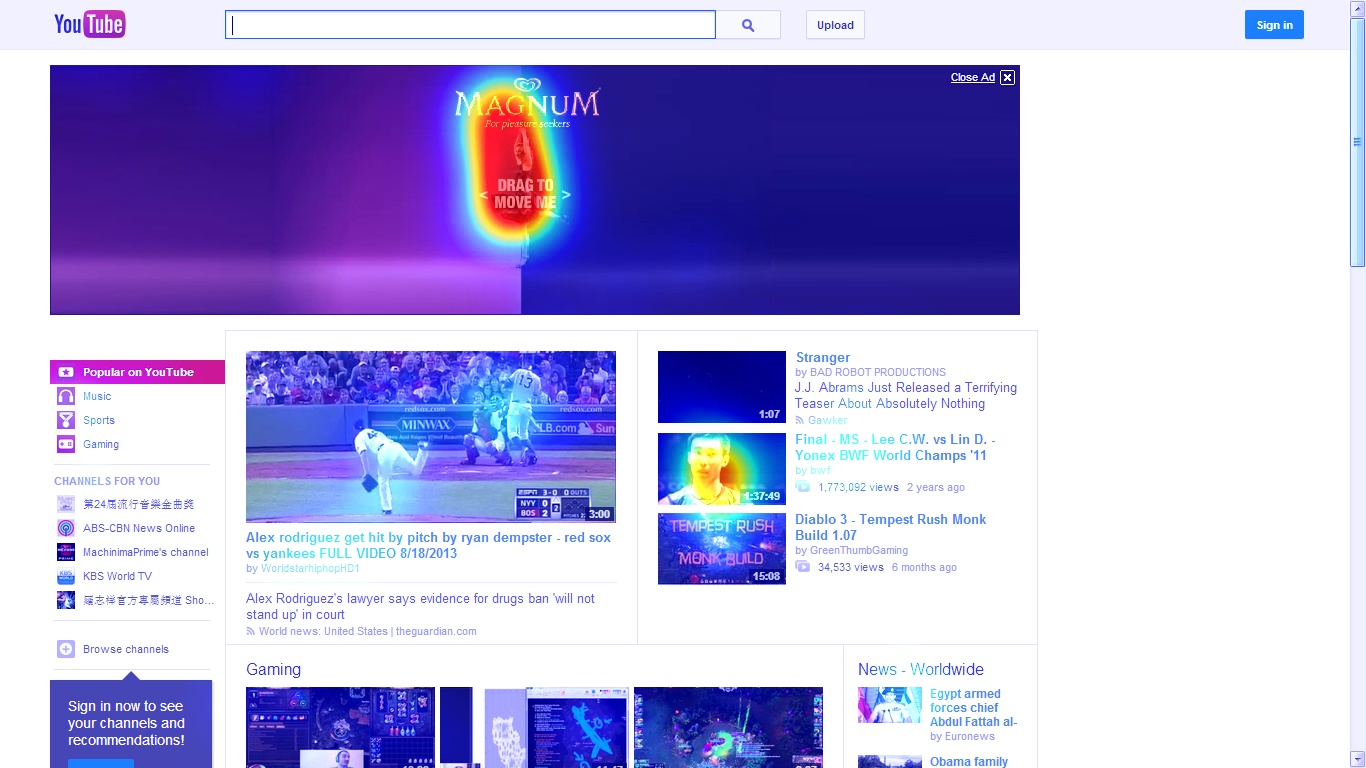} &
    \includegraphics[width=0.25\textwidth]{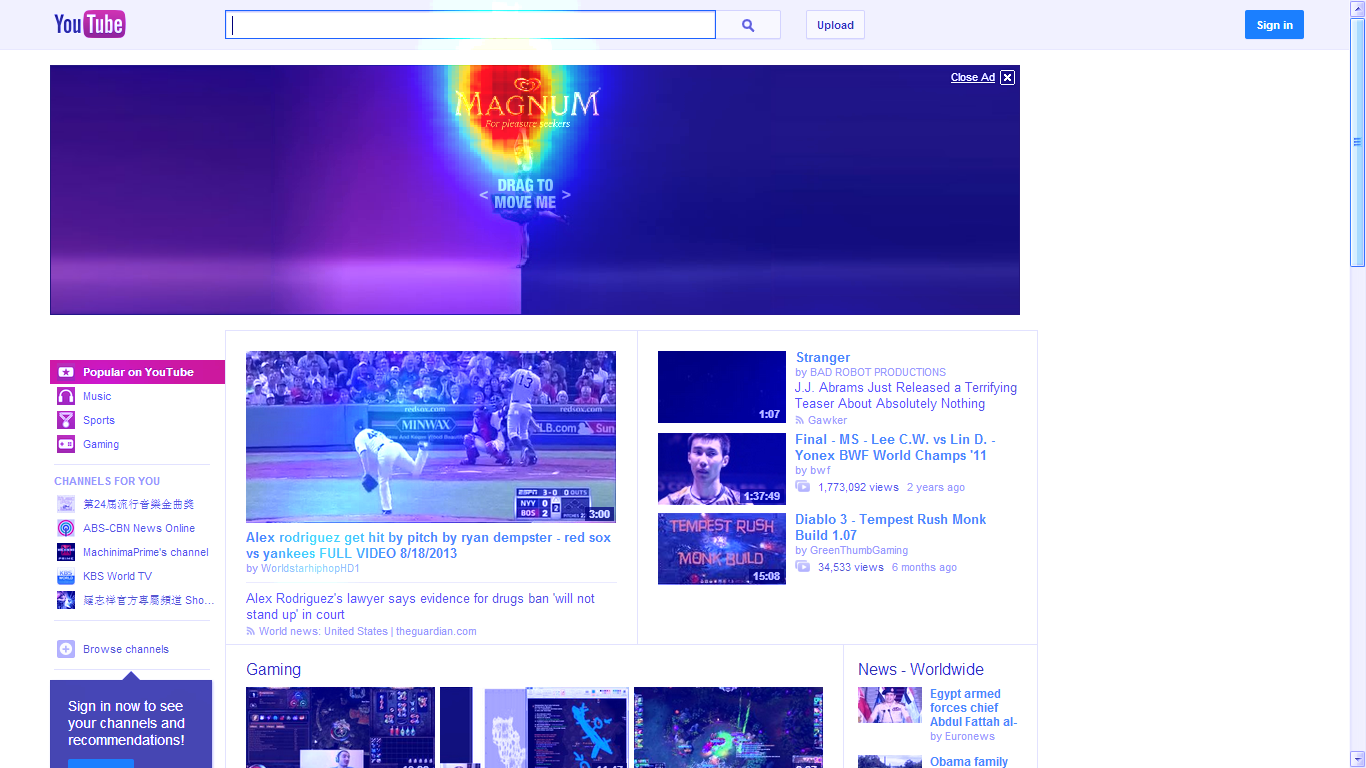} &
    \includegraphics[width=0.25\textwidth]{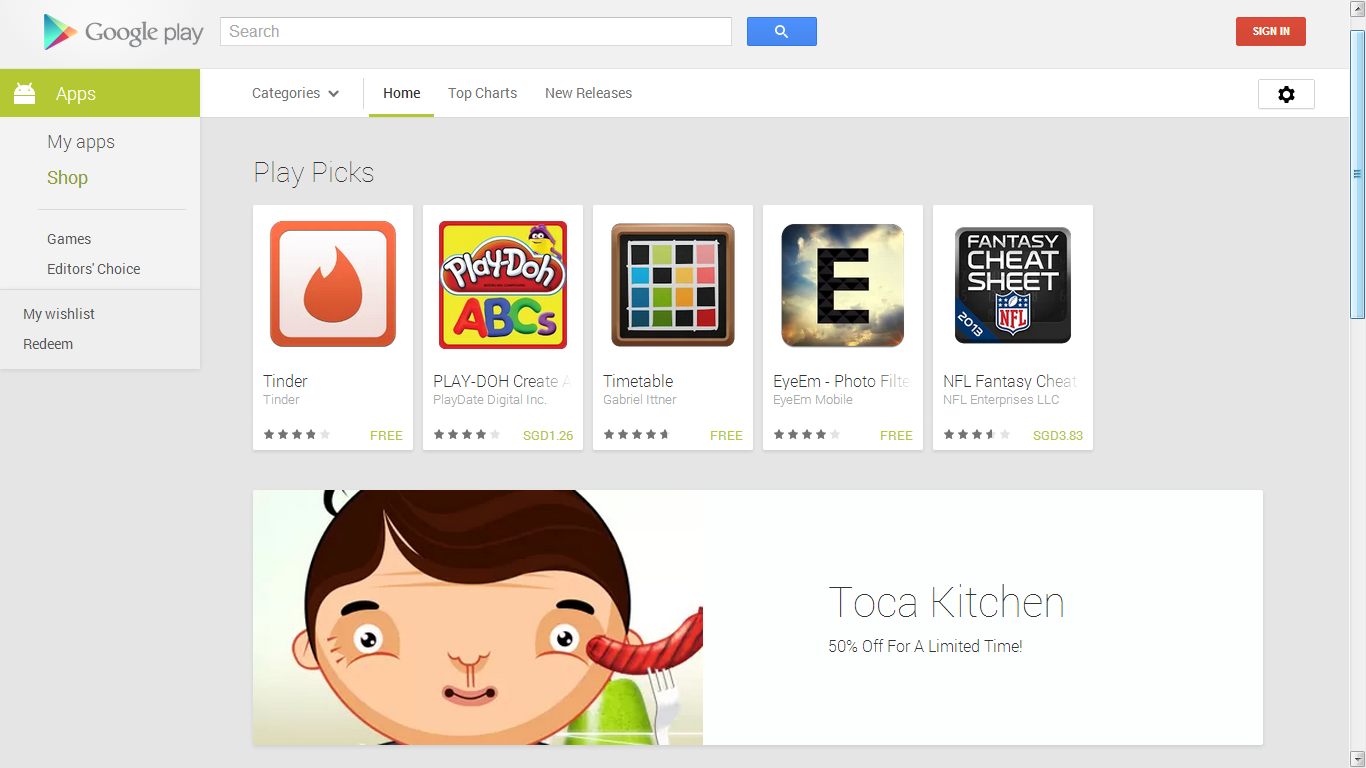} &
    \includegraphics[width=0.25\textwidth]{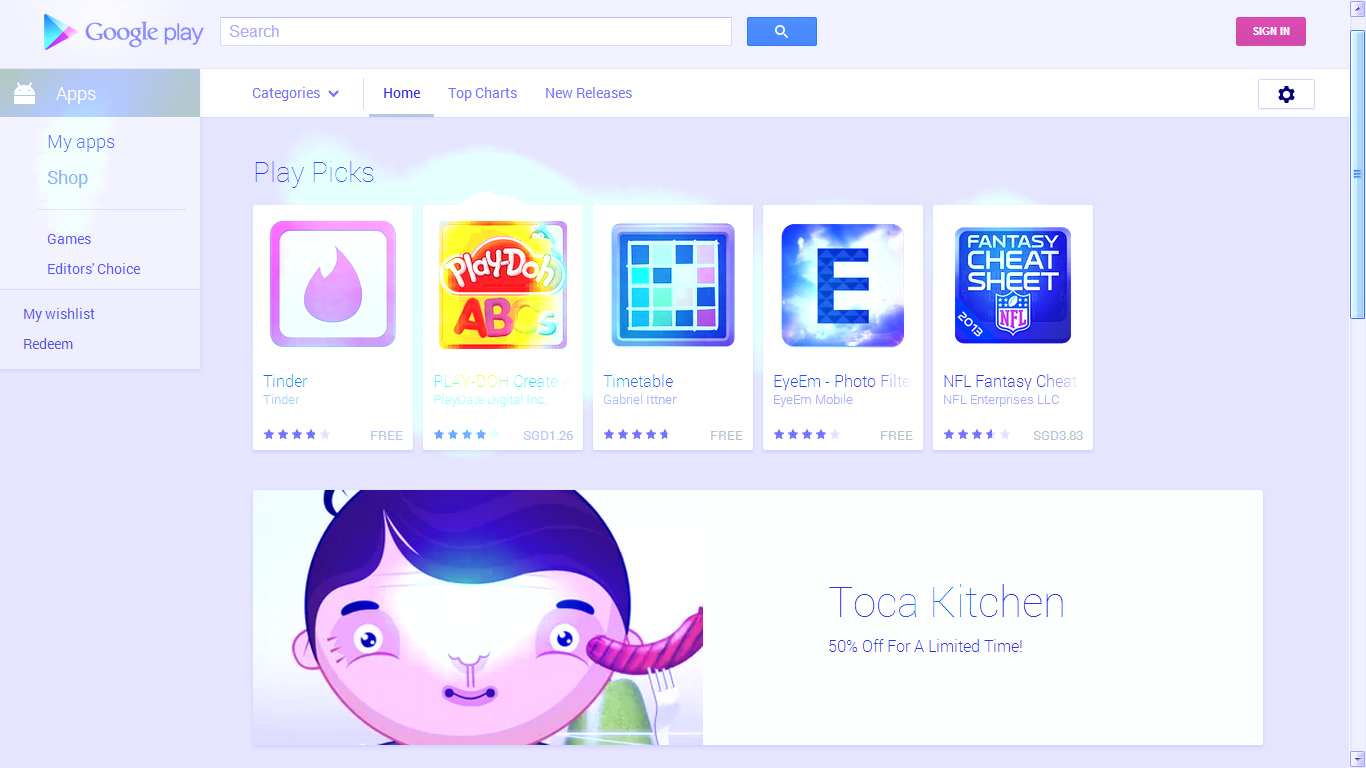} &
    \includegraphics[width=0.25\textwidth]{sec/figures/visualization_other_dataset/FIWI/2_out.png}  
    \\
    \\
    \includegraphics[width=0.25\textwidth]{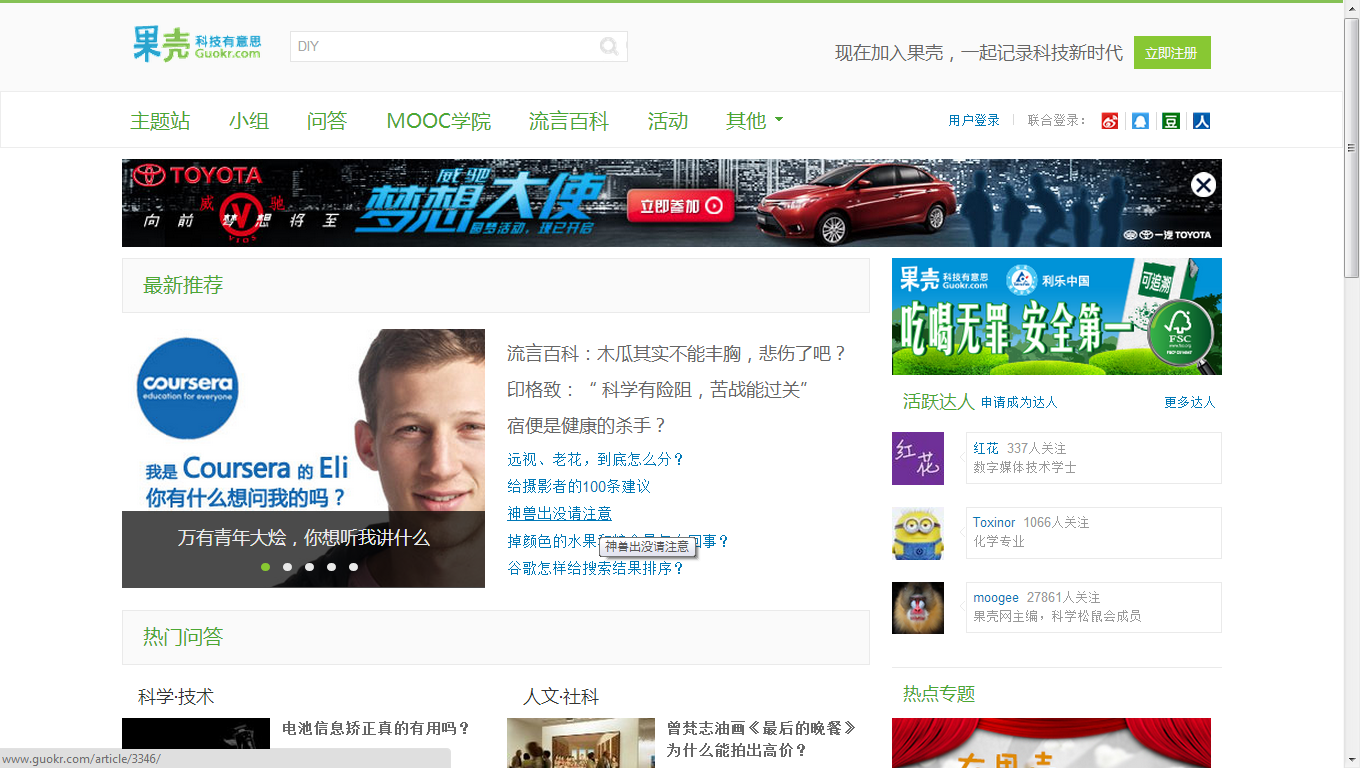} &
    \includegraphics[width=0.25\textwidth]{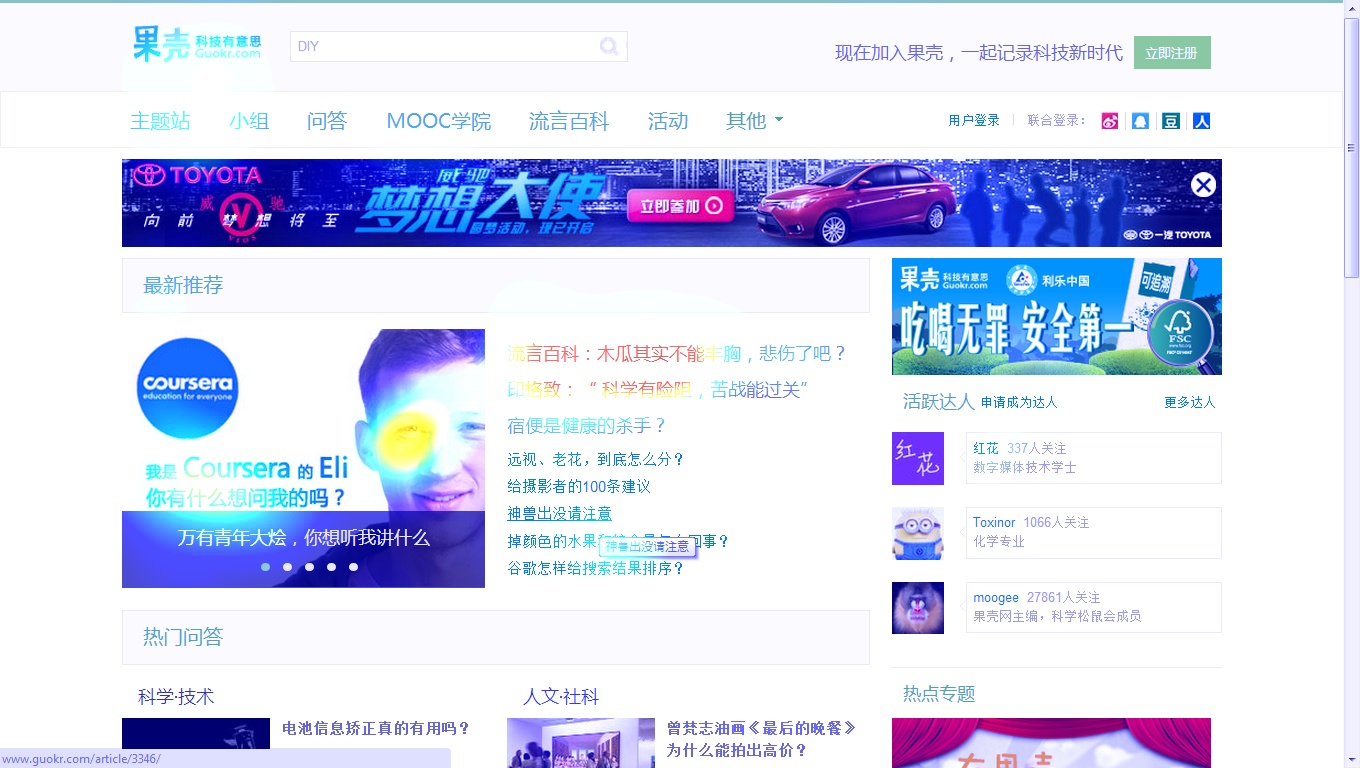} &
    \includegraphics[width=0.25\textwidth]{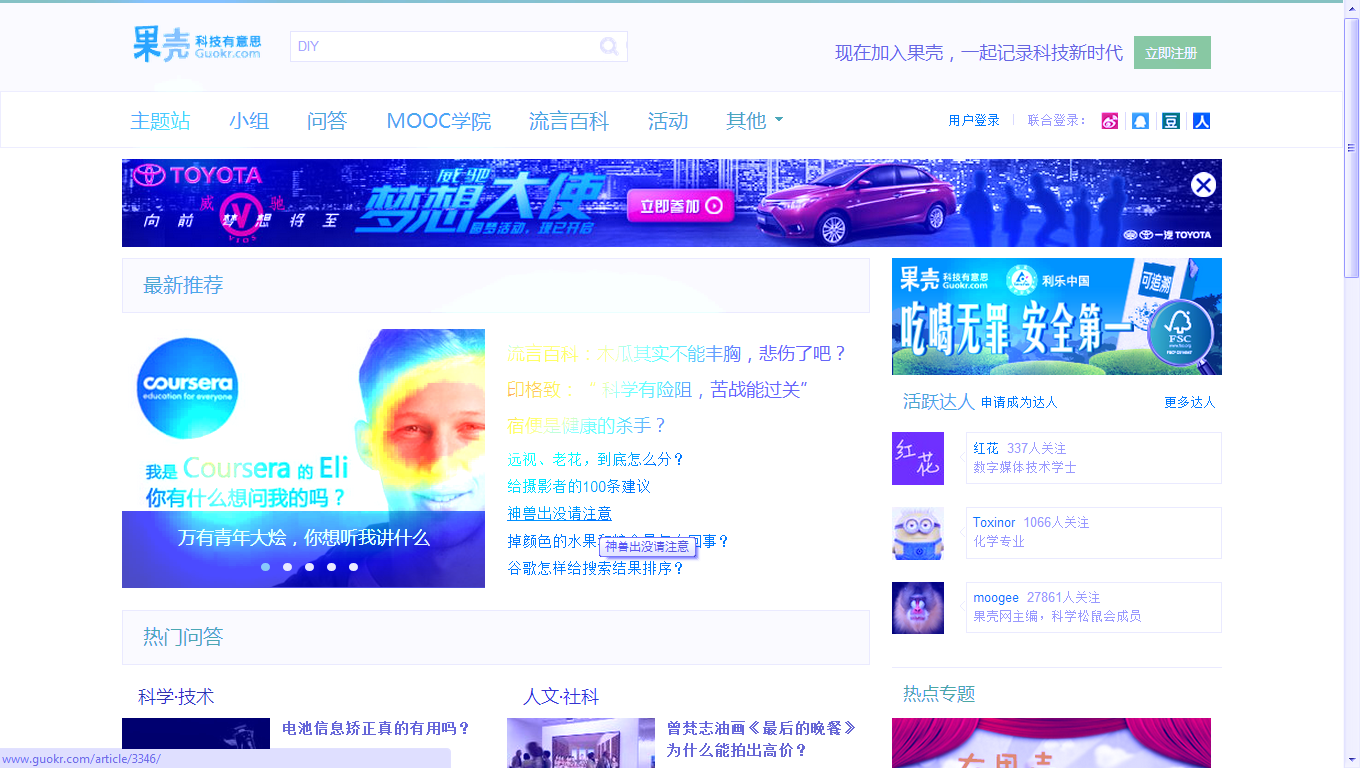} &
    \includegraphics[width=0.25\textwidth]{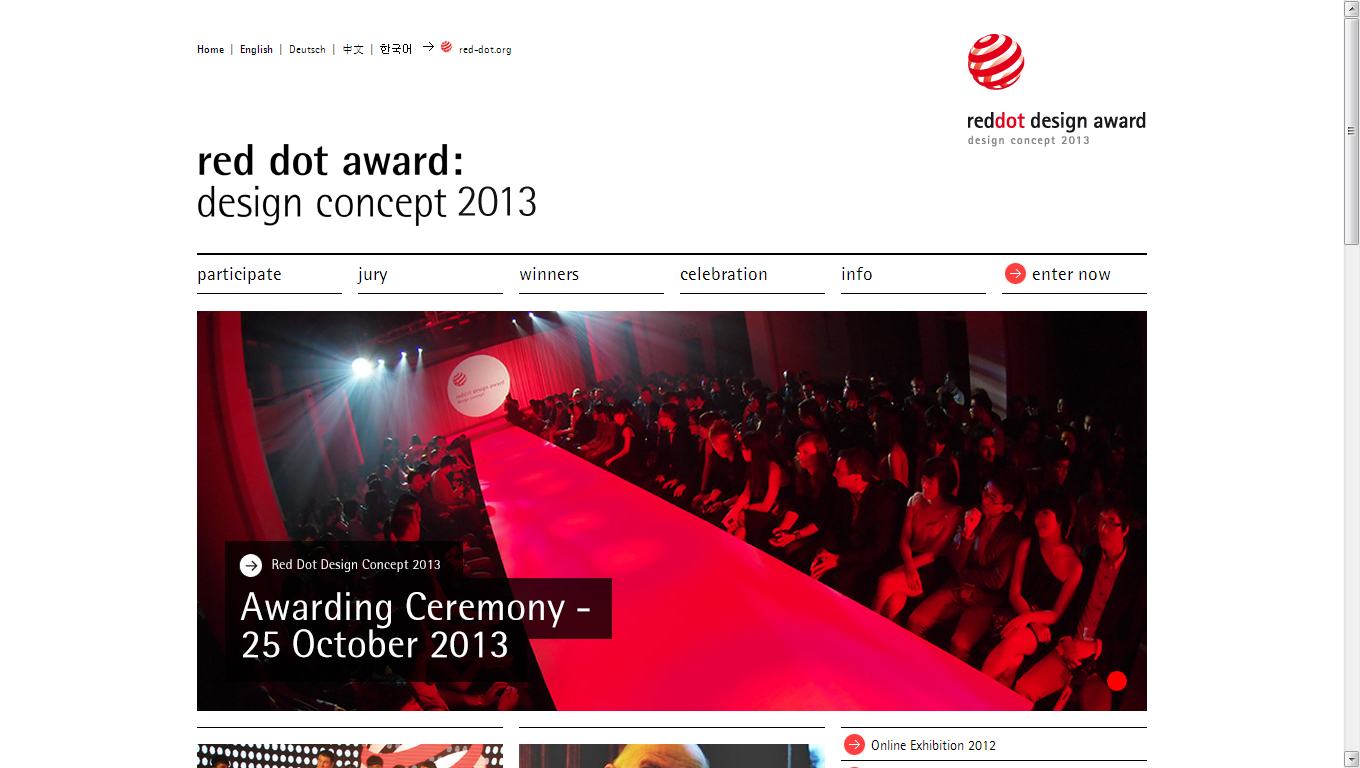} &
    \includegraphics[width=0.25\textwidth]{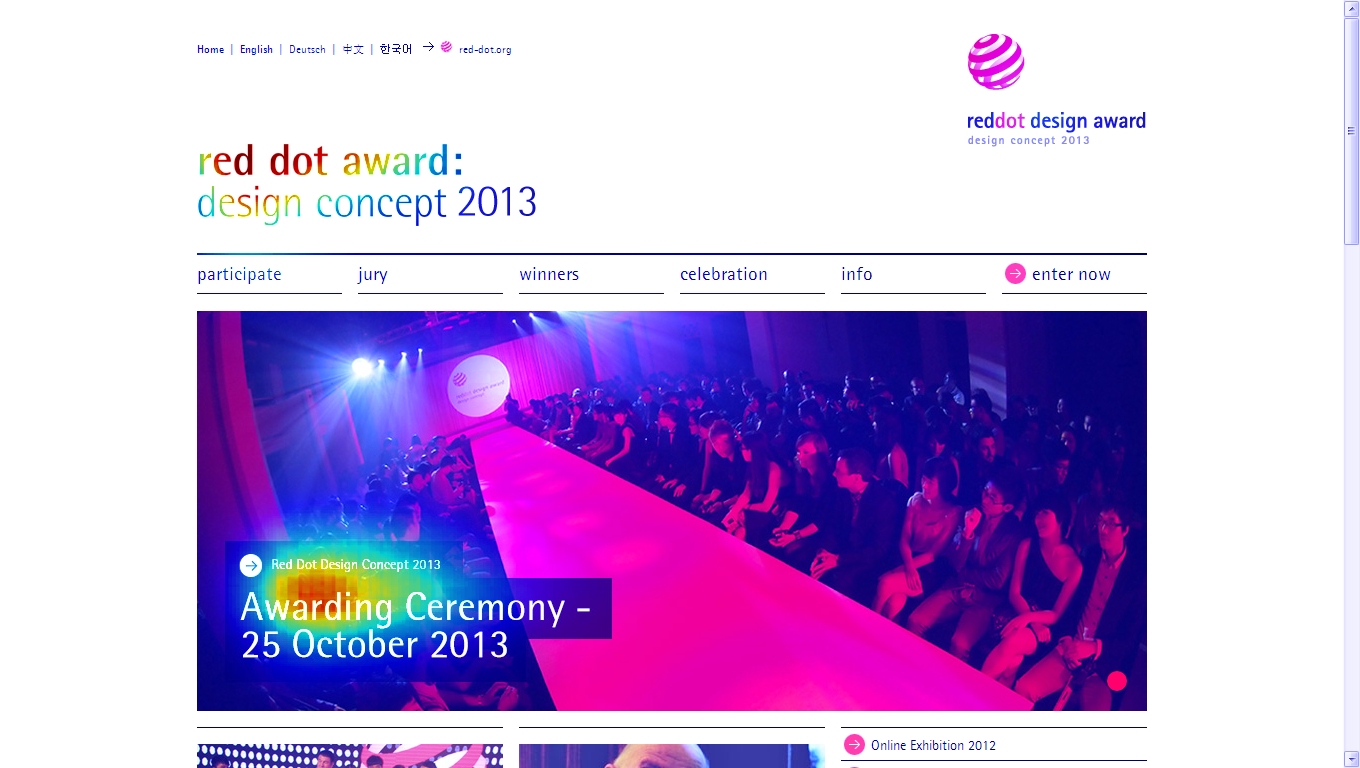} &
    \includegraphics[width=0.25\textwidth]{sec/figures/visualization_other_dataset/FIWI/4_out.png}  
    \\
    \\
    \includegraphics[width=0.25\textwidth]{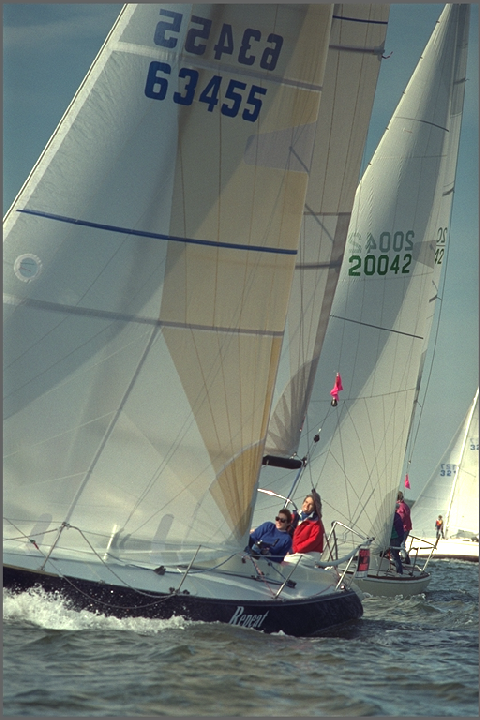} &
    \includegraphics[width=0.25\textwidth]{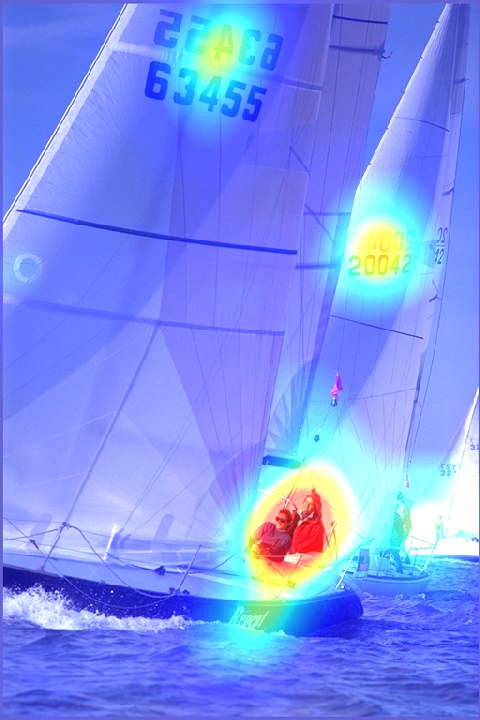} &
    \includegraphics[width=0.25\textwidth]{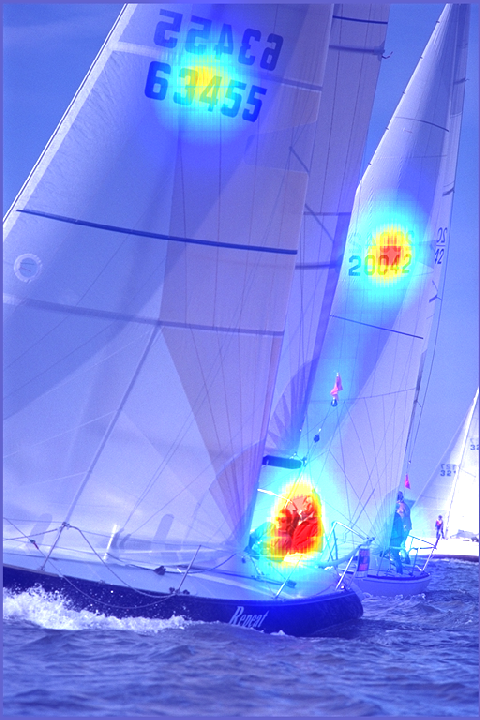} &
    \includegraphics[width=0.25\textwidth]{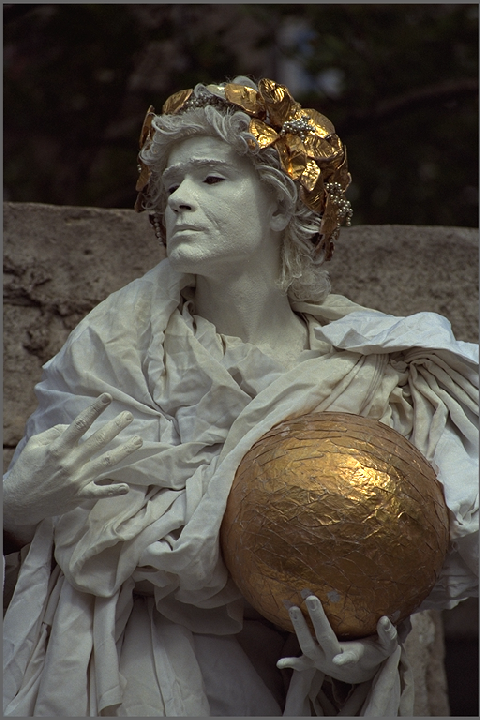} &
    \includegraphics[width=0.25\textwidth]{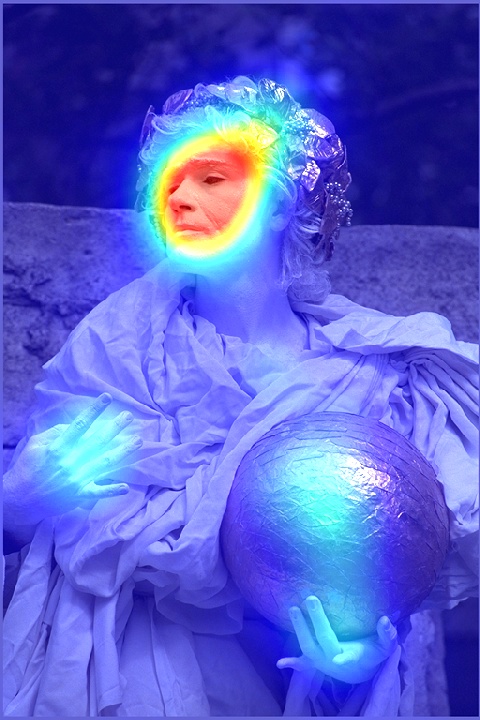} &
    \includegraphics[width=0.25\textwidth]{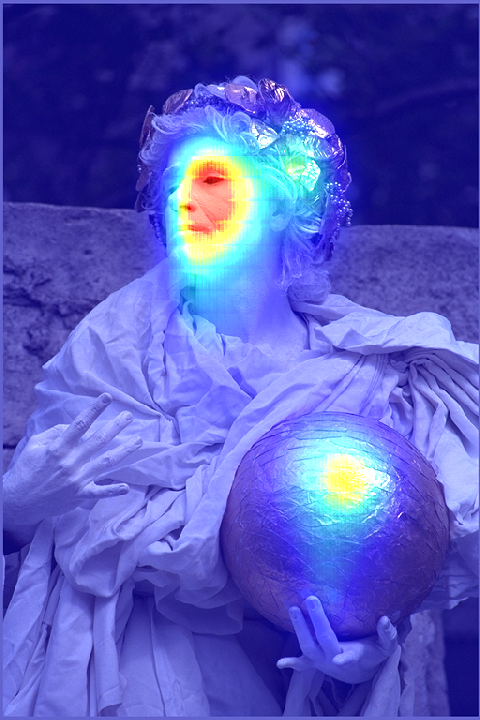} &
    \\
    \\
    \includegraphics[width=0.25\textwidth]{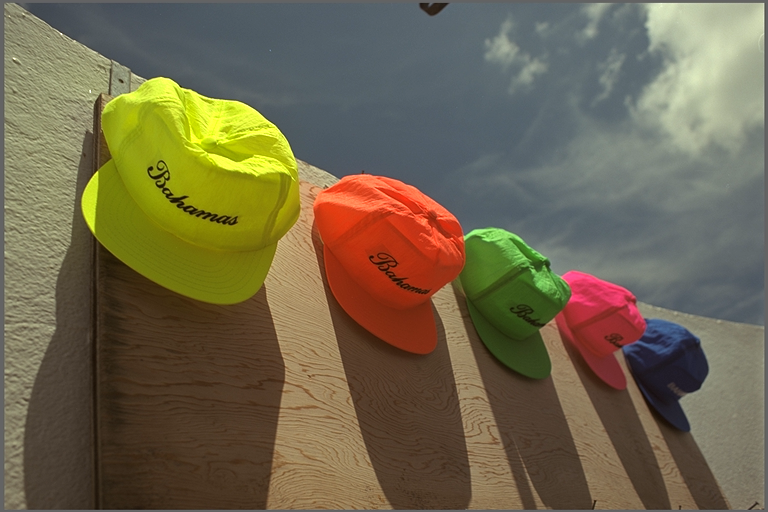} &
    \includegraphics[width=0.25\textwidth]{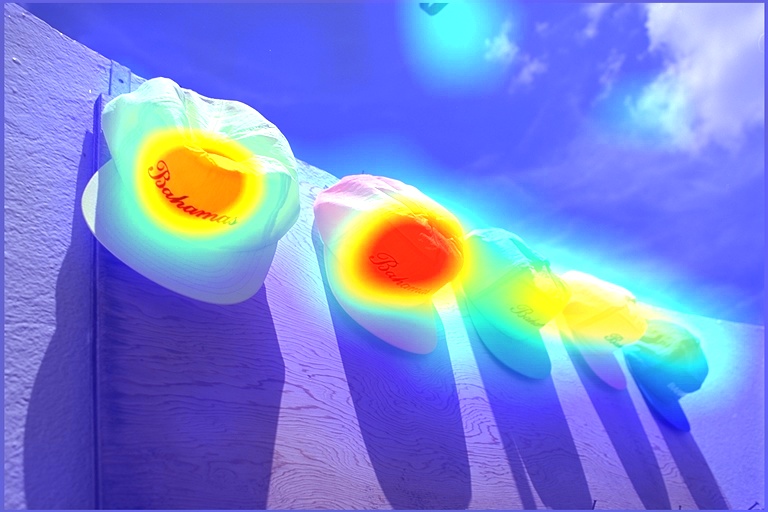} &
    \includegraphics[width=0.25\textwidth]{sec/figures/visualization_other_dataset/TUD1/3_gt.jpg} &
    \includegraphics[width=0.25\textwidth]{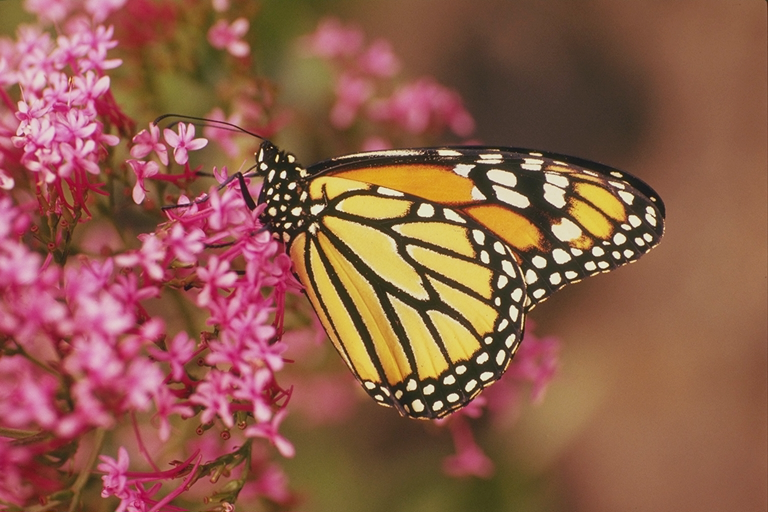} &
    \includegraphics[width=0.25\textwidth]{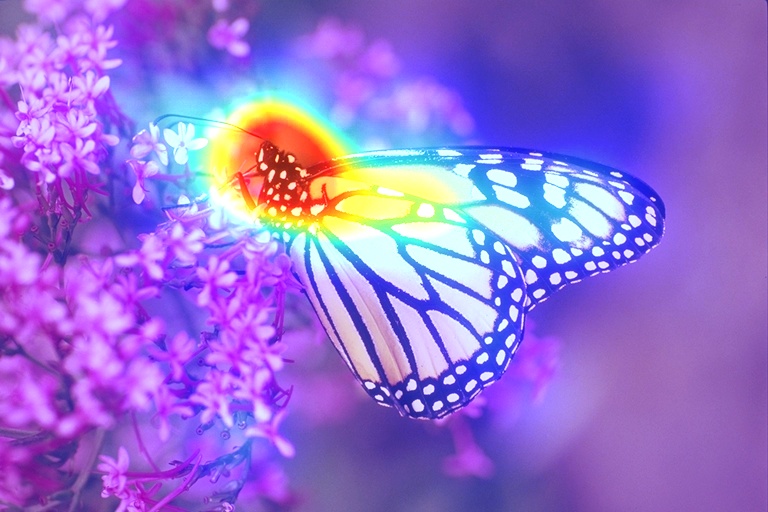} &
    \includegraphics[width=0.25\textwidth]{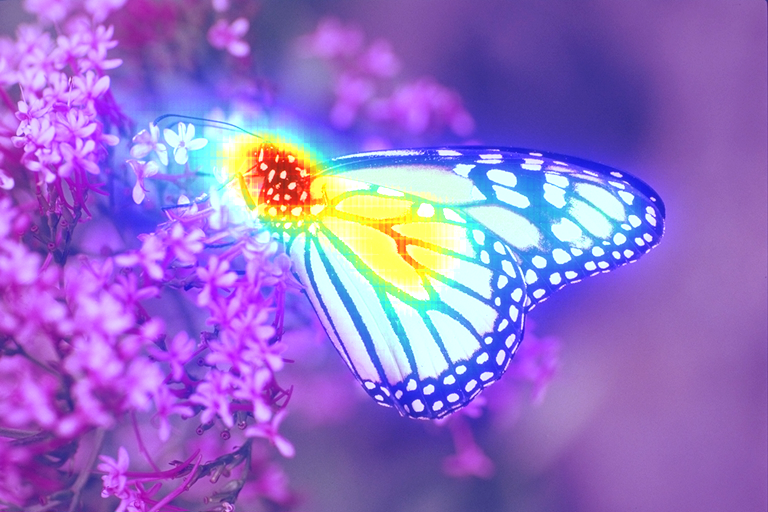} &
    \\
    \\
    \includegraphics[width=0.25\textwidth]{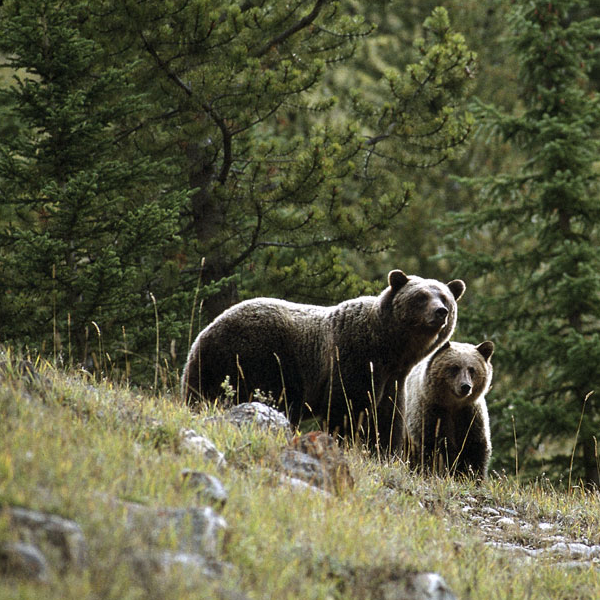} &
    \includegraphics[width=0.25\textwidth]{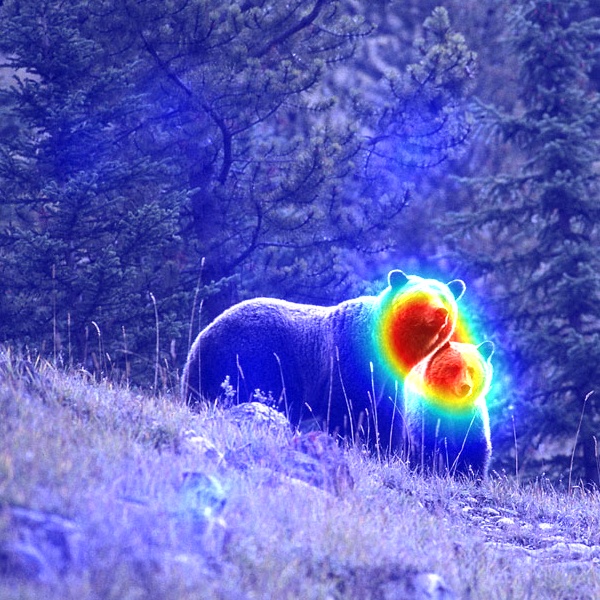} &
    \includegraphics[width=0.25\textwidth]{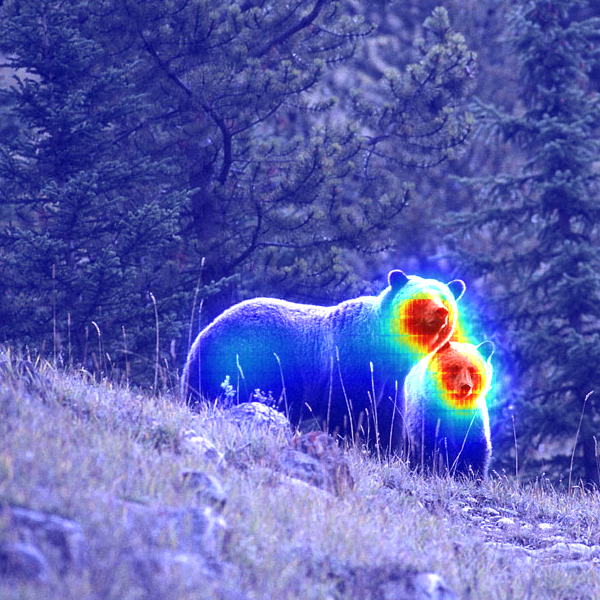} &
    \includegraphics[width=0.25\textwidth]{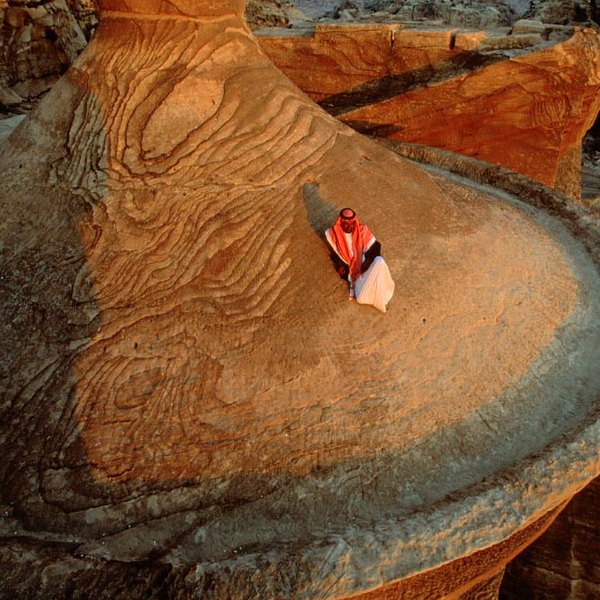} &
    \includegraphics[width=0.25\textwidth]{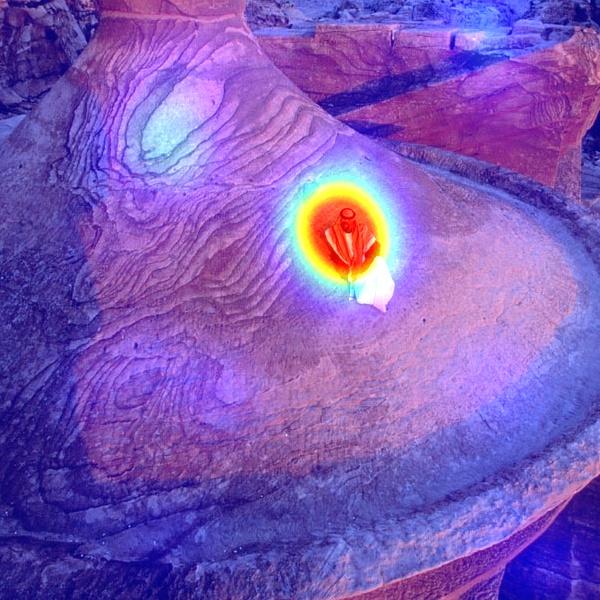} &
    \includegraphics[width=0.25\textwidth]{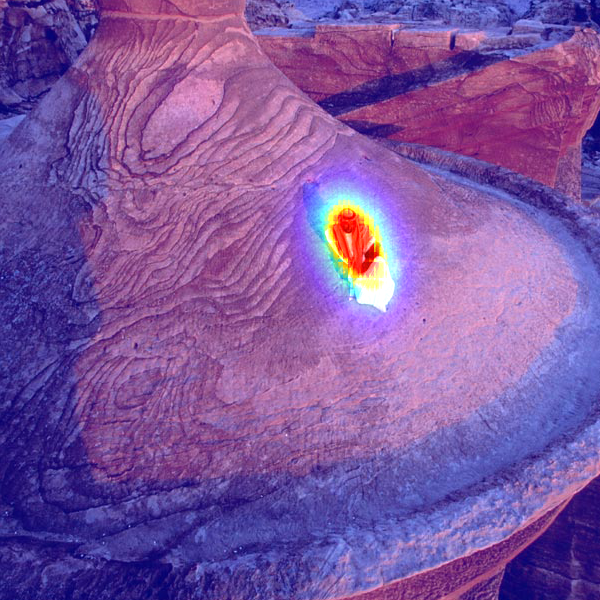} &
    \\
    \\
    \includegraphics[width=0.25\textwidth]{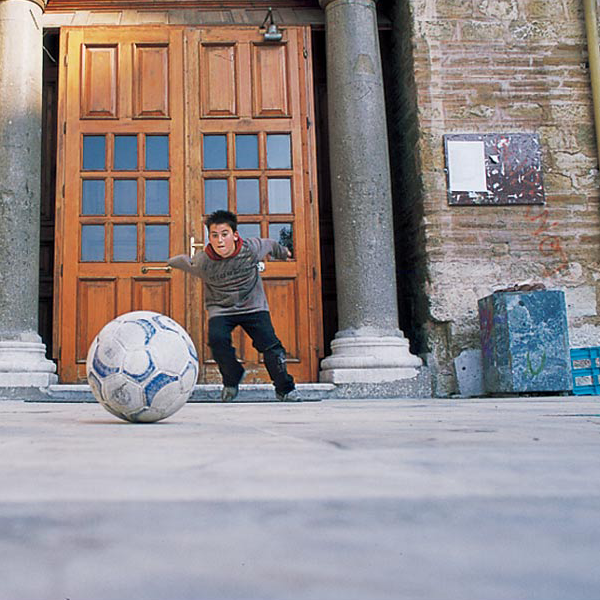} &
    \includegraphics[width=0.25\textwidth]{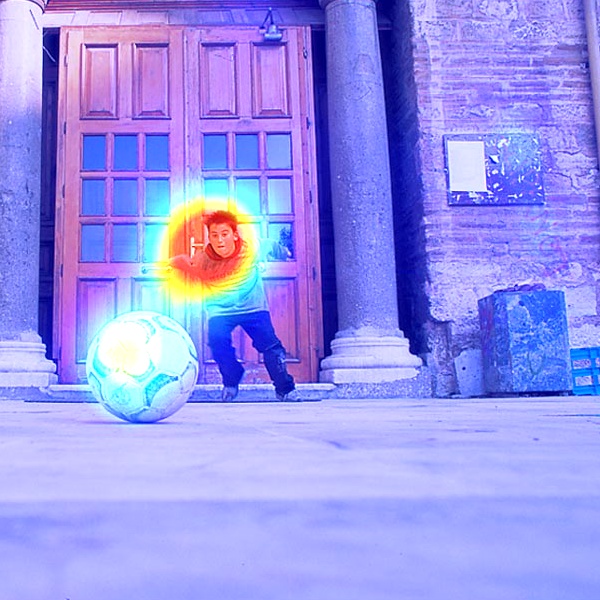} &
    \includegraphics[width=0.25\textwidth]{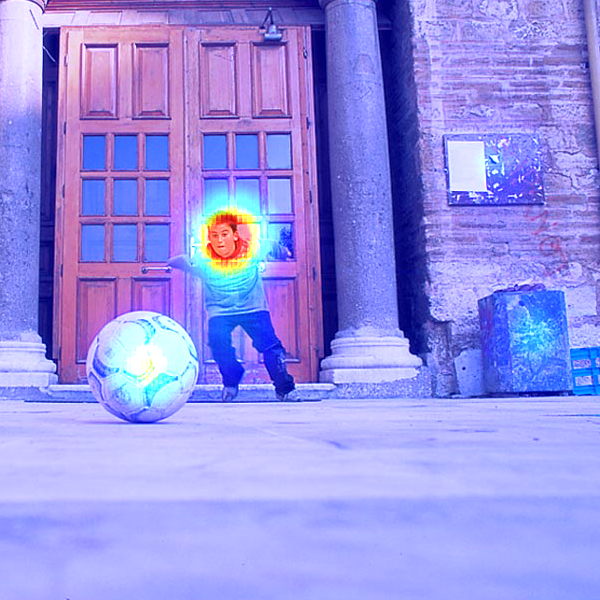} &
    \includegraphics[width=0.25\textwidth]{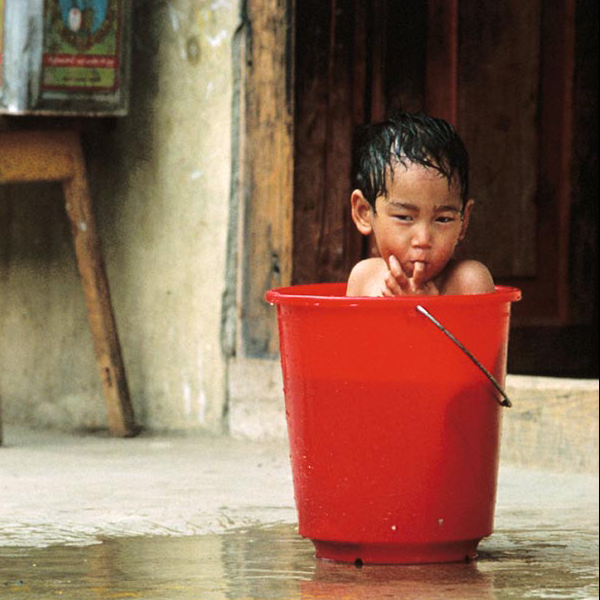} &
    \includegraphics[width=0.25\textwidth]{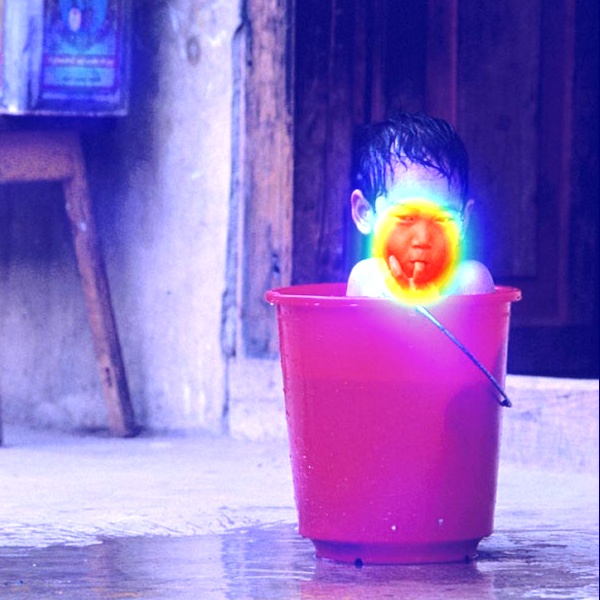} &
    \includegraphics[width=0.25\textwidth]{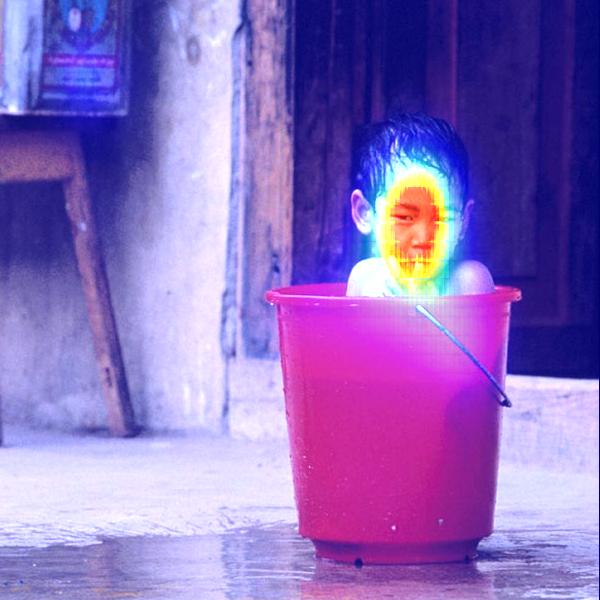} &
    \\
    \\
    { \Large Input Image} & {\Large Ground Truth} & {\Large \textbf{SUM}} & 
    {\Large Input Image} & {\Large Ground Truth} & {\Large \textbf{SUM}}\\
    \end{tabular}
}
    \caption{Visualizations of SUM’s predictions across different datasets. The first and second rows showcase the Toronto dataset~\cite{bruce2007attention}, while the third and fourth rows present the FIWI dataset~\cite{shen2014webpage}. The fifth and sixth rows display data from the TUD Image Quality Database 1~\cite{liu2009studying}, and the seventh and eighth rows exhibit data from the TUD Image Quality Database 2~\cite{alers2010studying}.}
    \label{fig:visual_sal_subb2}
\end{figure*}

We have included an additional visualization of SUM's predictions in \autoref{fig:visual_sal_subb}. Compared to ground truths, SUM consistently delivers accurate predictions across various image types and datasets, underscoring its robustness and versatility in visual saliency modeling. Moreover, to further validate the robustness of our proposed method, we conducted comparative analyses using publicly available datasets that had not been previously seen, as detailed in Table \autoref{datasets}. The performance, as depicted in \autoref{fig:visual_sal_subb2}, notably remains consistent when applied to new and previously unseen datasets. This suggests that SUM adeptly identifies and highlights the salient features in images, maintaining close alignment with the ground truth data. Therefore, SUM can be reliably utilized in diverse real-world applications where accuracy in visual recognition is critical.

\begin{table}[ht]
\centering
\caption{loss weighting coefficients \(\lambda_i\) (\(i = 1, \ldots, 5\)) as used in \autoref{tab:different_loss}.}
\label{tab:loss_coef}
\resizebox{0.6\columnwidth}{!}{
    \begin{tabular}{ccccc}
    \hline 
    KL & CC & SIM & NSS & MSE \\
    \hline  
    1 & 0 & 0 & 0 & 0 \\
    0 & -1 & 0 & 0 & 0 \\
    0 & 0 & -1 & 0 & 0 \\
    0 & 0 & 0 & -1 & 0 \\
    \hline 
    0 & 0 & 0 & 0 & 1 \\
    10 & 0 & -3 & 0 & 0 \\
    10 & 0 & 0 & -3 & 0 \\
    10 & -3 & 0 & 0 & 0 \\
    \hline 
    10 & 0 & 0 & 0 & 5 \\
    0 & -2 & 0 & -1 & 0 \\
    10 & -2 & -1 & 0 & 0 \\
    10 & 0 & -3 & 0 & 5 \\
    \hline 
    10 & 0 & 0 & -3 & 5 \\
    10 & -3 & 0 & 0 & 5 \\
    10 & -2 & -1 & -1 & 0 \\
    10 & -2 & -1 & 0 & 5 \\
    \hline 
    10 & -2 & -1 & -1 & 5 \\
    \end{tabular}
}
\end{table}

\begin{table}[t]
\caption{Details of unseen datasets used for quantitative analysis of SUM in \autoref{fig:visual_sal_subb2}.}
\label{tab:other datasets}
\centering
\resizebox{\columnwidth}{!}{
\begin{tabular}{lcccccc}
\toprule
Dataset & Image domain &  \# Image & Image Resolution  \\
\midrule
\textit{Toronto}~\cite{bruce2007attention} & Natural scene  & 120 & 681 $\times$ 511 \\
\textit{TUD Image Quality Database 1}~\cite{liu2009studying} & Natural scene & 29 & 768 $\times$ 512\\
\textit{TUD Image Quality Database 2}~\cite{alers2010studying} & Natural scene & 160 & 600 $\times$ 600\\
\textit{FIWI}~\cite{shen2014webpage} & Web page & 149 & 1360 $\times$ 768 \\

\bottomrule
\end{tabular}
}
\label{datasets}
\vspace{-0.5em}
\end{table}

\end{document}